%% file: main.tex
\documentclass[10pt,twocolumn,letterpaper]{article}

\usepackage{cvpr}

\input{preamble}

\definecolor{cvprblue}{rgb}{0.21,0.49,0.74}
\usepackage[accsupp]{axessibility}
\usepackage[table]{xcolor}
\usepackage{mathabx}
\usepackage{amssymb}
\usepackage{booktabs}
\usepackage{arydshln}
\usepackage{multirow}
\usepackage{makecell}
\usepackage{pifont}
\usepackage[latin, english]{babel}
\usepackage{titletoc}
\usepackage{tcolorbox}
\usepackage[linesnumbered,ruled,longend,commentsnumbered]{algorithm2e}
\usepackage[pagebackref,breaklinks,colorlinks,allcolors=cvprblue]{hyperref}
\usepackage[capitalize,noabbrev]{cleveref}
\usepackage{float}
\usepackage{newfloat}

\usepackage{listings}
\DeclareCaptionStyle{ruled}{labelfont=normalfont,labelsep=colon,strut=off}
\floatstyle{ruled}
\newfloat{listing}{tb}{lst}{}
\floatname{listing}{Listing}
\makeatletter
\def\adl@drawiv#1#2#3{%
        \hskip.5\tabcolsep
        \xleaders#3{#2.5\@tempdimb #1{1}#2.5\@tempdimb}%
                #2\z@ plus1fil minus1fil\relax
        \hskip.5\tabcolsep}
\newcommand{\cdashlinelr}[1]{%
  \noalign{\vskip\aboverulesep
           \global\let\@dashdrawstore\adl@draw
           \global\let\adl@draw\adl@drawiv}
  \cdashline{#1}
  \noalign{\global\let\adl@draw\@dashdrawstore
           \vskip\belowrulesep}}
\makeatother

\newcommand{\myparagraph}[1]{\vspace{1pt}\noindent{\bf{#1}}~~}
\newcommand{\cmark}{\textcolor{green!60!black}{\checkmark}}
\newcommand{\xmark}{\textcolor{red}{\ding{55}}}
\newcommand{\tmark}{\textcolor{yellow}{$\triangle$}}

\def\paperID{45166}
\def\confName{CVPR}
\def\confYear{2026}

\title{PosterGen: Aesthetic-Aware Multi-Modal Paper-to-Poster Generation via Multi-Agent LLMs}

\author{
Zhilin Zhang\textsuperscript{1,2}\footnotemark[1] \quad
Xiang Zhang\textsuperscript{3}$^*$ \quad
Jiaqi Wei\textsuperscript{4} \quad
Yiwei Xu\textsuperscript{5} \quad
Chenyu You\textsuperscript{1}\footnotemark[2]  \\
\textsuperscript{1}Stony Brook University \quad 
\textsuperscript{2}New York University \quad
\textsuperscript{3}University of British Columbia \quad \\
\textsuperscript{4}Zhejiang University \quad 
\textsuperscript{5}University of California, Los Angeles \\
}

\begin{document}
\maketitle
\renewcommand{\thefootnote}{\fnsymbol{footnote}}
\footnotetext[1]{Equal contribution.}
\footnotetext[2]{Corresponding author.}
\input{sec/0_abstract}    
\input{sec/1_intro}
\input{sec/2_rw}
\input{sec/3_design}
\input{sec/4_method}
\input{sec/5_exp}
\input{sec/6_conclusion}

{
    \small
    \bibliographystyle{ieeenat_fullname}
    \bibliography{main}
}

\input{sec/X_suppl}

\end{document}

%% file: preamble.tex
\newcommand{\TODO}[1]{\textbf{\color{red}[TODO: #1]}}
\renewcommand{\TODO}[1]{}



%% file: sec/0_abstract.tex
\begin{abstract}
Multi-agent systems built on large language models (LLMs) and multimodal large language models (MLLMs) have recently shown strong capabilities on complex compositional tasks. We apply this paradigm to paper-to-poster generation, a practical but time-consuming step in research communication. Most of the existing methods ignore core design and aesthetic principles, and often produce posters that still require substantial manual refinement.
To address these limitations, we propose \textbf{PosterGen}, a multi-agent framework that embeds fundamental design principles into a specialized agent workflow that mirrors professional designers. Our system incorporates collaborative agents to distill a paper's narrative (CuratorAgent), map content to a balanced spatial layout (LayoutAgent), apply a cohesive visual system of color and typography (StylistAgents), and render the final poster in PPTX format (Renderer). This design-centric approach produces posters that are both semantically grounded and visually compelling.
To systematically evaluate visual quality, we introduce a comprehensive VLM-based rubric measuring information fidelity and aesthetic design, which we validate against human preferences in a user study. Experimental results show that PosterGen achieves both content fidelity and design aesthetics comparable to human-designed posters while significantly and consistently outperforming state-of-the-art methods. 
Code is publicly available at \href{https://github.com/Y\string-Research\string-SBU/PosterGen}{https://github.com/Y-Research-SBU/PosterGen}.
\end{abstract}

%% file: sec/1_intro.tex
\section{Introduction}
\label{sec:intro}

Academic posters serve as an indispensable venue for fast, visual, and interactive research communication for conferences, as they provide a medium for direct, one-on-one dialogue between the target audience and the authors~\citep{faulkes2021better}. However, many researchers suffer from spending significant efforts and time on the deliberate design of an academic poster. Therefore, there is urgent need for an automatic method for generating high-quality academic posters that can significantly relieve manual burdens.
\begin{figure*}[!t]
\centering
\includegraphics[width=0.95\textwidth]{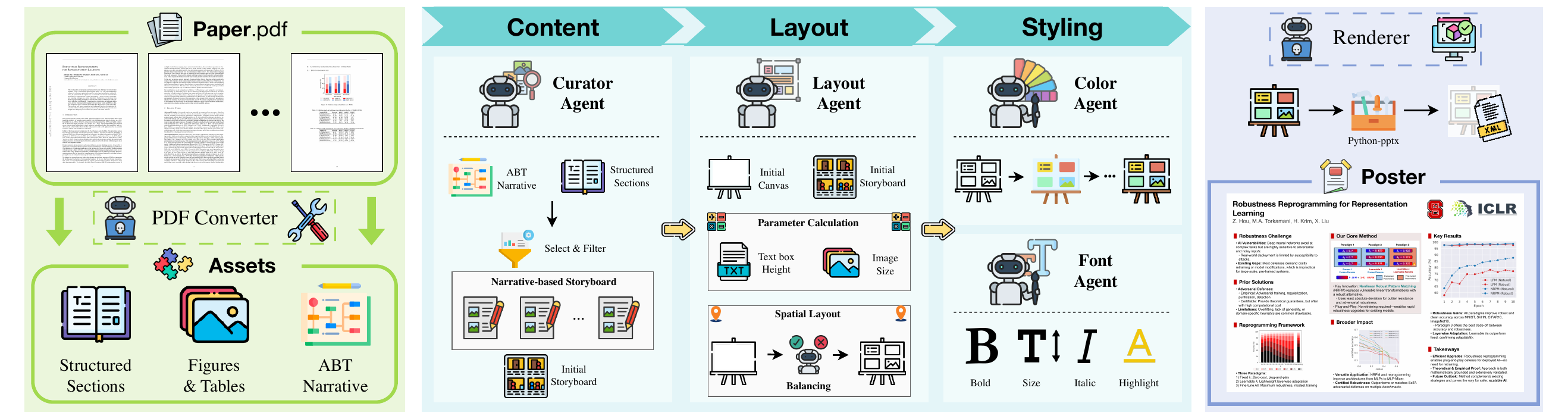}
\caption{\textbf{Overview of PosterGen.} The process consists of three main stages: (1) \textbf{ParserAgent} processes the input paper, extracting all text and visual assets and organizing them into a structured format focusing on an ABT narrative. (2) A series of aesthetic-aware agents then transform this content into a styled layout: \textbf{CuratorAgent} creates a narrative-based storyboard, \textbf{LayoutAgent} calculates the precise spatial arrangement and balances the columns, and \textbf{StylistAgents} apply a harmonious color palette and a hierarchical typographic system. (3) Finally, \textbf{Renderer} module takes the styled metadata and produces the output poster.}
\vspace{-3mm}
\label{fig::methods}
\end{figure*}

Although the automatic generation of commercial posters has received significant research attention~\citep{chen2025posta,chen2025postercraft,gao2025postermaker}, research on academic poster generation is far less explored. Early works on academic posters often used neural models~\citep{qiang2019learning,xu2022posterbot}. Other works focused on specific sub-tasks like layout generation~\citep{wang2024scipostlayout} or text summarization~\citep{saxena2025postersum,liu2022character}. These methods often produce posters with quality issues, such as content overflow~\citep{qiang2019learning}, and require further manual adjustment. Recently, LLM-powered multi-agent systems have shown strong performance on solving complex tasks. P2P~\citep{sun2026pp} and PosterAgent~\citep{pang2025paperposter} were the first to apply this approach to academic poster generation. However, their works do not sufficiently consider aesthetics and design principles, e.g., well-organized layout design that ensures natural reading flow, and styling choices for color and typography to present visual hierarchy.
\begin{table}[!t]
\caption{Comparison of existing paper-to-poster methods and their supported features. Here \cmark\;denotes fully supported, \tmark\;denotes  partially supported, and \xmark\;denotes not supported.}
\resizebox{\linewidth}{!}{
\begin{tabular}{lcccc}
\toprule
\multirow{2}{*}{Method} & Output Format & Outline & Layout & Styling \\
& / Editable Slides & Hierarchy & Alignment & Cohesion\\
\midrule
PosterBot~\cite{xu2022posterbot} & LaTeX /\;\tmark & \tmark & \tmark & \xmark \\
P2P~\cite{sun2026pp} & HTML /\;\xmark & \cmark & \cmark & \xmark  \\
PosterAgent~\cite{pang2025paperposter} & \textbf{PPTX} /\;\cmark & \cmark & \xmark & \xmark \\
\midrule
PosterGen (ours) & \textbf{PPTX} /\;\cmark & \cmark & \cmark & \cmark  \\
\bottomrule
\end{tabular}
}
\vspace{-3mm}
\label{tab:method_compare}
\end{table}

To move beyond the aesthetic limitations of current approaches and minimize the need for manual refinement, we propose the first aesthetic-driven multi-agent framework, \textbf{PosterGen}, designed to generate academic posters that are close to human-level aesthetic perfection. As shown in Table~\ref{tab:method_compare}, existing methods~\citep{xu2022posterbot, sun2026pp, pang2025paperposter} offer partial solutions, but PosterGen is the first to fully integrate both principled layout and stylings while supporting a more accessible \textbf{PPTX} format than LaTeX or HTML. We argue that embedding these aesthetic principles is indispensable for full automation. PosterGen implements a collaborative multi-agent workflow that mirrors a professional design process, and consists of four critical stages: (i) distilling the essential narrative and visual assets; (ii) constructing a narrative-driven storyboard based on the ABT structure; (iii) mapping the storyboard to a balanced and aesthetic-aware layout; and (iv) applying a cohesive system of colors and fonts.

We further evaluate PosterGen against state-of-the-art methods via VLM-as-Judge and human evaluation, which leads to several key findings: \ding{182} Text-to-image generation methods (GPT-4o-Image) frequently suffer from content hallucination and gibberish text. \ding{183} PosterGen achieves both content and aesthetic quality comparable to human-designed posters while consistently outperforming state-of-the-art multi-agent methods across almost all metrics. \ding{184} Quantitative and qualitative results confirm that our design-centric approach is highly effective, producing visually compelling and presentation-ready posters. 

Overall, our main contributions are as follows:
\begin{itemize}
    \item We propose \textbf{PosterGen}, a novel aesthetic-aware multi-modal multi-agent framework for academic poster generation that, to our knowledge, is the first to incorporate core design principles directly into its agent workflow.
    \item We introduce a comprehensive VLM-based evaluation rubric to assess the functional and aesthetic quality of generated posters, covering information, layout, color, and typography, and validated against human preferences.
    \item We present both quantitative and qualitative ablation studies that validate our fine-grained agent decomposition and the contribution of each specialized agent.
\end{itemize}

%% file: sec/2_rw.tex
\section{Related Work}
\label{sec:related_work}
\begin{figure*}[!t]
\centering
\includegraphics[width=0.95\textwidth]{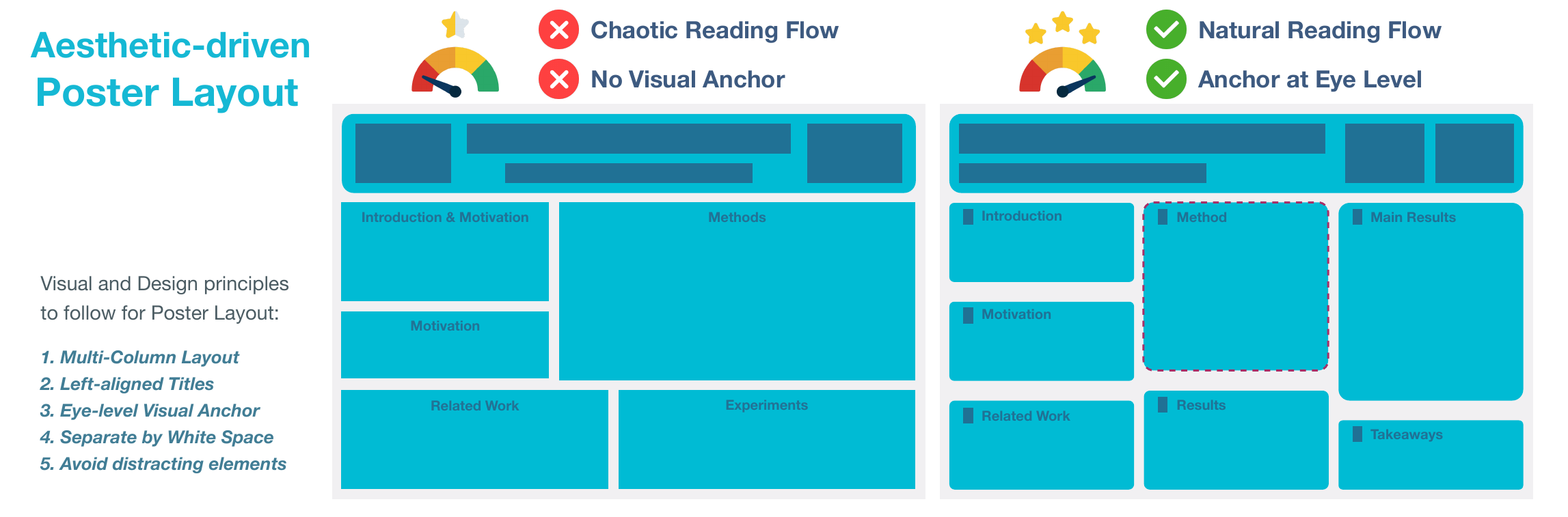}
\caption{A comparison of different layout structures. The vertically unaligned layout (left) results in a chaotic reading flow. In contrast, the vertically aligned grid (right), adopted by our LayoutAgent, establishes a natural reading flow and places the visual anchor at eye-level for emphasis.}
\vspace{-3mm}
\label{fig::layout}
\end{figure*}
\myparagraph{Poster Generation.}
Recent research has explored the automatic generation of artistic and product posters in a broad sense. For example, some works utilize modular~\citep{chen2025posta} or unified~\citep{chen2025postercraft} frameworks to achieve a high aesthetic quality in generated posters. Other methods focus on precise generation control, such as layout structure control using language models~\citep{seol2024posterllama}, text accuracy~\citep{gao2025postermaker}, or handling multiple user-provided conditions~\citep{zhang2025creatidesign}, all of which excel at creating visually appealing posters for art or marketing purposes. While these methods excel at visual appeal, an academic poster differs in its primary goal to convey complex research with precision and clarity in limited space. Early works used neural models to generate posters from papers~\citep{qiang2019learning,xu2022posterbot}. Recently, \citep{wang2024scipostlayout,saxena2025postersum} proposed benchmarks for this task; however, these methods suffer from several limitations, such as content overflow~\citep{qiang2019learning}, and restrictions to layout generation~\citep{wang2024scipostlayout} or text summary~\citep{saxena2025postersum} only.

Recent studies show that LLM-powered multi-agent frameworks can outperform single models on complex multimodal tasks~\citep{guo2024large, li2023camel,yin2023ttida,liu2024agentbench,jin2024contranovo, zhang2024autoregressive+,zhang2024cross,wu2024autogen,you2024calibrating,you2025uncovering,xiong2025quantagent,wei2025ai,pan2025beyond,zhao2025timeseriesscientist,sun2025docagent,liang2025slidegen,cao2025multi2,sun2026coma} by letting agents take on specialized roles and coordinate through mechanisms like self-reflection~\citep{bo2024reflective,wei2025retrieval}. P2P~\citep{sun2026pp} and PosterAgent~\citep{pang2025paperposter} were the first to apply this multi-agent solution to scientific poster generation. However, these methods lack a thorough consideration of design principles and aesthetics, and require extensive manual adjustments before they are ready for use in a conference poster session.

\myparagraph{Aesthetic Design with MLLMs.}
Recent works have explored multimodal large language models (MLLMs) for aesthetics-aware visual design. For instance, DesignProbe~\citep{lin2024designprobe} investigates the aesthetic reasoning capabilities of MLLMs through benchmarks that assess models in terms of color harmony, typography, and composition. Other works utilize MLLMs to generate aesthetics-constrained layouts, such as hierarchical layouts~\citep{cheng2025graphic}, generalized content-aware layouts~\citep{hsu2025postero} and aesthetic-aligned layout training schemes~\citep{patnaik2025aesthetiq}. Beyond layout prediction, POSTA~\citep{chen2025posta} integrates MLLMs with diffusion-based rendering to enable customizable artistic poster creation; PosterMaker~\citep{gao2025postermaker} introduces a high-quality rendering pipeline that improves the visual and linguistic fidelity for posters.

%% file: sec/3_design.tex
\section{Design Principles}\label{sec::design_principles}
Academic posters are visual media that require deliberate design to initiate conversations. Instead of relying on arbitrary heuristics, our framework is built upon a foundation of established research into high-quality poster design~\citep{faulkes2021better}. We distill these best practices into four core principles that are embedded directly into our agent designs.

\noindent \textbf{Narrative.}
A coherent narrative is the foundation of a design-aware poster. Following a schema widely adopted in scientific writing, we adopt the ``And, But, Therefore'' (\textbf{ABT}) structure~\citep{olson2019narrative} to distill the paper's core message, which establishes context (And), identifies problems (But), and presents solutions (Therefore). This narrative then guides the creation of specific, content-driven section titles.

\noindent \textbf{Layout Structure.}
Since a poster is a two-dimensional space with width and height, a three-column grid is a common and effective method to ensure a natural reading flow, as shown in Figure~\ref{fig::layout} (right). This structure strategically places a key visual anchor at the eye-level hot zone (top of the center column) and utilizes white space to separate elements and reduce visual clutter.

\noindent \textbf{Color Design.}
Color is used to create hierarchy and ensure accessibility. One optimal approach is to employ a restrained, theme-based, and monochromatic palette to maintain visual harmony (Figure~\ref{fig::color_palette}). This framework establishes a three-tier system: theme color for primary emphasis, monochromatic variants for section backgrounds, and a high-contrast accent color for highlights, with all text following the \underline{WCAG 4.5:1} contrast ratio to ensure readability.

\noindent \textbf{Typography Design.}
Typography complements color to reinforce clarity and readability from a standard viewing distance. We prioritize legible sans-serif typefaces and establish two hierarchy types: (1) a visual hierarchy using different font sizes (title, headings, body) to organize content structure, and (2) a semantic hierarchy using formatting like \textbf{bold}, \textit{italics}, or contrast color for emphasis.

%% file: sec/4_method.tex
\section{Method}\label{sec::method}
In this work, we propose a novel multi-agent framework for generating functionally effective and aesthetic-aware scientific posters, as shown in~Figure \ref{fig::methods}. Our framework implements the core design criteria from Section~\ref{sec::design_principles} by incorporating them as core logic within each specialized agent. This architecture establishes a cascade of structured design constraints throughout the entire generation process.

The PosterGen workflow consists of five specialist agents and one post-processing module: ParserAgent~(Section \ref{sec::parser}), CuratorAgent~(Section \ref{sec::curator_agent}), LayoutAgent~(Section \ref{sec::layout_agent}), ColorAgent and FontAgent~(Section \ref{sec::styling_agent}), and Renderer~(Section \ref{sec::renderer}).

\subsection{ParserAgent}\label{sec::parser}
Given a research paper in PDF format, ParserAgent initiates the workflow and is responsible for extracting the raw text and all available visual assets (e.g., figures and tables). To accomplish this, we utilize an external PDF converter tool, Marker~\citep{paruchuri2025marker}, which converts the paper's content into Markdown format and saves all visual assets as PNG images. 

To minimize token usage for downstream agents (particularly for lengthy papers), ParserAgent concurrently performs several processing functions. (a) It distills the paper's core narrative into the ABT structure (see Section~\ref{sec::design_principles}), to establish a guiding framework for all subsequent content organization; (b) it restructures the raw text into logical sections that focus on main content and essential details, under rigid limitation of maximum 1000 words per section; and (c) it classifies the extracted visual assets into distinct categories based on their narrative role: a single ``key\_visual'' representing the core research; visuals for ``problem\_illustration'' and ``method\_workflow''; figures depicting ``main\_results'' and ``comparative\_results''; and all other visual elements as ``supporting'' material. Prompts of ParserAgent can be found in~\cref{fig::prompt_parser_part1,fig::prompt_parser_part2,fig::prompt_parser_part3,fig::prompt_parser_part4} in the supplementary material.

\subsection{CuratorAgent}\label{sec::curator_agent}
CuratorAgent functions as a spatial narrative designer. Its primary design consideration is to orchestrate all parsed content elements tightly around the ABT narrative. This narrative-centric approach ensures that the poster's structure is fluid and engaging. By establishing a strong narrative foundation early, it also minimizes the need for unnecessary content and visual refinements in later stages.

Operating on the ABT structure and structured sections provided by ParserAgent, CuratorAgent performs the initial strategic placement of content, and maps the narrative onto a preliminary three-column storyboard. To follow the narrative and visual strategy, CuratorAgent enforces a strict limit of five to eight sections for the entire poster. This constraint guarantees that the three-column layout is fully utilized while preventing content overflow, which imitates a typical human design pattern that progresses logically from introduction and methods to results and discussion. Representative prompts of CuratorAgent can be found in~\cref{fig::prompt_curator_part1,fig::prompt_curator_part2,fig::prompt_curator_part3} in the supplementary material.

\subsection{LayoutAgent}\label{sec::layout_agent}
\begin{figure}[!t]
\centering
\includegraphics[width=\columnwidth]{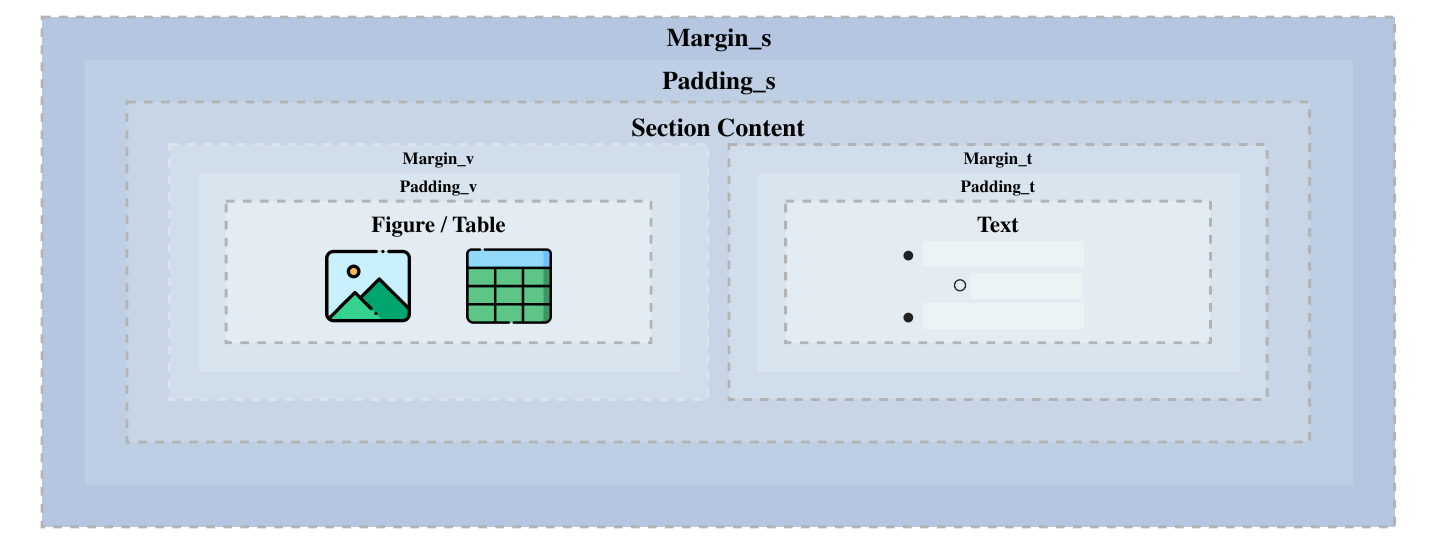}
\caption{An illustration of the CSS-like box model used to control the spacing between poster elements for XML files.}
\vspace{-3mm}
\label{fig::box_model}
\end{figure}

\begin{algorithm}[!t]
\caption{Optimal TextFrame Height Estimate}
\label{alg:text_height_estimation}
\KwIn{Text $T$, width $w$, font attr.\ $f$, precision $\varepsilon$}
\KwData{Initial bounds for binary search}
\KwOut{$h^\star$ (estimated height)}

$h_{\min}, h_{\max} \leftarrow \text{initial bounds}$\;

\While{$h_{\max} - h_{\min} > \varepsilon$}{
    $h_{\text{test}} \leftarrow (h_{\min} + h_{\max}) / 2$\;
    $B \leftarrow \textbf{SimulateTextbox}(T, w, h_{\text{test}}, f)$\;
    \eIf{\textbf{IsOverflowing}$(B)$}{
        $h_{\min} \leftarrow h_{\text{test}}$\;
    }{
        $h_{\max} \leftarrow h_{\text{test}}$\;
    }
    \textbf{Delete}$(B)$\;
}

$h^\star \leftarrow h_{\max} + \textbf{NewlineOffset}(T, f.\text{size})$\;
\Return{$h^\star$}\;
\end{algorithm}

LayoutAgent is a hybrid procedure that transforms the storyboard from CuratorAgent into a precise spatial layout. It operates in three phases: Phases 1 and 3 are procedural, computing the exact coordinates and dimensions $\{x, y, w, h\}$ for each element within a three-column grid (Figure~\ref{fig::layout}); Phase 2 invokes an LLM-based Balancer to optimize column utilization based on the spatial analysis from Phase 1. We use two key tools support this process: an element height estimation algorithm (Algorithm~\ref{alg:text_height_estimation}) for accurate vertical space allocation, and a CSS-like box model (Figure~\ref{fig::box_model}) for precise whitespace control in XML.

While calculating the height for visual assets is straightforward due to their fixed aspect ratios, determining the height for textFrames is much more complex for \texttt{PPTX}. This challenge derives from a discrepancy between the \texttt{python-pptx} library, which acts as an XML editor, and the final rendering engine (e.g., Microsoft PowerPoint) that determines the actual appearance. To bridge this gap, we propose the estimation algorithm detailed in Algorithm~\ref{alg:text_height_estimation}. The algorithm first employs a binary search to identify the minimum text box height that avoids any font size reduction by the rendering engine. It then applies a corrective offset, calculated from the number of newline characters, to compensate for subtle deviations in the engine's behavior.

To control whitespace precisely, we implement a CSS-like box model (Figure~\ref{fig::box_model}) that encapsulates every element with distinct `margin' and `padding' settings for fine-grained spacing control. This approach significantly narrows the layout capability gap that typically exists between automated HTML-based and PPTX-based layout generation methods. The prompts of LayoutAgent can be found in~\cref{fig::prompt_balancer_part1,fig::prompt_balancer_part2} in the supplementary material.

\subsection{StylistAgents}\label{sec::styling_agent}
Once the spatial layout is determined, StylistAgents apply the visual and typographic details to generate styled layouts. This stage consists of two specialized components: ColorAgent and FontAgent. Rather than the simple assignment of colors and fonts, we highlight the importance of a design thinking process rooted in the principles of poster aesthetics. This perspective is based on a core understanding that in academic posters, color and typography are not merely decorative; instead, they serve as essential media for both visual and semantic hierarchy.

\begin{figure}[!t]
\centering
\includegraphics[width=0.8\columnwidth]{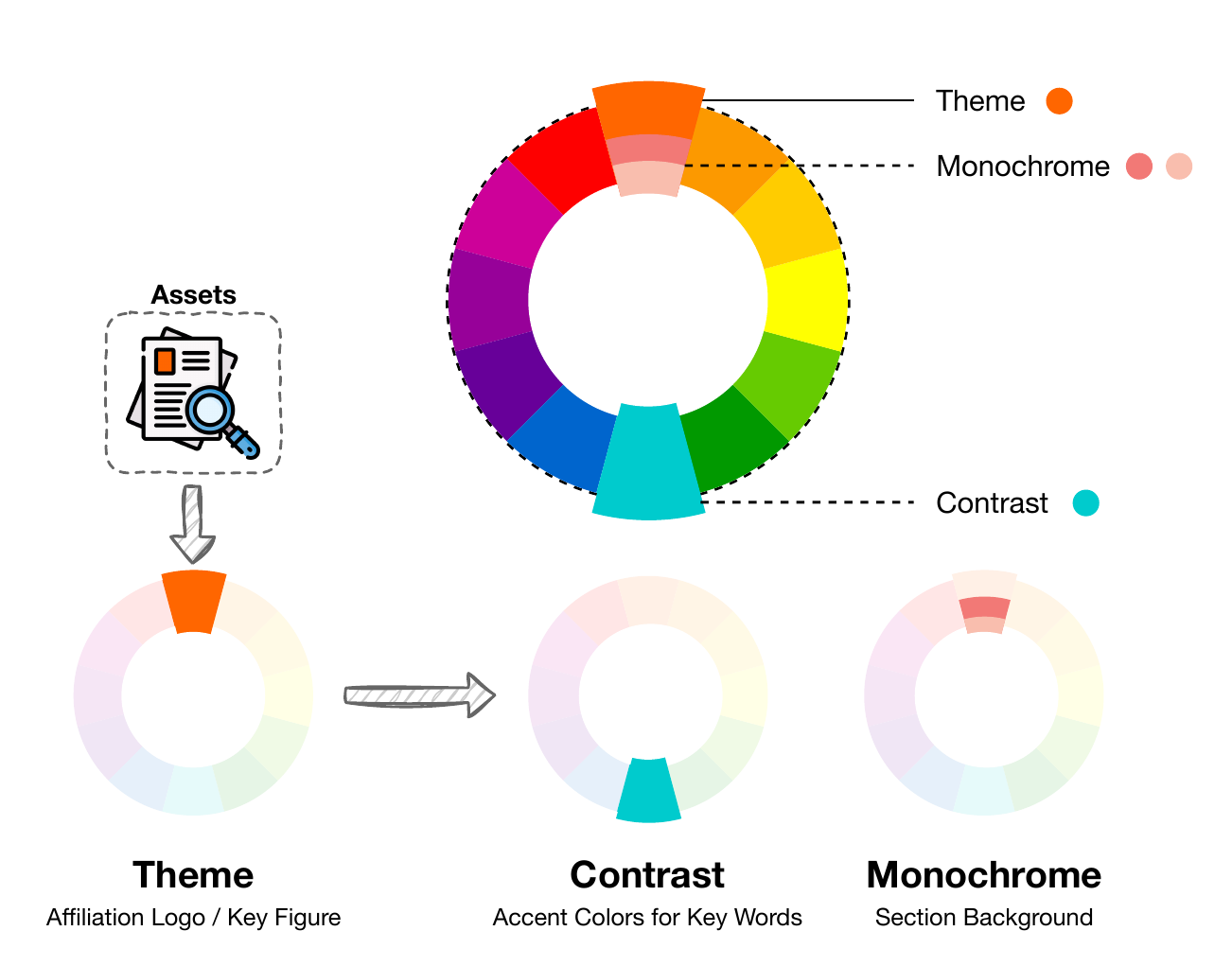}
\caption{Overview of the color palette generation. A primary color is extracted from a source image to get monochromatic and contrast colors.}
\vspace{-3mm}
\label{fig::color_palette}
\end{figure}

\myparagraph{ColorAgent.}
ColorAgent focuses on creating a suitable and harmonious color palette, as shown in Figure~\ref{fig::color_palette}. It first searches for the author's affiliation logo. If it exists, a VLM is adopted to analyze the image and extract a dominant theme color. This method leverages the institution's official branding to ensure an official appearance. For a fallback plan, ColorAgent can also analyze the key figure from the paper to identify a suitable theme color. After selecting the primary theme color, the next step for ColorAgent is to generate a complete color scheme strictly following color theory principles. For instance, given the theme color, ColorAgent will create the following color scheme:
\begin{itemize}
    \item monochromatic shades for backgrounds and accents, e.g., monochromatic light and dark;
    \item a high-contrast color that is used specifically for highlighting important keywords.
\end{itemize}

In this way, the ColorAgent generates a limited color palette that ensures aesthetic cohesion and high readability. Complete prompts of ColorAgent can be found in~\cref{fig::prompt_color} in the supplementary material.

\myparagraph{FontAgent.}
FontAgent manages typography and works to establish a clear visual hierarchy and emphasize key information within the text. It operates in a two-stage process: it first employs one LLM call to analyze the summarized text of the paper, which extracts a list of important keywords for each section. Next, FontAgent applies styling by using a set of predefined interfaces to assign different font families and sizes. FontAgent also highlights the keywords identified in the previous stage via the contrast color from ColorAgent. To avoid a tedious appearance, we adopt diverse highlighting styles, i.e., \textbf{bolding} and \textit{italics}, to make the poster more visually engaging. Complete prompts of FontAgent can be found in~\cref{fig::prompt_font} in the supplementary material.

\subsection{Renderer}\label{sec::renderer}
The renderer post-processes the fully styled layout metadata from the previous agents and renders a standard \texttt{PPTX} file using the \texttt{python-pptx} library. Additionally, it attaches affiliation and conference logos to the top-right corner of the poster. In the final step, the renderer uses \texttt{LibreOffice} (headless mode) to convert this \texttt{PPTX} file into a high-quality \texttt{PNG} image for visual inspection and evaluation.

%% file: sec/5_exp.tex
\definecolor{darkblue}{RGB}{123, 166, 180}
\definecolor{lightblue}{RGB}{173, 216, 230}
\definecolor{lightpurple}{RGB}{216, 191, 216}
\definecolor{darkpurple}{RGB}{166, 141, 166}

\section{Experiments}
\subsection{Metrics}
\begin{table}[!t]
\centering
\caption{VLM-as-Judge evaluation criteria for poster content and design, all on a 1-5 scale.}
\resizebox{0.9\linewidth}{!}{
\begin{tabular}{ccc}
\toprule
\textbf{Domain} & \textbf{Focus Area} & \textbf{Dimension} \\
\midrule
\multirow{3}{*}{\textbf{Content}} & \multirow{2}{*}{Information} & Layering, Coverage, \\
& & Depth, Completeness\\
\cmidrule(lr){2-3}
& Narrative & ABT, Flow, Conciseness \\
\midrule
\multirow{7}{*}{\textbf{Aesthetics}} & \multirow{2}{*}{Layout} & Grid, Whitespace, \\
& & Anchor, Balance\\
\cmidrule(lr){2-3}
& \multirow{2}{*}{Color} & Palette, Hierarchy, \\
& & Contrast, Source\\
\cmidrule(lr){2-3}
& \multirow{2}{*}{Typography} & Font, Emphasis, \\
& & Legibility\\
\cmidrule(lr){2-3}
& Strictness & Alignment, Robustness \\
\bottomrule
\end{tabular}
}
\vspace{-1mm}
\label{table:criteria}
\end{table}
\begin{table}[!t]
    \centering 
    \setlength{\tabcolsep}{2pt}
    \caption{Quantitative results on \textbf{content metrics} across different poster generation methods, with scores averaged over 30 posters rated on a 1-5 scale. The best scores for each content metric are \textbf{bolded}. The second best scores are \underline{underlined}.}
    \resizebox{1.0\linewidth}{!}{
    \begin{tabular}{llcccccccc}
        \toprule
        \multirow{2}{*}{\textbf{VLM}} & \multirow{2}{*}{\textbf{Method}} & \multicolumn{4}{c}{\textbf{Information}} & \multicolumn{3}{c}{\textbf{Narrative}} & \multirow{2}{*}{\textbf{Avg.}}\\
        \cmidrule(lr){3-6}
        \cmidrule(lr){7-9}
        & & Layer.$\uparrow$ & Cover.$\uparrow$ & Depth$\uparrow$ & Compl.$\uparrow$ & ABT$\uparrow$ & Flow$\uparrow$ & Conc.$\uparrow$ & \\
        \midrule
        \multirow{5}{*}{GPT-4.1} & \cellcolor{darkblue!25}Human-designed & \cellcolor{lightblue!25}\textbf{3.47} & \cellcolor{lightblue!25}\underline{3.83} & \cellcolor{lightblue!25}\textbf{3.97} & \cellcolor{lightblue!25}4.50 & \cellcolor{lightblue!25}4.53 & \cellcolor{lightblue!25}4.47 & \cellcolor{lightblue!25}4.47 & \cellcolor{darkblue!25}\underline{4.18}\\
        
        & \cellcolor{darkblue!25}GPT-4o-Image & \cellcolor{lightblue!25}2.23 & \cellcolor{lightblue!25}2.10 & \cellcolor{lightblue!25}2.20 & \cellcolor{lightblue!25}2.20 & \cellcolor{lightblue!25}2.73 & \cellcolor{lightblue!25}2.87 & \cellcolor{lightblue!25}2.57 & \cellcolor{darkblue!25}2.41\\
        
        & \cellcolor{darkblue!25}P2P~\citep{sun2026pp} & \cellcolor{lightblue!25}2.63 & \cellcolor{lightblue!25}3.33 & \cellcolor{lightblue!25}3.17 & \cellcolor{lightblue!25}3.83 & \cellcolor{lightblue!25}\underline{4.67} & \cellcolor{lightblue!25}\underline{4.57} & \cellcolor{lightblue!25}\underline{4.53} & \cellcolor{darkblue!25}3.82\\
        
        & \cellcolor{darkblue!25}PosterAgent~\citep{pang2025paperposter} & \cellcolor{lightblue!25}3.00 & \cellcolor{lightblue!25}3.33 & \cellcolor{lightblue!25}3.33 & \cellcolor{lightblue!25}\textbf{4.80} & \cellcolor{lightblue!25}4.07 & \cellcolor{lightblue!25}4.17 & \cellcolor{lightblue!25}4.07 & \cellcolor{darkblue!25}3.82 \\
        
        \cdashlinelr{2-10}
        
        & \cellcolor{darkblue!25}PosterGen (ours) & \cellcolor{lightblue!25}\underline{3.27} & \cellcolor{lightblue!25}\textbf{3.93} & \cellcolor{lightblue!25}\underline{3.83} & \cellcolor{lightblue!25}\underline{4.70} & \cellcolor{lightblue!25}\textbf{4.87} & \cellcolor{lightblue!25}\textbf{4.90} & \cellcolor{lightblue!25}\textbf{4.83} & \cellcolor{darkblue!25}\textbf{4.33} \\
        \midrule
        \multirow{5}{*}{\makecell[l]{Claude \\ Sonnet 4}} & \cellcolor{darkpurple!25}Human-designed & \cellcolor{lightpurple!25}\underline{3.73} & \cellcolor{lightpurple!25}\underline{3.33} & \cellcolor{lightpurple!25}\underline{4.13} & \cellcolor{lightpurple!25}4.47 & \cellcolor{lightpurple!25}\underline{4.40} & \cellcolor{lightpurple!25}4.27 & \cellcolor{lightpurple!25}3.73 & \cellcolor{darkpurple!25}\underline{4.01}\\
        
        & \cellcolor{darkpurple!25}GPT-4o-Image & \cellcolor{lightpurple!25}2.23 & \cellcolor{lightpurple!25}2.03 & \cellcolor{lightpurple!25}2.87 & \cellcolor{lightpurple!25}2.10 & \cellcolor{lightpurple!25}2.83 & \cellcolor{lightpurple!25}2.60 & \cellcolor{lightpurple!25}2.13 & \cellcolor{darkpurple!25}2.40\\

        & \cellcolor{darkpurple!25}P2P~\citep{sun2026pp} & \cellcolor{lightpurple!25}2.90 & \cellcolor{lightpurple!25}3.13 & \cellcolor{lightpurple!25}3.50 & \cellcolor{lightpurple!25}\underline{4.73} & \cellcolor{lightpurple!25}4.27 & \cellcolor{lightpurple!25}4.27 & \cellcolor{lightpurple!25}3.77 & \cellcolor{darkpurple!25}3.80 \\
        
        & \cellcolor{darkpurple!25}PosterAgent~\citep{pang2025paperposter} & \cellcolor{lightpurple!25}3.03 & \cellcolor{lightpurple!25}3.07 & \cellcolor{lightpurple!25}3.50 & \cellcolor{lightpurple!25}\textbf{4.80} & \cellcolor{lightpurple!25}4.23 & \cellcolor{lightpurple!25}\underline{4.30} & \cellcolor{lightpurple!25}\underline{3.90} & \cellcolor{darkpurple!25}3.83 \\
        
        \cdashlinelr{2-10}
        
        & \cellcolor{darkpurple!25}PosterGen (ours) & \cellcolor{lightpurple!25}\textbf{3.97} & \cellcolor{lightpurple!25}\textbf{3.43} & \cellcolor{lightpurple!25}\textbf{4.27} & \cellcolor{lightpurple!25}\textbf{4.80} & \cellcolor{lightpurple!25}\textbf{4.60} & \cellcolor{lightpurple!25}\textbf{4.87} & \cellcolor{lightpurple!25}\textbf{4.53} & \cellcolor{darkpurple!25}\textbf{4.35} \\
        \bottomrule
    \end{tabular}
    }
    \vspace{-3mm}
    \label{table:result_content}
\end{table}

\begin{table*}[!t]
    \centering 
    \setlength{\tabcolsep}{2.5pt}
    \caption{Quantitative results on \textbf{aesthetic metrics} across different poster generation methods, with scores averaged over 30 posters rated on a 1-5 scale. The best scores for each aesthetic metric are \textbf{bolded}. The second best scores are \underline{underlined}.}
    \resizebox{\linewidth}{!}{
    \begin{tabular}{llcccccccccccccc}
        \toprule
        \multirow{2}{*}{\textbf{VLM}} & \multirow{2}{*}{\textbf{Method}} & \multicolumn{4}{c}{\textbf{Layout}} & \multicolumn{4}{c}{\textbf{Color}} & \multicolumn{3}{c}{\textbf{Typography}} & \multicolumn{2}{c}{\textbf{Strictness}} & \multirow{2}{*}{\textbf{Avg.}}\\
        \cmidrule(lr){3-6}
        \cmidrule(lr){7-10}
        \cmidrule(lr){11-13}
        \cmidrule(lr){14-15}
        & & Grid$\uparrow$ & Wh.Sp.$\uparrow$ & Anch.$\uparrow$ & Bal.$\uparrow$ & P.l.t.$\uparrow$ & Hier.$\uparrow$ & Ctrst.$\uparrow$ & Src.$\uparrow$ & Font$\uparrow$ & Empha.$\uparrow$ & Legi.$\uparrow$ & Align.$\uparrow$ & Robust.$\uparrow$ & \\
        \midrule
        \multirow{5}{*}{GPT-4.1} & \cellcolor{darkblue!25}Human-designed & \cellcolor{lightblue!25}3.77 & \cellcolor{lightblue!25}3.80 & \cellcolor{lightblue!25}\underline{4.47} & \cellcolor{lightblue!25}3.83 & \cellcolor{lightblue!25}\textbf{3.83} & \cellcolor{lightblue!25}\textbf{3.93} & \cellcolor{lightblue!25}\underline{4.97} & \cellcolor{lightblue!25}\underline{4.10} & \cellcolor{lightblue!25}\underline{3.93} & \cellcolor{lightblue!25}\textbf{4.30} & \cellcolor{lightblue!25}4.70 & \cellcolor{lightblue!25}\underline{4.93} & \cellcolor{lightblue!25}\textbf{5.00} & \cellcolor{darkblue!25}\underline{4.27} \\
        
        & \cellcolor{darkblue!25}GPT-4o-Image & \cellcolor{lightblue!25}2.80 & \cellcolor{lightblue!25}3.07 & \cellcolor{lightblue!25}4.00 & \cellcolor{lightblue!25}2.80 & \cellcolor{lightblue!25}2.90 & \cellcolor{lightblue!25}2.73 & \cellcolor{lightblue!25}\textbf{5.00} & \cellcolor{lightblue!25}2.43 & \cellcolor{lightblue!25}3.37 & \cellcolor{lightblue!25}3.17 & \cellcolor{lightblue!25}4.13 & \cellcolor{lightblue!25}3.33 & \cellcolor{lightblue!25}\underline{1.13} & \cellcolor{darkblue!25}3.14 \\
        
        & \cellcolor{darkblue!25}P2P~\citep{sun2026pp} & \cellcolor{lightblue!25}3.80 & \cellcolor{lightblue!25}3.90 & \cellcolor{lightblue!25}4.20 & \cellcolor{lightblue!25}3.90 & \cellcolor{lightblue!25}\underline{3.53} & \cellcolor{lightblue!25}\underline{3.63} & \cellcolor{lightblue!25}\underline{4.97} & \cellcolor{lightblue!25}3.33 & \cellcolor{lightblue!25}3.13 & \cellcolor{lightblue!25}3.17 & \cellcolor{lightblue!25}4.37 & \cellcolor{lightblue!25}\textbf{5.00} & \cellcolor{lightblue!25}\textbf{5.00} & \cellcolor{darkblue!25}3.99\\
        
        & \cellcolor{darkblue!25}PosterAgent~\citep{pang2025paperposter} & \cellcolor{lightblue!25}\underline{3.97} & \cellcolor{lightblue!25}\underline{4.00} & \cellcolor{lightblue!25}4.13 & \cellcolor{lightblue!25}\underline{3.97} & \cellcolor{lightblue!25}3.20 & \cellcolor{lightblue!25}3.27 & \cellcolor{lightblue!25}\textbf{5.00} & \cellcolor{lightblue!25}3.17 & \cellcolor{lightblue!25}3.17 & \cellcolor{lightblue!25}3.00 & \cellcolor{lightblue!25}\underline{4.73} & \cellcolor{lightblue!25}\textbf{5.00} & \cellcolor{lightblue!25}\textbf{5.00} & \cellcolor{darkblue!25}3.97 \\
        \cdashlinelr{2-16}
        
        & \cellcolor{darkblue!25}PosterGen (ours) & \cellcolor{lightblue!25}\textbf{4.20} & \cellcolor{lightblue!25}\textbf{4.27} & \cellcolor{lightblue!25}\textbf{4.97} & \cellcolor{lightblue!25}\textbf{4.27} & \cellcolor{lightblue!25}3.50 & \cellcolor{lightblue!25}3.53 & \cellcolor{lightblue!25}\textbf{5.00} & \cellcolor{lightblue!25}\textbf{4.47} & \cellcolor{lightblue!25}\textbf{4.17} & \cellcolor{lightblue!25}\underline{4.23} & \cellcolor{lightblue!25}\textbf{5.00} & \cellcolor{lightblue!25}\textbf{5.00} & \cellcolor{lightblue!25}\textbf{5.00} & \cellcolor{darkblue!25}\textbf{4.43} \\
        \midrule
        \multirow{5}{*}{\makecell[l]{Claude \\ Sonnet 4}} & \cellcolor{darkpurple!25}Human-designed & \cellcolor{lightpurple!25}4.27 & \cellcolor{lightpurple!25}\underline{3.97} & \cellcolor{lightpurple!25}\underline{4.47} & \cellcolor{lightpurple!25}\underline{4.27} & \cellcolor{lightpurple!25}\textbf{4.17} & \cellcolor{lightpurple!25}\textbf{4.13} & \cellcolor{lightpurple!25}\textbf{5.00} & \cellcolor{lightpurple!25}\textbf{4.20} & \cellcolor{lightpurple!25}\textbf{3.30} & \cellcolor{lightpurple!25}\underline{4.77} & \cellcolor{lightpurple!25}\underline{3.87} & \cellcolor{lightpurple!25}4.47 & \cellcolor{lightpurple!25}\textbf{5.00} & \cellcolor{darkpurple!25}\textbf{4.30} \\
        
        & \cellcolor{darkpurple!25}GPT-4o-Image & \cellcolor{lightpurple!25}2.90 & \cellcolor{lightpurple!25}2.83 & \cellcolor{lightpurple!25}3.73 & \cellcolor{lightpurple!25}2.60 & \cellcolor{lightpurple!25}2.10 & \cellcolor{lightpurple!25}1.97 & \cellcolor{lightpurple!25}\underline{4.93} & \cellcolor{lightpurple!25}2.17 & \cellcolor{lightpurple!25}3.00 & \cellcolor{lightpurple!25}3.10 & \cellcolor{lightpurple!25}2.87 & \cellcolor{lightpurple!25}2.77 & \cellcolor{lightpurple!25}\underline{2.00} & \cellcolor{darkpurple!25}2.84 \\

        & \cellcolor{darkpurple!25}P2P~\citep{sun2026pp} & \cellcolor{lightpurple!25}4.30 & \cellcolor{lightpurple!25}3.90 & \cellcolor{lightpurple!25}4.37 & \cellcolor{lightpurple!25}4.10 & \cellcolor{lightpurple!25}\underline{3.37} & \cellcolor{lightpurple!25}\underline{3.60} & \cellcolor{lightpurple!25}\textbf{5.00} & \cellcolor{lightpurple!25}3.43 & \cellcolor{lightpurple!25}2.97 & \cellcolor{lightpurple!25}3.70 & \cellcolor{lightpurple!25}3.50 & \cellcolor{lightpurple!25}\textbf{4.87} & \cellcolor{lightpurple!25}\textbf{5.00} & \cellcolor{darkpurple!25}4.01 \\
        
        & \cellcolor{darkpurple!25}PosterAgent~\citep{pang2025paperposter} & \cellcolor{lightpurple!25}\textbf{4.50} & \cellcolor{lightpurple!25}3.93 & \cellcolor{lightpurple!25}4.03 & \cellcolor{lightpurple!25}\underline{4.27} & \cellcolor{lightpurple!25}2.90 & \cellcolor{lightpurple!25}3.10 & \cellcolor{lightpurple!25}\textbf{5.00} & \cellcolor{lightpurple!25}2.83 & \cellcolor{lightpurple!25}3.10 & \cellcolor{lightpurple!25}3.70 & \cellcolor{lightpurple!25}3.63 & \cellcolor{lightpurple!25}\underline{4.80} & \cellcolor{lightpurple!25}\textbf{5.00} & \cellcolor{darkpurple!25}3.91 \\

        \cdashlinelr{2-16}
        
        & \cellcolor{darkpurple!25}PosterGen (ours) & \cellcolor{lightpurple!25}\underline{4.37} & \cellcolor{lightpurple!25}\textbf{4.10} & \cellcolor{lightpurple!25}\textbf{4.87} & \cellcolor{lightpurple!25}\textbf{4.57} & \cellcolor{lightpurple!25}3.07 & \cellcolor{lightpurple!25}3.20 & \cellcolor{lightpurple!25}\textbf{5.00} & \cellcolor{lightpurple!25}\underline{4.10} & \cellcolor{lightpurple!25}\underline{3.27} & \cellcolor{lightpurple!25}\textbf{4.83} & \cellcolor{lightpurple!25}\textbf{4.00} & \cellcolor{lightpurple!25}4.70 & \cellcolor{lightpurple!25}\textbf{5.00} & \cellcolor{darkpurple!25}\underline{4.24} \\
        \bottomrule
    \end{tabular}
    }
    \vspace{-3mm}
    \label{table:result_design}
\end{table*}

We evaluate the generated posters using the comprehensive criteria detailed in Table~\ref{table:criteria}. To simulate how humans perceive and evaluate academic posters, we utilize Vision-Language Models (VLMs) as judges, and use both GPT-4.1 and Claude Sonnet 4 to mitigate assessment biases.

The evaluation is divided into two fundamental domains: \textbf{Content} and \textbf{Aesthetics}, which are detailed in Table~\ref{table:criteria}. The Content domain assesses the poster's fidelity to the source paper, focusing on two key areas: Information (evaluating the layering, coverage, depth, and completeness) and Narrative (evaluating the ABT structure, logical flow, and conciseness). The Aesthetic domain evaluates the poster's visual execution based on the principles from Section~\ref{sec::design_principles}. This is broken down into four focus areas: Layout (grid, whitespace, anchor, balance), Color (palette, hierarchy, contrast, source), and Typography (font, emphasis, legibility). Finally, the Strictness focus area assesses the technical quality, such as element Alignment and Robustness against critical flaws like content overlap and overflow.

\subsection{Baselines}
We choose two types of baselines: an end-to-end text-to-image generation method, i.e., GPT-4o-Image, and the state-of-the-art multi-agent poster generation methods.

\noindent\textbf{GPT-4o-Image Generation} is directly based on the ChatGPT web interface. We provide the GPT-4o model with the source PDF file, along with a text prompt that instructs it to generate an academic poster of a given size. This method produces the final poster as a single image in an end-to-end way, without explicit intermediate generation stages.

\noindent\textbf{P2P}~\citep{sun2026pp} is the first LLM-based multi-agent framework for academic poster generation. It uses three agents for visual element extraction, textual content generation, and final poster assembly, respectively. However, P2P renders its output exclusively in HTML and CSS, which is neither directly editable nor easily portable for practical use.

\noindent\textbf{PosterAgent}~\citep{pang2025paperposter} proposes a top-down, multi-agent pipeline that consists of (1) Parser to distill the source paper into a structured asset library; and (2) Planner agent that arranges assets into a binary-tree layout, which is subsequently refined by a (3) Painter-Commenter loop that leverages VLM feedback to address layout issues. Although this baseline provides a solid technical solution for poster generation in the \texttt{PPTX} format, it does not sufficiently incorporate aesthetic and design principles into its agent workflow, which marks a key difference from our approach.

\subsection{Quantitative Results and Comparisons}
We compare PosterGen with baseline approaches on 30 top-conference papers, which are detailed in \cref{AdditionalQualitativeResults} in the supplementary material. The results are presented in Table~\ref{table:result_content} and Table~\ref{table:result_design}. Overall, PosterGen achieves best or second-best performance across almost every metric, thus \textbf{significantly outperforming all baselines} and achieving a level of quality that is \textbf{comparable to} human-designed posters.
\begin{figure}[!t]
\centering
\includegraphics[width=0.95\columnwidth]{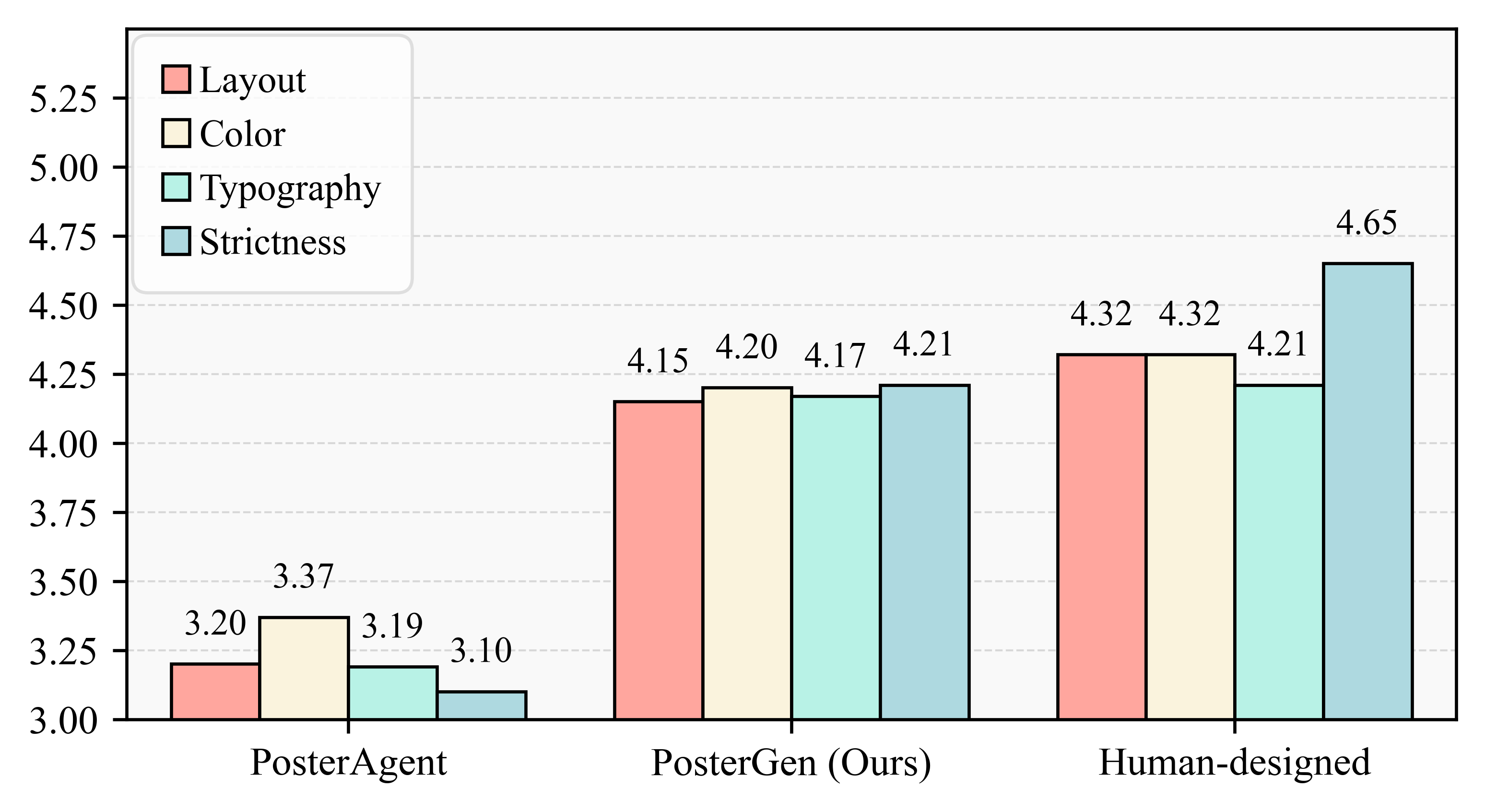}
\vspace{-8pt}
\caption{Human evaluation on aesthetic scores across 10 randomly picked sets of posters. Specific dimension scores are averaged to represent design aesthetics.}
\label{fig::vlm_humanstudy}
\end{figure}
\begin{figure*}[!t]
\centering
\includegraphics[width=0.98\textwidth]{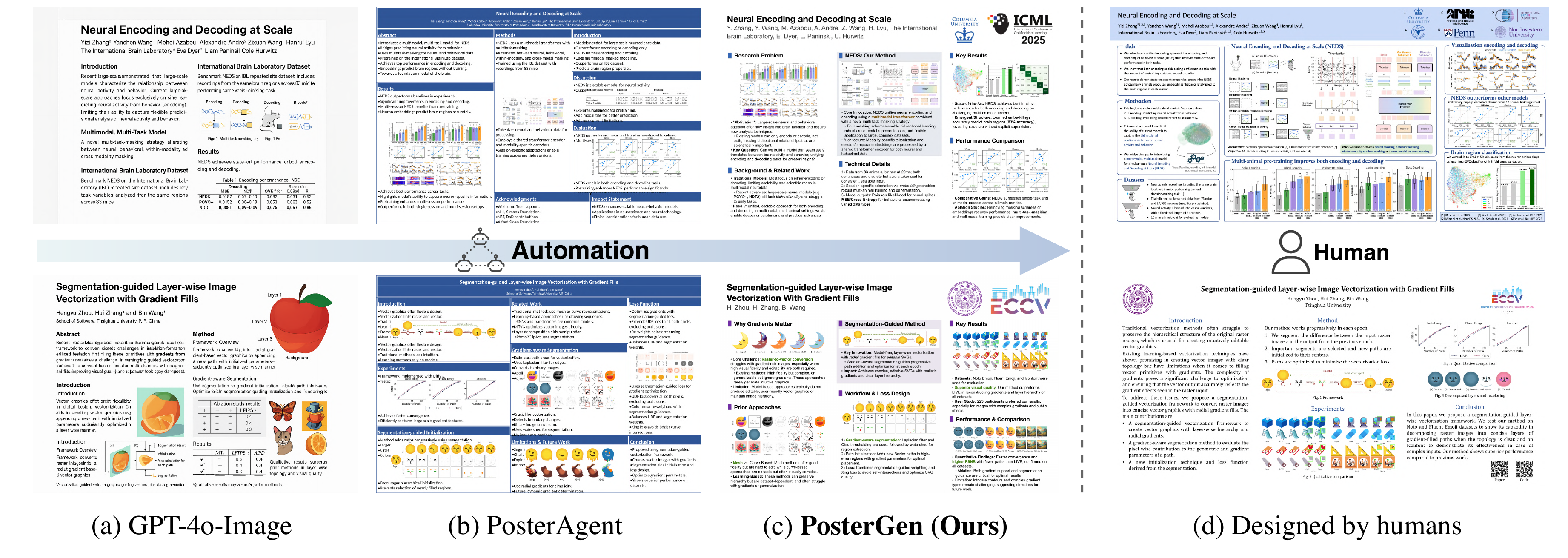}
\vspace{-3mm}
\caption{Qualitative comparison on two representative papers~\citep{zhang2025neural,zhou2024segmentation}. (a) Posters generated by GPT-4o-Image; (b) Posters generated by PosterAgent~\citep{pang2025paperposter}; (c) Posters generated by PosterGen (ours); (d) Posters designed by humans (Ground truth).}
\vspace{-3mm}
\label{fig::case_study}
\end{figure*}

\myparagraph{Content Performance.} As shown in Table~\ref{table:result_content}, PosterGen achieves the highest average content score across both VLM judges. With GPT-4.1, PosterGen attains an average score of \textbf{4.33}, surpassing both P2P~\cite{sun2026pp} and PosterAgent~\cite{pang2025paperposter} by \textbf{13.4\%}. The improvement is driven mainly by the superior narrative structure, as PosterGen achieves near-perfect scores in narrative metrics. This contrasts sharply with PosterAgent, which scores well on section completeness, but much lower across narrative metrics (\textbf{-15.7\%}). This highlights the efficacy of our \textbf{CuratorAgent}, which leverages the ABT narrative structure to create a coherent storyboard rather than just aggregating content. Meanwhile, the Claude Sonnet 4 judge gives highly similar results, which further reinforces PosterGen's informative capabilities.

\myparagraph{Aesthetic Performance.} As shown in Table~\ref{table:result_design}, PosterGen again achieves the highest average aesthetic score of \textbf{4.43} (per GPT-4.1), which is comparable to the Human-designed score (4.27), outperforming P2P by \textbf{11.0\%} and PosterAgent by \textbf{11.6\%}. With regard to layout design, PosterGen leads across all four related metrics, including a near-perfect 4.97 on visual anchor (``Anch."). Notably, although P2P benefits from HTML and CSS, which natively handle element alignment and responsive spacing, PosterGen surpasses it across all layout metrics under both VLM judges. This demonstrates the effectiveness of \textbf{LayoutAgent} in implementing core design principles, such as placing a visual anchor at eye-level, and using its CSS-like box model to effectively manage white space and visual balance.

While subjective metrics like color scheme (``P.l.t.") are competitive with human designers, our \textbf{StylistAgents} perform much better in systematic and aesthetic design. Per GPT-4.1, PosterGen achieves perfect legibility and the highest score for color source, validating its principled approach to typography and color. This confirms that while agent-based methods are feasible, PosterGen's design-centric agents are what truly deliver satisfactory aesthetic quality.

\subsection{Human Evaluation}
\begin{table*}[!t]
    \centering
    \caption{Quantitative results of ablation experiments over 10 posters. The best scores for each metric are \textbf{bolded}.}
    \vspace{-1mm}
    \resizebox{0.95\textwidth}{!}{
    \begin{tabular}{lccccccccccc}
        \toprule
        \multirow{2}{*}{\textbf{VLM Model}} & \multicolumn{3}{c}{\textbf{Agent Stage}} & \multicolumn{3}{c}{\textbf{Content Metrics}} & \multicolumn{5}{c}{\textbf{Aesthetic Metrics}} \\  
        \cmidrule(lr){2-4}
        \cmidrule(lr){5-7}
        \cmidrule(lr){8-12}
        & {\textbf{Curator}} & {\textbf{Layout}} & {\textbf{Styling}} & \textbf{Information}$\uparrow$ & \textbf{Narrative}$\uparrow$ & \textbf{Avg.}$\uparrow$ & \textbf{Layout}$\uparrow$ & \textbf{Color}$\uparrow$ & \textbf{Typography}$\uparrow$ & \textbf{Strictness}$\uparrow$ & \textbf{Avg.}$\uparrow$ \\
        
        \midrule
        
        \multirow{3}{*}{GPT-4.1} & \cmark & \xmark & \xmark & 3.70 & \textbf{4.90} & 4.30 & 3.68 & 3.10 & 3.17 & 4.35 & 3.57 \\
         & \cmark & \cmark & \xmark & 3.80 & 4.83 & 4.32 & \textbf{4.40} & 3.25 & 3.47 & \textbf{5.00} & 4.03 \\
         & \cmark & \cmark & \cmark & \textbf{3.93} & \textbf{4.90} & \textbf{4.41} & 4.28 & \textbf{4.33} & \textbf{4.50} & \textbf{5.00} & \textbf{4.53} \\
        \bottomrule
    \end{tabular}
    }
    \vspace{-3mm}
    \label{tab:ablation_2}
\end{table*}

Motivated by previous works~\citep{chen2025postercraft, xue2025comfybench}, we conduct a user study to establish correlation with human preferences. We randomly chose 10 sets of posters generated by PosterAgent, PosterGen, and humans to create 10 questions, and invited 30 researchers to evaluate these posters, rating each using a 1-5 Likert scale across four aesthetic dimensions, i.e., layout, color, typography, and technical quality. As shown in Figure~\ref{fig::vlm_humanstudy}, the results indicate that PosterGen significantly outperforms PosterAgent in all aesthetic scores. Moreover, PosterGen's scores (4.15 for layout, 4.20 for color, 4.17 for typography) tend to be very close to those of the human-made posters (4.32, 4.32 and 4.21, respectively). This further verifies the effectiveness of our aesthetic-centric framework. Additionally, it also shows that human evaluators are stricter in comparison to VLM judges, as PosterAgent's aesthetic scores in the human evaluation are around \textbf{15\%} lower than its VLM-as-Judge scores, suggesting that it potentially requires more manual refinement. More details of the user study can be found in Section~\ref{HumanEvaluationDetails} in the supplementary material.

\begin{figure*}[!t]
\centering
\includegraphics[width=0.95\textwidth]{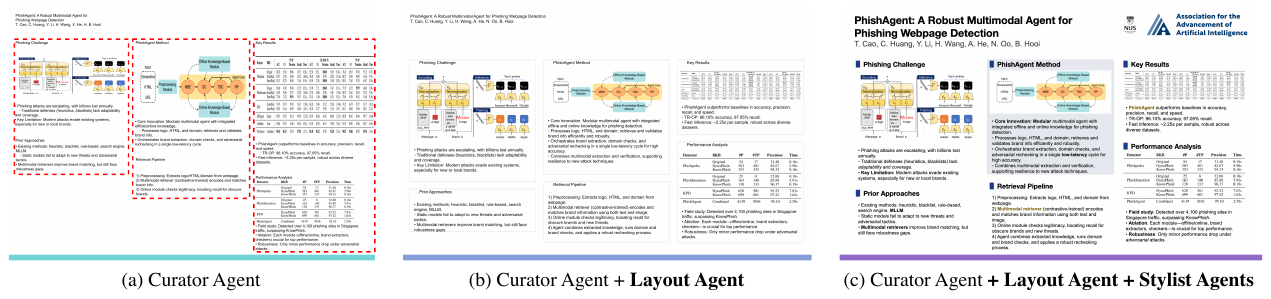}
\vspace{-3mm}
\caption{Qualitative results of ablation experiments on~\cite{cao2025phishagent}. (a) Output of CuratorAgent. Chaotic layouts are highlighted in the \textcolor{red}{red} dashed boxes. (b) Output of LayoutAgent. LayoutAgent applies spatial adjustments and balances space usage of columns. (c) Output of the entire multi-agent pipeline. StylistAgents apply visually appealing color and font elements to the poster.}
\vspace{-3mm}
\label{fig::ablation_study}
\end{figure*}

\subsection{Qualitative Results}\label{sec::qualitative}
To provide visually convincing demonstration, we conduct a qualitative comparison on two representative papers \citep{zhang2025neural,zhou2024segmentation}, as shown in Figure~\ref{fig::case_study}. A visual inspection shows that PosterGen significantly outperforms the baselines.

A major flaw in the end-to-end GPT-4o-Image method (\cref{fig::case_study} (a)) is the presence of critical content failures, despite a clear layout at first glance. Its outputs frequently suffer from sections of gibberish text, duplicated or broken content blocks, and the hallucination of visual assets not present in the source paper. In comparison, PosterAgent (\cref{fig::case_study} (b)) represents a significant improvement in content fidelity using multi-agent workflow. However, it remains limited due to lack of aesthetic consideration. The layout often suffers from misaligned sections or huge wasted space, and fails to establish a logical reading flow. Furthermore, its stylings merely rely on identically sized, black bullet points and monotonous plain text that cannot create a visual hierarchy or emphasize any key information.

In contrast, the posters generated by PosterGen (\cref{fig::case_study} (c)) exhibit a superior level of design quality that approaches the human-designed ones (\cref{fig::case_study} (d)). The region around the title bar is designed with elegance, applying varied fonts to the title, authors, and affiliation, along with their respective logos, which impose an instant hierarchy. Instead of applying lined borders, PosterGen adopts an easy-reading approach that utilizes colored section blocks and deliberate use of whitespace. The textual content is also enriched, as key phrases are highlighted using contrasting colors and varied formats to direct the viewer's attention. This is well demonstrated by the main anchor section, which applies a light monochromatic background to provide emphasis without imposing visual strains. More detailed qualitative results can be found in Section~\ref{AdditionalQualitativeResults} in the supplementary material.

\subsection{Ablation Study}
We conduct an ablation study on 10 randomly selected posters to validate the contribution of each agent, with results reported in Table~\ref{tab:ablation_2}. The full pipeline consistently achieves the highest performance across both content and aesthetic metrics.

CuratorAgent effectively establishes the poster content, as ``Information'' and ``Narrative'' metrics remain high with negligible variation across all stages, suggesting that the core content is successfully curated at this early stage. LayoutAgent then significantly improves the ``Layout'' score and technical quality via spatial arrangement. StylistAgents further enhance color and typography, underscoring their contribution to the final visual quality.

We also visualize the progressive output on a representative paper~\citep{cao2025phishagent} in Figure~\ref{fig::ablation_study}. CuratorAgent (a) produces the storyboard but with layout flaws such as imbalanced columns and content overflow, which are highlighted in the red dashed boxes. LayoutAgent (b) resolves these spatial issues through its box model and balancing loop, yielding a properly aligned three-column grid. StylistAgents (c) then apply aesthetic refinement by introducing themed colors to section titles and using a light background to emphasize the key method section, while highlighting keywords in contrasting colors and varied typography.

\subsection{Efficiency Analysis}
\begin{table}[!t]
    \centering
    \setlength{\tabcolsep}{2pt}
    \caption{Runtime and cost analysis averaged over 10 posters.}
    \resizebox{1.0\linewidth}{!}{
    \begin{tabular}{ccccccc}
        \toprule
        \multirow{2}{*}{\textbf{Runtime}} & \multirow{2}{*}{\textbf{API Calls}} & \multirow{2}{*}{\textbf{Cost (GPT-4.1)}} & \multicolumn{4}{c}{\textbf{Runtime Composition}} \\
        \cmidrule(lr){4-7}
        & & & \textbf{Parser} & \textbf{Curator} & \textbf{Layout} & \textbf{Other} \\
        \midrule
        5.32 min & 8 & \$0.20/poster & 62.73\% & 13.11\% & 17.89\% & 6.27\% \\
        \bottomrule
    \end{tabular}
    }
    \vspace{-3mm}
    \label{tab:runtime_cost}
\end{table}
We evaluate the runtime efficiency of PosterGen on a MacBook Air M3 without GPU acceleration or parallelism, over 10 randomly picked posters, as reported in Table~\ref{tab:runtime_cost}. PosterGen generates a complete poster in 5.32 minutes on average, and requires 8 API calls at a cost of \$0.20 per poster with GPT-4.1. ParserAgent accounts for the majority of runtime (62.73\%), as it relies on Marker for PDF parsing and visual asset extraction. The remaining agents collectively complete in under two minutes, demonstrating the practical efficiency of our multi-agent workflow.

%% file: sec/6_conclusion.tex
\section{Conclusion}
In this work, we present \textbf{PosterGen}, the first aesthetic-aware multi-agent framework capable of producing visually compelling posters approaching human-level quality. Our work is thoughtfully guided by fundamental design and aesthetic principles, incorporated into specialized agent collaboration that mirrors the process of professional designers. To systematically evaluate the visual quality, we introduce VLM-based criteria that measure information fidelity and aesthetic design, and conduct a user study to establish the correlation with human preferences. Experimental results demonstrate significant improvements in design aesthetics over state-of-the-art methods. Our method substantially alleviates the manual burden of preparing academic posters for researchers, and offers a principled and reproducible framework that narrows the aesthetic gap to human-designed posters, which lays a foundation for future design-aware multi-agent systems in scientific communication.

%% file: sec/X_suppl.tex
\clearpage
\setcounter{page}{1}
\maketitlesupplementary
\setcounter{section}{0}

\renewcommand{\thesection}{\Alph{section}}
\renewcommand{\thesubsection}{\thesection.\arabic{subsection}}
\renewcommand{\theHsection}{\Alph{section}}
\renewcommand{\theHsubsection}{\Alph{section}.\arabic{subsection}}

\section*{Table of Contents}
\startcontents[appendices]
\printcontents[appendices]{l}{1}{\setcounter{tocdepth}{3}}

\section{Extended Related Work}\label{ExtendedRelatedWork}
\input{sec/S7_ext_rw}

\section{Implementation Details}\label{ImplementationDetails}
\input{sec/S8_implement_details}

\section{Prompts}\label{Prompts}
\input{sec/S9_prompts}

\section{Human Evaluation Details}\label{HumanEvaluationDetails}
\input{sec/S10_human_eval_details}

\section{Additional Qualitative Results}\label{AdditionalQualitativeResults}

\input{sec/S11_additional_qualitative_results}

\section{Limitations}\label{limitations}
\input{sec/S12_limitations}

\section{Broader Impacts}\label{BroaderImpacts}
\input{sec/S13_broader_impacts}

%% file: sec/S7_ext_rw.tex
\paragraph{Slide Generation.}
A similar task to poster generation is the automatic generation of presentation slides from documents~\citep{mondal2024presentations, zheng2025pptagent, jung2025talk, fu2022doc2ppt, hu2014ppsgen, zhang2025tokenization, kumar2024slidespawn, sravanthi2009slidesgen, shi2025presentagent}. Some works develop agents for general purposes, such as efficient slide editing~\citep{jung2025talk} or narrated presentation videos~\citep{shi2025presentagent}. Other methods~\citep{zheng2025pptagent} focus on holistically improving the content, design, and coherence of the slides. Among these, several works specifically aim to generate slides for academic presentations. Early approaches like PPSGen~\citep{hu2014ppsgen}, SlidesGen~\citep{sravanthi2009slidesgen}, and SlideSpawn~\citep{kumar2024slidespawn} utilized summarization and information extraction techniques to generate draft slides. More recent approaches utilize end-to-end systems~\citep{fu2022doc2ppt} or persona-aware models~\citep{mondal2024presentations} to generate more tailored slides. However, academic poster design is far more challenging than slide generation, as slides can distribute content across multiple pages, and work together with a presenter's oral explanation to convey the full message of the paper, while an academic poster needs to contain all necessary information from a paper onto a single page, and ought to be visually appealing to attract the attention and help initiate a dialogue~\citep{faulkes2021better} with the authors.

%% file: sec/S8_implement_details.tex
\subsection{Agent Specifications}\label{AgentSpecifications}
Table~\ref{tab:agent_specs} details the type and role of each agent in our pipeline. Most agents (CuratorAgent, ColorAgent, FontAgent) operate via a single LLM call, while ParserAgent and LayoutAgent adopt a hybrid design that runs mostly procedural logic and only calls the LLM for specific subtasks. Renderer is entirely procedural and requires no LLM call.

\begin{table}[!t]
    \centering
    \setlength{\tabcolsep}{2pt}
    \caption{Agent specifications in PosterGen.}
    \resizebox{1.0\linewidth}{!}{
    \begin{tabular}{lccc}
        \toprule
        \textbf{Agent/Module} & \textbf{Type} & \textbf{Model} & \textbf{Role} \\
        \midrule
        Parser & Hybrid & GPT 4.1 & Structure section content \\
        $\drsh$Marker & Tool & N/A & Extract content from PDF \\
        Curator & LLM & GPT 4.1 & Create spatial content plan and storyboard \\
        Color & LLM & GPT 4.1 & Generate color palette \\
        \textbf{Layout} & \textbf{Hybrid} & N/A & Control precise element coords (CSS-style) \\
        \textbf{$\drsh$Balancer} & \textbf{LLM} & GPT 4.1 & Optimize column utilization\\
        Font & LLM & GPT 4.1 & Apply typography and keyword highlighting \\
        Renderer & Proced. & N/A & Generate final PPTX and PNG files \\
        \bottomrule
    \end{tabular}
    }
    \label{tab:agent_specs}
\end{table}

\subsection{Experimental Environment}\label{ExperimentalEnvironment}
\begin{table}[!t]
\centering
\caption{Key hyperparameters for PosterGen Implementations. The default LLM/VLM model is \texttt{gpt-4.1-2025-04-14}. The ``Content Constraints'' are controlled via instructions within the agent prompts. The symbol `\#' denotes ``number of.''}
\resizebox{\linewidth}{!}{
\begin{tabular}{llc}
\toprule
\textbf{Category} & \textbf{Parameter} & \textbf{Value} \\
\midrule
\multirow{5}{*}{\makecell{LLM/VLM \\ Configuration}} & Model & \texttt{GPT-4.1} \\
& \hspace{1em}$\drsh$~Alternatives & \texttt{GPT-4o}\\ 
&  & \texttt{GPT-4.1 mini}\\ 
&  & \texttt{Claude-Sonnet-4}\\ 
& Temperature & 0.7 \\
\midrule
\multirow{3}{*}{Content Constraints}& \# Sections & $[5, 8]$\\
& \# Visual Assets & $[4, 6]$\\
& Max Words per Section & 1000 \\
\midrule
\multirow{2}{*}{Text Height Estimation} & Height Precision ($\epsilon$) & 0.001 inches \\
& Newline Offset Ratio & 1.0 \\
\bottomrule
\end{tabular}
}
\vspace{-3mm}
\label{table:hyperparams}
\end{table}

The framework is compatible with Windows, Linux and macOS operating systems, and is implemented in Python 3.11. While a GPU is not strictly required, it is strongly recommended to accelerate document parsing and optical character recognition (OCR) tasks handled by the \texttt{marker-pdf} tool. The framework relies on several key libraries; the most critical dependencies for reproducing our work include:
\begin{itemize} 
    \item $\texttt{python-pptx} = 1.0.2$ for PPT generation. 
    \item $\texttt{langchain} = 0.3.25$ and $\texttt{langgraph} = 0.4.8$ for building the multi-agent workflow. 
    \item $\texttt{marker-pdf} = 1.8.2$ for PDF-to-Markdown conversion and parsing. Notably, we slightly modify the official \texttt{marker-pdf} source code to extract tables as \underline{images} rather than Markdown text. 
    \item $\texttt{Pillow} = 10.4.0$ for image manipulation. 
\end{itemize}

\subsection{Configuration Parameters}\label{ConfigurationParameters}
To ensure the reproducibility of our implementation and experiments, we list the most critical parameters that directly influence the behavior of the agents and the quality of generated posters in Table~\ref{table:hyperparams}. Aesthetic parameters related to layout, color and typography (font sizes, margins, color values) are defined in the configuration files within our source code but omitted here for clarity.

%% file: sec/S9_prompts.tex
\subsection{Baseline Prompt}\label{BaselinePrompt}
We present the prompt of GPT-4o Image Generation via ChatGPT web interface (as illustrated in Figure~\ref{fig::prompt_baseline}), which is alongside the input paper file. 
\begin{figure*}[!t]
\begin{tcolorbox}[colback=white!10!white, colframe=black!70!black, title=GPT-4o Image Generation]
You are an expert specializing in designing automated academic poster.

\textbf{Primary Task:} Analyze the provided research paper and autonomously design and generate a complete, professional academic poster.

\textbf{Key Guidelines:}
\begin{itemize}
    \item \textbf{Fixed Dimensions:} The final poster layout must be exactly \texttt{width} pixels wide and \texttt{height} pixels high.
    \item \textbf{Content Fidelity:} All content (text, figures, tables) must be extracted or summarized exclusively from the source paper. Do not invent or infer information.
    \item \textbf{Visual Design:} The poster must be visually appealing. Apply a clean and professional theme consistently across all elements.
    \item \textbf{Layout Design:} The layout must be well-balanced, effectively utilizing the available space. Actively avoid element overflow and large, underutilized blank areas.
\end{itemize}

\textbf{Required Structure $\&$ Content:}
\begin{itemize}
    \item \textbf{Header:} Must include the full paper title and a complete list of all authors.
    \item \textbf{Body:} The main content area must be organized into distinct sections. These sections should feature a well-balanced and carefully arranged composition of:
    \begin{itemize}
        \item Concise text summaries.
        \item high-quality figures from paper pdf.
        \item Key data tables from paper pdf.
    \end{itemize}
\end{itemize}

\textbf{Output Format}: 
The final output must be a single PNG image file with dimensions of \texttt{width} $\times$ \texttt{height} pixels.
\end{tcolorbox}
\vspace{-3mm}
\caption{Prompt for GPT-4o Image Generation.}
\label{fig::prompt_baseline}
\end{figure*}

\subsection{PosterGen Prompts}\label{PosterGenPrompts}
We present the detailed prompt design~\citep{zhang2025prompt} used in our PosterGen multi-agent workflow as follows.

\textbf{ParserAgent.} This includes: (1) Title and Authors Extraction~(Figure \ref{fig::prompt_parser_part1}); (2) Narrative (ABT) Extraction~(Figure \ref{fig::prompt_parser_part2}); (3) Visual Asset Classification~(Figure \ref{fig::prompt_parser_part3}); and (4) Structured Section Extraction~(Figure \ref{fig::prompt_parser_part4}).
\begin{figure*}[!t]
\begin{tcolorbox}[colback=white!10!white, colframe=black!70!black, title=ParserAgent (1): Title and Authors Extraction]
You are an expert academic paper parser. Your task is to extract the title and authors from the provided academic paper text.

Please extract:
\begin{enumerate}
    \item \textbf{Title:} The main title of the paper
    \item \textbf{Authors:} All author names using initials (no affiliations, emails, or other metadata)
\end{enumerate}

\textbf{Strict Formatting Requirements:}
\begin{itemize}
    \item \textbf{Title:} Use proper title case where each word has only the first letter capitalized, EXCEPT for established acronyms, technical terms, or proper nouns that are conventionally written in all uppercase letters (such as abbreviations for organizations, technologies, or methodologies). Keep such terms in their original case. Example: ``A Study of Machine Learning Methods'' but preserve acronyms like ``Using LLM for Data Analysis'' or ``CNN Architecture Design''.
    \item \textbf{Authors:} Use initials for authors' names. Convert full names to initials format, preserving middle initials when present. Examples: ``Kevin W. Jones" $\to$ "K.W. Jones'', ``Yann LeCun'' $\to$ "Y. LeCun", "Mary Smith Johnson" $\to$ "M.S. Johnson". Separate multiple authors with ``, ". Remove all affiliations, emails, institutions, departments, addresses, and other metadata.
\end{itemize}

\textbf{Input text:} \{\{ markdown document \}\}

\textbf{Required JSON structure:}
{\shorthandoff{"}%
\begin{lstlisting}
{
  "title": "Title With Proper Case Formatting",
  "authors": "F. Author, S. Author, T. Author"
}
\end{lstlisting}
}
\end{tcolorbox}
\vspace{-3mm}
\caption{Prompt for ParserAgent to extract the title and authors from the source paper.}
\label{fig::prompt_parser_part1}
\end{figure*}

\begin{figure*}[!t]
\begin{tcolorbox}[colback=white!10!white, colframe=black!70!black, title=ParserAgent (2): Narrative (ABT) Extraction]
Extract ABT narrative structure optimized for poster presentation:

\textbf{Input:} \{\{ Academic paper markdown text \}\}

\textbf{Output:} JSON with poster-ready ABT structure

\textbf{Guidelines:}
\begin{itemize}
    \item Each section (and/but/therefore) should be 1-2 concise sentences
    \item Focus on visual impact and poster audience understanding
    \item Emphasize key contributions and results
    \item Avoid technical jargon where possible
\end{itemize}

\textbf{Required JSON structure:}
\begin{lstlisting}
{
  "and": "Current knowledge and established facts (background context)",
  "but": "Specific problem, gap, or challenge identified", 
  "therefore": "Your solution, contribution, and key findings",
  "poster_hook": "One compelling sentence that grabs attention",
  "key_impact": "Why this research matters (practical implications)"
}
\end{lstlisting}

\textbf{Paper content:} {{ markdown$ $document }}

\end{tcolorbox}
\vspace{-3mm}
\caption{Prompt for ParserAgent to extract ABT-structured narratives.}
\label{fig::prompt_parser_part2}
\end{figure*}

\begin{figure*}[!t]
\begin{tcolorbox}[colback=white!10!white, colframe=black!70!black, title=ParserAgent (3): Classify Visual Assets]
Classify visual assets by column-aware poster placement and research role:

\textbf{Available visuals with captions:} {{ visuals list }}

Classify each visual into exactly one category based on column-specific poster design:

\textbf{Column-Aware Categories:}
\begin{enumerate}
    \item \textbf{key visual:} Most important method visual representing core research innovation (max 1, middle column)
    \item \textbf{problem illustration:} Visuals showing research problem, challenges, or motivation (left column introduction)
    \item \textbf{method workflow:} Method architecture, system diagrams, algorithmic workflows (middle column method)
    \item \textbf{main results:} Primary experimental results, performance tables, key findings (right column)
    \item \textbf{comparative results:} Baseline comparisons, ablation studies, validation charts (right column)
    \item \textbf{supporting:} Background concepts, supplementary analysis, minor details (flexible placement)
\end{enumerate}

\textbf{Classification Guidelines:}
\begin{itemize}
    \item Problem Context: Figures showing ``what's wrong" or ``why this matters" $\to$ problem illustration
    \item Method Core: Most important technical diagram $\to$ key visual
    \item Method Details: Architecture/workflow diagrams $\to$ method workflow
    \item Primary Evidence: Main performance results $\to$ main results
    \item Validation Evidence: Comparisons with baselines $\to$ comparative results
    \item Background/Supplementary: Minor or supporting content $\to$ supporting
\end{itemize}

\textbf{Consider:}
\begin{itemize}
    \item Visual content and research narrative role
    \item Optimal column placement for logical flow
    \item Visual impact and audience comprehension
\end{itemize}

\textbf{Required JSON output:}
\begin{lstlisting}
{
  "key_visual": "visual_id or null",
  "problem_illustration": ["visual_id1", ...],
  "method_workflow": ["visual_id1", ...],
  "main_results": ["visual_id1", ...],
  "comparative_results": ["visual_id1", ...],
  "supporting": ["visual_id1", ...]
}
\end{lstlisting}
Ensure every visual id appears exactly once across all categories.
\end{tcolorbox}
\vspace{-3mm}
\caption{Prompt for ParserAgent to classify extracted visual assets.}
\label{fig::prompt_parser_part3}
\end{figure*}

\begin{figure*}[!t]
\begin{tcolorbox}[colback=white!10!white, colframe=black!70!black, title=ParserAgent (4): Structured Section Extraction]
Extract structured sections from academic paper text for poster creation.

\textbf{Paper Text:} \{\{ raw text \}\}

\textbf{Task:}
Extract all major sections from the paper and organize them with their content. Focus on sections that would be relevant for an academic poster.

\textbf{Section Extraction Guidelines:}
\begin{enumerate}
    \item \textbf{Identify Major Sections:}
    \begin{itemize}
        \item Introduction/Background
        \item Related Work (if substantial)
        \item Methodology/Approach
        \item Experiments/Results
        \item Discussion/Analysis
    \end{itemize}
    \item \textbf{Content Processing:}
    \begin{itemize}
        \item Extract the main content for each section
        \item Keep section content under \textbf{1000} words
        \item Preserve key technical details, formulas, and findings
        \item Maintain important bullet points and lists
        \item Remove excessive citations and references
    \end{itemize}
    \item \textbf{Section Classification:}
        \begin{itemize}
        \item \textbf{foundation:} Introduction, background, motivation, problem statement
        \item \textbf{method:} Methodology, approach, algorithm, system design
        \item \textbf{evaluation:} Experiments, results, analysis, validation
    \end{itemize}
\end{enumerate}

\textbf{Required JSON structure:}
\begin{lstlisting}
{
  "paper_sections": [
    {
      "section_name": "Introduction",
      "section_type": "foundation",
      "content": "Main content of the section (max 1000 words)",
      "key_points": [ ... ],
      "importance": "high|medium|low",
      "contains_figures": ["figure_1", "figure_2"],
      "contains_tables": ["table_1"]
    }
  ],
  "paper_structure": {
    "total_sections": 5,
    "foundation_sections": 2,
    "method_sections": 2,
    "evaluation_sections": 1
  }
}
\end{lstlisting}

\textbf{Critical Requirements:}
\begin{itemize}
    \item Extract ALL major sections (don't skip any)
    \item Keep each section under 1000 words
    \item Preserve technical accuracy
    \item Identify which figures/tables belong to each section
    \item Classify section importance for poster layout
\end{itemize}

Generate structured sections that provide comprehensive paper coverage for poster creation.

\end{tcolorbox}
\vspace{-3mm}
\caption{Prompt for ParserAgent to extract structured sections.}
\label{fig::prompt_parser_part4}
\end{figure*}

\textbf{CuratorAgent.} CuratorAgent generates an effective storyboard through strategic content planning and by applying visual height constraints. Due to space limitations, we split the prompt into three parts: (1) Input and Design Patterns (Figure~\ref{fig::prompt_curator_part1}), (2) visual asset selection and content organization (Figure~\ref{fig::prompt_curator_part2}), and (3) output format (Figure~\ref{fig::prompt_curator_part3}). We also omit less important parts and replace them with ellipsis mark ($...$). The full prompt is available in the source code.
\begin{figure*}[!t]
\begin{tcolorbox}[colback=white!10!white, colframe=black!70!black, title=CuratorAgent: Spatial Story Board Generation (Part 1 of 3)]
You are an Expert Academic Poster Designer specializing in visual-dense poster layouts with strategic spatial organization.

\textbf{Mission:}
Transform research papers into spatially-organized poster sections that maximize visual asset utilization while following human design patterns. Prioritize visual impact over text density.

\textbf{Input:}
\begin{itemize}
    \item Paper Structured Sections: \{\{ structured sections \}\}
    \item Enhanced ABT Narrative: \{\{ narrative content \}\}
    \item Classified Visuals: \{\{ classified visuals \}\}
    \item Available Images: \{\{ available images \}\}
    \item Available Tables: \{\{ available tables \}\}
    \item Visual Heights Information: \{\{ visual heights info \}\}
    \item Available Height Per Column: \{\{ available height per column \}\}
\end{itemize}

\textbf{Human Poster Design Patterns:}\\
Based on analysis of successful academic posters:
\begin{enumerate}
    \item \textbf{Left Column Strategy - Foundation \& Context:}
    \begin{itemize}
        \item Introduction/Background/Motivation (priority placement)
        \item Problem definition and challenges
        \item Related work and background context
        \item Method overview or workflow diagrams
    \end{itemize}
    \item \textbf{Middle Column Strategy - Core Technical Content:}
    \begin{itemize}
        \item Primary methodology (highest priority content)
        \item Technical details and algorithms
        \item Theoretical analysis and key innovations
        \item System architecture diagrams
    \end{itemize}
    \item \textbf{Right Column Strategy - Experiments \& Results:}
    \begin{itemize}
        \item Experimental results (tables and performance charts)
        \item Key findings and validation data
        \item Performance comparisons and analysis
    \end{itemize}
\end{enumerate}
\end{tcolorbox}
\vspace{-3mm}
\caption{Part 1 of CuratorAgent prompt, focusing on the inputs, high-level instructions and human design patterns.}
\label{fig::prompt_curator_part1}
\end{figure*}

\begin{figure*}[!t]
\begin{tcolorbox}[colback=white!10!white, colframe=black!70!black, title=CuratorAgent: Spatial Story Board Generation (Part 2 of 3)]
\textbf{Oversized Visual Exclusion:}
\begin{itemize}
    \item \textbf{Exclusion Rule:} Any visual with height percentage $>50\%$ in visual heights info MUST BE EXCLUDED from poster
    \item \textbf{Reasoning:} Even with 80\% shrinking, these visuals would still exceed 40\% column height
    \item \textbf{Smart Substitution:} \dots
    \item \textbf{FALLBACK RULE:} If only ONE oversized visual ($>50\%$) is selected, allow it to proceed. For multiple oversized visuals, only select the one with SMALLEST height percentage.
\end{itemize}

\textbf{Visual Asset Strategic Selection Process}
\begin{enumerate}
    \item \textbf{Key Visual Mandatory Placement:}
    \begin{itemize}
    \item Identify the ``key visual'' from classified visuals. This is the MOST important visual
    \item Place key visual in middle column, top priority section
    \item This anchors the entire poster layout around the core research contribution
    \end{itemize}
    \item \textbf{Column-Based Visual Distribution:}
    \begin{itemize}
        \item \textbf{Column 1 (Left) - Foundation \& Context:}
        \begin{itemize}
            \item \textbf{MINIMUM:} 1 visual asset required
            \item \textbf{Purpose:} Express core research problem or contradiction visually
            \item \textbf{Selection Priority:} Choose visuals that illustrate problem context, background concepts, or prior work limitations
            \item \textbf{Maximum:} 2 visual assets
        \end{itemize}
        \item \textbf{Column 2 (Middle) - Methodology:}
        \begin{itemize}
            \item \textbf{MANDATORY:} Contains key visual from classified visuals
            \item \textbf{Additional:} May include 1 supporting method diagram
            \item \textbf{Maximum:} 2 visual assets
        \end{itemize}
        \item \textbf{Column 3 (Right) - Results \& Impact:}
        \begin{itemize}
            \item \textbf{STRICT MAXIMUM:} 2 visual assets ONLY
            \item \textbf{Selection Criteria:} Choose the 2 most critical visuals that directly validate main claims
            \item \textbf{Priority Order:} \dots
        \end{itemize}
    \end{itemize}
    \item \textbf{Visual Distribution Enforcement:} \dots
    \item \textbf{Column Space Optimization Strategy:} \dots
\end{enumerate}

\textbf{Core Task:}
Create 5-8 poster sections with BOTH content organization AND strategic spatial placement to achieve perfect space utilization across all three columns. DO NOT create any conclusion, takeaway, future work, or impact sections. Focus ONLY on problem, method, and results/experiments.

\textbf{Content Organization Guidelines:}
\begin{enumerate}
    \item \textbf{Section Requirements:}
    \begin{itemize}
        \item \textbf{Section titles}: Maximum 4 words (e.g., ``Our Method'', ``Key Results'')
        \item \textbf{Text content}: 2-3 concise entries using different rich hierarchical formatting (see examples below) based on section contents
        \item \textbf{Visual integration}: Each visual assigned to exactly ONE section
        \item \textbf{Complete content}: No ellipsis (\dots), write full bullet points
    \end{itemize}
    \item \textbf{Rich Text Formatting Options:}
    \begin{itemize}
        \item \textbf{A) Nested Bullet Structure:} 
\begin{lstlisting}
"* Primary concept or finding",
"   - Supporting detail or sub-point",
"   - Additional supporting evidence"
\end{lstlisting}
        \item Other formats like Bold Headers and Ordered Lists are also available.
    \end{itemize}
\end{enumerate}
\end{tcolorbox}
\vspace{-3mm}
\caption{Part 2 of CuratorAgent prompt, specifying the detailed rules for visual asset selection, content organization, and other planning requirements.}
\label{fig::prompt_curator_part2}
\end{figure*}

\begin{figure*}[!t]
\begin{tcolorbox}[colback=white!10!white, colframe=black!70!black, title=CuratorAgent: Spatial Story Board Generation (Part 3 of 3)]
\textbf{Output Format:}
\begin{lstlisting}
{
  "spatial_content_plan": {
    "poster_strategy": {
      "narrative_flow": "How the story progresses across columns",
      "space_utilization_approach": "Strategy for filling all three columns",
      "column_balance_rationale": "Why content is distributed this way"
    },
    "sections": [
      {
        "section_id": "unique_identifier",
        "section_title": "Max 4 Words",
        "column_assignment": "left|middle|right",
        "vertical_priority": "top|middle|bottom", 
        "importance_level": 1,
        "content_type": "foundation|method|results",
        "expected_content_density": "high|medium|low",
        "text_content": [
          "* **Key Innovation:** Core contribution with bold emphasis",
          "    - Supporting technical detail",
          "* **Impact:** Quantifiable result or benefit"
        ],
        "visual_assets": [
          {
            "visual_id": "figure_1",
            "visual_purpose": "How this supports the section",
            "placement_rationale": "Why this visual belongs in this spatial location"
          }
        ],
        "spatial_rationale": "Why this section belongs in this column/position"
      }
    ]
  },
  "column_distribution": {
    "left_column": {
      "focus": "Foundation and context",
      "assigned_sections": ["section_id_1", "section_id_2"],
      "content_strategy": "Build problem understanding and motivation"
    },
    "middle_column": {
      "focus": "Core methodology", 
      "assigned_sections": ["section_id_3", "section_id_4"],
      "content_strategy": "Present technical innovation and approach"
    },
    "right_column": {
      "focus": "Results and impact",
      "assigned_sections": ["section_id_5", "section_id_6"], 
      "content_strategy": "Demonstrate effectiveness and validations"
    }
  }
}
\end{lstlisting}
\end{tcolorbox}
\caption{Part 3 of CuratorAgent prompt, defining the exact JSON output format and data structure required from the agent.}
\label{fig::prompt_curator_part3}
\end{figure*}

\textbf{Layout Balancer.} This is a sub-agent of LayoutAgent designed to improve column utilization and prevent overflows. Its prompt is detailed in Figure~\ref{fig::prompt_balancer_part1} and Figure~\ref{fig::prompt_balancer_part2}.
\begin{figure*}[!t]
\begin{tcolorbox}[colback=white!10!white, colframe=black!70!black, title=Balancer Agent: Balance Column Space Utilization (Part 1 of 2)]
You are an expert academic poster layout optimization specialist. Your goal is to achieve optimal three-column space utilization through conservative within-column content adjustments only.

\textbf{Current Column Status:}
\begin{itemize}
    \item \textbf{Column 1 (Left):} \{left utilization\} utilization - \{left status\}
    \item \textbf{Column 2 (Middle):} \{middle utilization\} utilization - \{middle status\}  
    \item \textbf{Column 3 (Right):} \{right utilization\} utilization - \{right status\}
    \item \textbf{Available Height per Column:} \{available height\} inches
\end{itemize}

\textbf{Target Utilization:} 85-95\% for each column

\textbf{Core Optimization Principle:}
Prioritize content reduction over content expansion. Better to have 80\% utilization than risk overflow beyond available space.

\textbf{Column Content Rules:}
\begin{enumerate}
    \item Left Column: Foundation \& Context
    \begin{itemize}
        \item \textbf{Purpose:} Introduction, background, prior work, problem setup, supporting context
        \item \textbf{Content Types:} Motivation, challenges, related work, problem definitions, supporting materials
        \item \textbf{Reading Role:} Sets up the research problem and provides necessary background
    \end{itemize}
    \item Middle Column: Core Methodology
    \begin{itemize}
        \item \textbf{Purpose:} Method details, algorithms, implementation, technical innovation
        \item \textbf{Content Types:} Core methods, algorithms, technical approach, key innovations
        \item \textbf{Reading Role:} Presents the technical contribution and methodology
        \item \textbf{CRITICAL:} Contains key visual (importance level=1). NEVER remove method sections
    \end{itemize}
    \item Right Column: Results \& Impact
    \begin{itemize}
        \item \textbf{Purpose:} Experiments, evaluation, findings, conclusions, future work
        \item \textbf{Content Types:} Experimental results, performance analysis, conclusions, future directions
        \item \textbf{Reading Role:} Demonstrates validation and impact of the proposed method
    \end{itemize}
\end{enumerate}
\end{tcolorbox}
\caption{Part 1 of the Balancer sub-agent prompt, outlining its role, the current column status, and the fundamental content rules for each column.}
\label{fig::prompt_balancer_part1}
\end{figure*}

\begin{figure*}[!t]
\begin{tcolorbox}[colback=white!10!white, colframe=black!70!black, title=Balancer Agent: Balance Column Space Utilization (Part 2 of 2)]
\textbf{Within-Column Optimization Strategies:}
\begin{enumerate}
    \item Strategy A: Conservative Text Content Adjustment (for 80-100\% utilization)
    \begin{itemize}
        \item \textbf{When to use:} Column utilization is close to optimal range (80-100\%)
        \item \textbf{Actions allowed:} 
        \begin{itemize}
            \item MINIMAL text expansion: Add only 1-2 short phrases to underutilized columns (75-85\%)
            \item Aggressive text reduction: Significantly shorten content in overflow columns ($>$95\%)
            \item CONSERVATIVE APPROACH: Prefer slight underutilization over any risk of overflow
        \end{itemize}
        \item \textbf{Text Length Limits:} 
        \begin{itemize}
            \item Maximum per bullet: 25 words (count carefully)
            \item Maximum sub-bullets: 2 per main bullet
            \item Expansion limit: Add maximum 10-15 words total per section
            \item Reduction target: Remove 30-50\% of content from overflow sections
        \end{itemize}
    \end{itemize}
    \item Strategy B: Section Management (for $<$80\% or $>$100\% utilization)
    \begin{itemize}
        \item \textbf{When to use:} Column has severe underutilization ($<$80\%) or overflow ($>$100\%)
        \item \textbf{Actions allowed:} 
        \begin{itemize}
            \item Add sections from structured sections: Use additional content from paper sections that fit the column's purpose
            \item Remove less important sections: Remove sections with $\text{importance level}=3$ or lower importance
        \end{itemize}
        \item \textbf{Section Removal Priority:} 
        \begin{itemize}
            \item NEVER remove: Method sections with key visual ($\text{importance level}=1$)
            \item NEVER remove: Core experimental results or main findings
            \item Remove first: Supporting context, minor experiments, supplementary details ($\text{importance level}=3$)
            \item Remove second: Secondary analysis, additional background ($\text{importance level}=2$)
        \end{itemize}
    \end{itemize}
    
\end{enumerate}

\textbf{Strict Constraints:}
\begin{enumerate}
    \item \textbf{NO CROSS-COLUMN MOVES:} Never change column assignment for any existing section
    \item \textbf{PRESERVE READING FLOW:} Maintain left→middle→right logical progression 
    \item \textbf{SECTION ID PRESERVATION:} Never change section id, section title, visual assets, or other identifying fields
    \item \textbf{IMPORTANCE RESPECT:} Never remove critical sections (importance level=1 or core results)
    \item \textbf{TARGET UTILIZATION:} Achieve 85-95\% utilization for each column
\end{enumerate}

\textbf{Input:} \{\{structured sections\}, \{current story board\}, \{column analysis\}\}

\textbf{Output Format:}

Output the complete optimized story board JSON. Each section's `text content' must be an array of complete strings only:

\begin{lstlisting}
"text_content": [
  "* **Point Title:** Complete description text here",
  " - Supporting detail in complete sentences",
  "* **Another Point:** Full explanation without truncation"
]
\end{lstlisting}

Preserve all original structure and field names. Only modify content within string values.
\end{tcolorbox}
\caption{Part 2 of the Balancer sub-agent prompt, detailing the specific optimization strategies, strict constraints, and the required input/output format.}
\label{fig::prompt_balancer_part2}
\end{figure*}

\textbf{ColorAgent.} We present only the prompt for extracting the theme color from an affiliation logo using a VLM (as shown in Figure~\ref{fig::prompt_color}); the fallback method, which uses a key visual asset, is omitted for clarity.
\begin{figure*}[!t]
\begin{tcolorbox}[colback=white!10!white, colframe=black!70!black, title=ColorAgent: Theme Color Extraction]
Extract a sophisticated theme color from an affiliation logo that will work well as a poster accent color.

\textbf{Core Task:}
Analyze the provided affiliation logo and identify the most prominent, meaningful color that can serve as a poster theme color. This color should be:
\begin{itemize}
    \item Representative of the organization's visual identity
    \item Suitable for poster design applications (text highlights, accents)
    \item Professional and readable when used on white backgrounds
    \item Harmonious for academic poster contexts
\end{itemize}

\textbf{Color Extraction Guidelines:}
\begin{enumerate}
    \item \textbf{Primary Color Identification:}
    \begin{itemize}
        \item Look for the main brand color of the organization
        \item Ignore pure white, black, and very light grays (background/outline colors)
        \item Focus on colored elements that define the logo's visual identity
        \item Consider text colors, graphic elements, symbols, and emblematic elements
    \end{itemize}
    \item \textbf{Color Suitability Assessment:}
    \begin{itemize}
        \item Too Bright: If the main color is very bright/saturated (e.g., neon yellow \#FFFF00), generate a more subdued version
        \item Appropriate Saturation: Aim for colors that are vibrant but professional
        \item Readability: Ensure the color provides sufficient contrast on white backgrounds for text
    \end{itemize}
    \item \textbf{Color Adjustment Rules:}
    \begin{itemize}
        \item If original color is too bright (lightness $>$ 85\% or saturation $>$ 90\%), reduce brightness by 15-25\%
        \item If original color is too dark (lightness $>$ 25\%), lighten slightly for better visibility
        \item Maintain the color's hue character while optimizing for poster applications
    \end{itemize}
\end{enumerate}

\textbf{Output Requirements:}
Return ONLY a JSON object with the following structure:
\begin{lstlisting}
{
  "extracted_color": "#1E3A8A",
  "color_name": "Professional Navy Blue",
  "adjustment_made": "reduced_brightness | lightened | none",
  "original_color": "#0000FF",
  "suitability_score": 8.5,
  "reasoning": "Extracted the primary blue from the university emblem. Reduced brightness from bright blue to professional navy to ensure readability and sophisticated appearance on white backgrounds.",
  "usage_notes": "Excellent for text highlights, section headers, and accent elements. Provides strong contrast while maintaining professional appearance."
}
\end{lstlisting}

\textbf{Scoring Criteria (1-10 scale):}
\begin{itemize}
    \item \textbf{Contrast/Readability:} How well it works on white background
    \item \textbf{Professional Appearance:} Appropriate for academic/research contexts  
    \item \textbf{Brand Representation:} How well it represents the organization
    \item \textbf{Poster Suitability:} Effectiveness for highlights and accents
\end{itemize}

\end{tcolorbox}
\caption{Prompt for ColorAgent to extract theme color from affiliation logo.}
\label{fig::prompt_color}
\end{figure*}

\textbf{FontAgent.} FontAgent calls LLM once to extract and classify different keywords. The detailed prompt is illustrated in Figure~\ref{fig::prompt_font}.
\begin{figure*}[!t]
\begin{tcolorbox}[colback=white!10!white, colframe=black!70!black, title=FontAgent: Keyword Extraction]
Analyze poster content and identify keywords for strategic visual highlighting using three distinct formatting styles.

\textbf{Input Data:}
\begin{itemize}
    \item Enhanced Narrative: {{ enhanced narrative }}
    \item Curated Content: {{ curated content }}
\end{itemize}

\textbf{Core Task:}
For each section, identify keywords and assign them to specific highlighting styles based on their semantic importance and role in the research narrative.

\textbf{Highlighting Style Categories:}
\begin{enumerate}
    \item \textbf{BOLD + CONTRAST COLOR:}
    \begin{itemize}
        \item \textbf{Purpose:} Core method/methodology names that represent the paper's unique contribution
        \item \textbf{Criteria:} Novel algorithms, architectures, or techniques introduced by this work; the main methodological innovation that defines the paper; must be unique to this research (not generic terms)
        \item \textbf{Limit:} Maximum 2 per section, prefer 1 if it captures the main contribution
    \end{itemize}
    \item \textbf{BOLD:}
    \begin{itemize}
        \item \textbf{Purpose:} Important quantitative results and core technical terms within each section
        \item \textbf{Criteria:} Performance metrics and numerical results (e.g., ``\textbf{95\% accuracy}", ``\textbf{5.2$\times$ speedup}"); key technical concepts central to understanding the section; architecture names, dataset names, established method names; word-level emphasis, not entire phrases
        \item \textbf{Limit:} Maximum 3 per section
    \end{itemize}
    \item \textbf{ITALIC:}
    \begin{itemize}
        \item \textbf{Purpose:} Defining terms, single-word emphasis, and foreign terminology 
        \item \textbf{Criteria:} Technical terms being defined or introduced for the first time; single-word emphasis (e.g., ``This was the \textit{only} experiment"); foreign words, Latin terms, or specialized vocabulary; word-level application only, never entire sentences
        \item \textbf{Limit:} Maximum 2 per section
    \end{itemize}

\end{enumerate}

\textbf{Output Format:}
\begin{lstlisting}
{
  "section_keywords": {
    "motivation": {
      "bold_contrast": ["DP-CutMixSL"],
      "bold": ["Vision Transformers", "privacy leakage"],
      "italic": ["federated"]
    },
    "method": {
      "bold_contrast": ["CutMix", "differential privacy"],
      "bold": ["95% accuracy", "ResNet-50"],
      "italic": ["only"]
    },
    "results": {
      "bold_contrast": ["TransformerNet"],
      "bold": ["top-1 accuracy", "5.2x speedup", "CIFAR-10"],
      "italic": ["a priori"]
    }
  },
  "formatting_summary": {
    "total_bold_contrast": 4,
    "total_bold": 7,
    "total_italic": 3,
    ...
  }
}
\end{lstlisting}

Return a JSON object with the exact schema above that maximizes research impact through strategic visual emphasis.

\end{tcolorbox}
\caption{Prompt for FontAgent to extract different types of keywords.}
\label{fig::prompt_font}
\end{figure*}

\subsection{VLM-as-Judge Evaluation Prompt Template}\label{evaluation}
We present the prompt template used for our VLM-as-Judge evaluation, as shown in Figure~\ref{fig:prompt_eval}. This standardized template is applied to every evaluation focus area and uses a 5-point scale. To counteract the tendency of Vision-Language Models (VLMs) to provide overly generous scores, our prompt design incorporates a targeted few-shot example strategy. For high scores (4 and 5), we provide positive examples of desired qualities, while for low-to-mid scores (1, 2, and 3), we provide negative examples of common flaws. This approach is designed to calibrate the VLM's judgment and yield more accurate, evidence-based scoring.

\begin{figure*}[!t]
\begin{tcolorbox}[colback=white!10!white, colframe=black!70!black, title=VLM-as-Judge Evaluation Template]
You are an expert academic poster design critic. Your evaluation must be strict, detailed, and evidence-based. For each dimension, assess the poster against the specific examples provided in the 5-point scale. \textbf{High scores require adherence to professional design principles (positive examples). Low scores should be assigned when common design failures (negative examples) are present.} A poster that is merely functional but exhibits poor design choices must be scored significantly lower on design metrics than one that is both functional and visually excellent.

Please carefully examine the provided poster image and evaluate it across the following metrics. Provide a detailed explanation and score on a 5-point scale.

\textbf{FOCUS AREA: $<\texttt{Focus Area}>$}

\textbf{Metric : $<\texttt{Metric Description}>$}
\begin{itemize}
    \item \textbf{Score 5 (Excellent):}
    \begin{itemize}
        \item \textbf{Descriptor:} \dots
        \item \textbf{Positive Examples:} \dots
    \end{itemize}
    \item \textbf{Score 4 (Good):}
    \begin{itemize}
        \item \textbf{Descriptor:} \dots
        \item \textbf{Positive Examples:} \dots
    \end{itemize}
    \item \textbf{Score 3 (Acceptable):}
    \begin{itemize}
        \item \textbf{Descriptor:} \dots
        \item \textbf{Negative Examples:} \dots
    \end{itemize}
    \item \textbf{Score 2 (Poor):}
    \begin{itemize}
        \item \textbf{Descriptor:} \dots
        \item \textbf{Negative Examples:} \dots
    \end{itemize}
    \item \textbf{Score 1 (Failed):}
    \begin{itemize}
        \item \textbf{Descriptor:} \dots
        \item \textbf{Negative Examples:} \dots
    \end{itemize}
\end{itemize}

\textbf{Output Format:}
Please provide your evaluation as a JSON array with exactly 1 object for the metric:
\begin{lstlisting}
[
  {
    "metric": "Core Graphic Principles",
    "explanation": "Detailed analysis of the application of repetition, alignment, contrast, and proximity principles...",
    "score": 4
  }
]
\end{lstlisting}

\textbf{Evaluation Instructions:}
\dots

Please evaluate the poster step by step according to these criteria.
\end{tcolorbox}
\caption{Prompt Template for VLM-as-Judge Evaluation.}
\label{fig:prompt_eval}
\end{figure*}

%% file: sec/S10_human_eval_details.tex
\begin{figure*}[t]
\centering
\includegraphics[width=0.95\textwidth]{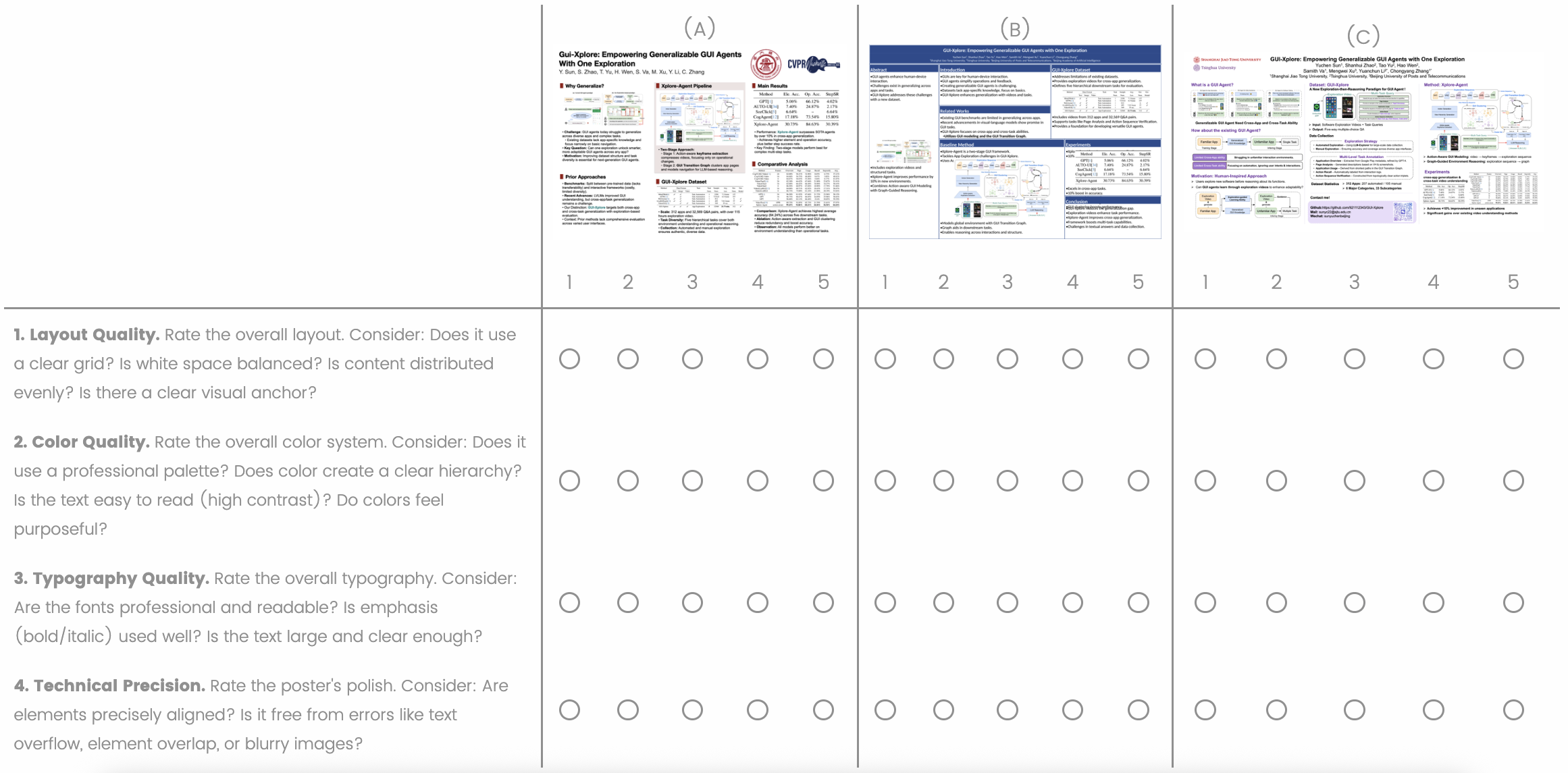}
\caption{A sample set of questions from Qualtrics survey for human evaluation.}
\label{fig::humanstudy_example}
\end{figure*}
\begin{figure*}[t]
\centering
\includegraphics[width=0.95\textwidth]{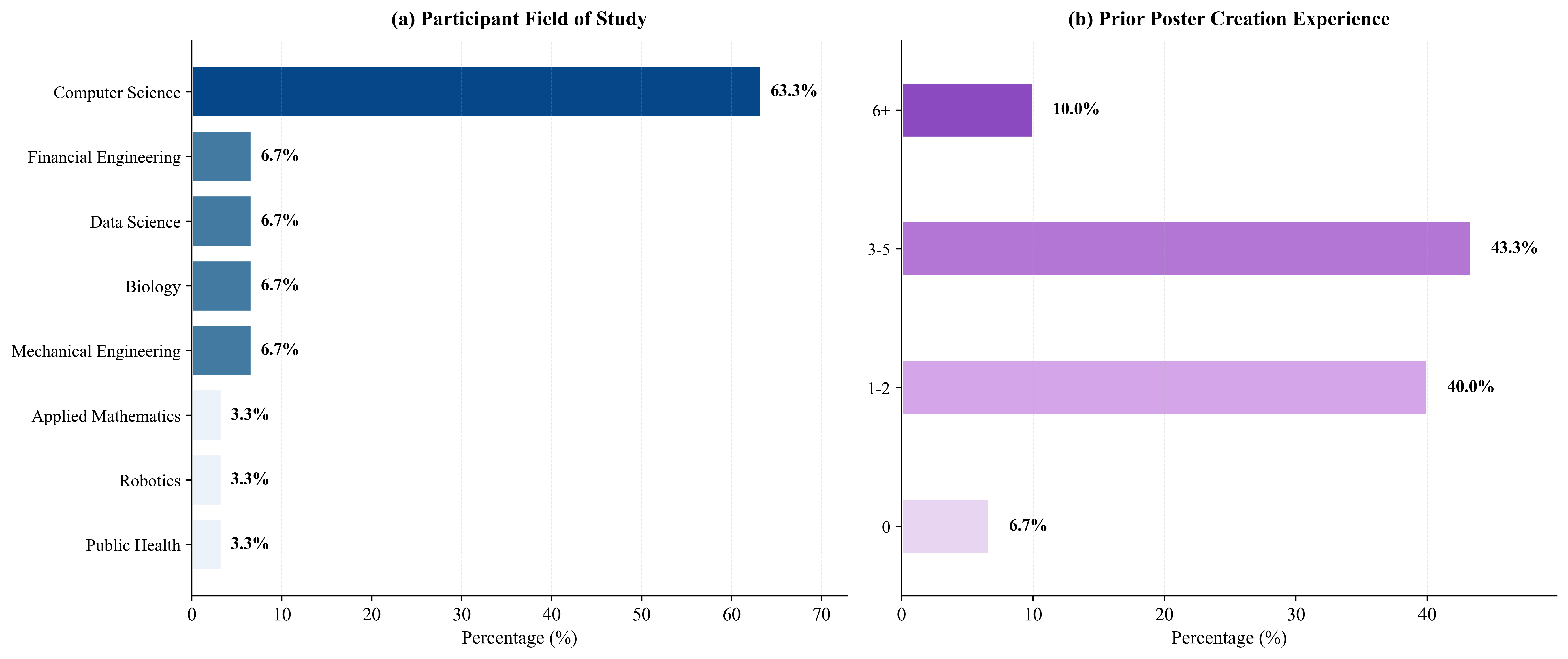}
\caption{Participant demographics regarding field of study and prior poster creation experience.}
\label{fig::humanstudy_bg}
\end{figure*}

We invite 30 participants to take part in our user study using the Qualtrics platform. To gain objective evaluation results, we survey each user for information regarding their field of study and experience in making academic posters, as shown in~\cref{fig::humanstudy_bg}. This verifies that our group of human evaluators possesses adequate research background to evaluate academic posters properly.

To alleviate evaluator fatigue and cognitive load, we did not ask each participant to assess all of our benchmark. Instead, each participant only needs to evaluate aesthetic scores for a random subset of 10 posters. Each set consists of three variations of posters: generated by our approach PosterGen, the baseline PosterAgent~\citep{pang2025paperposter}, and designed by authors of the paper. A rigorous randomization procedure is adopted to reduce possible bias: (i) each of the 10 sets is randomly ordered for evaluation, and (ii) for each set, the order of the three posters is also randomized. This makes sure that the evaluation is completely blind to all of our approaches for its validity to be maximized. An example set of questions is shown in \cref{fig::humanstudy_example}.

In addition, we did not include the GPT-4o-Image approach in our human study. As discussed in the qualitative analysis in Section 5.5 of the main paper, diffusion-based generation via GPT-4o-Image constantly suffers from gibberish text, distorted text blocks, and the hallucination of visual assets. It is thus pointless to consider it for comparison with the multi-agent approach and human designs.

%% file: sec/S11_additional_qualitative_results.tex
\begin{figure*}[!t]
\centering
\includegraphics[width=0.75\textwidth]{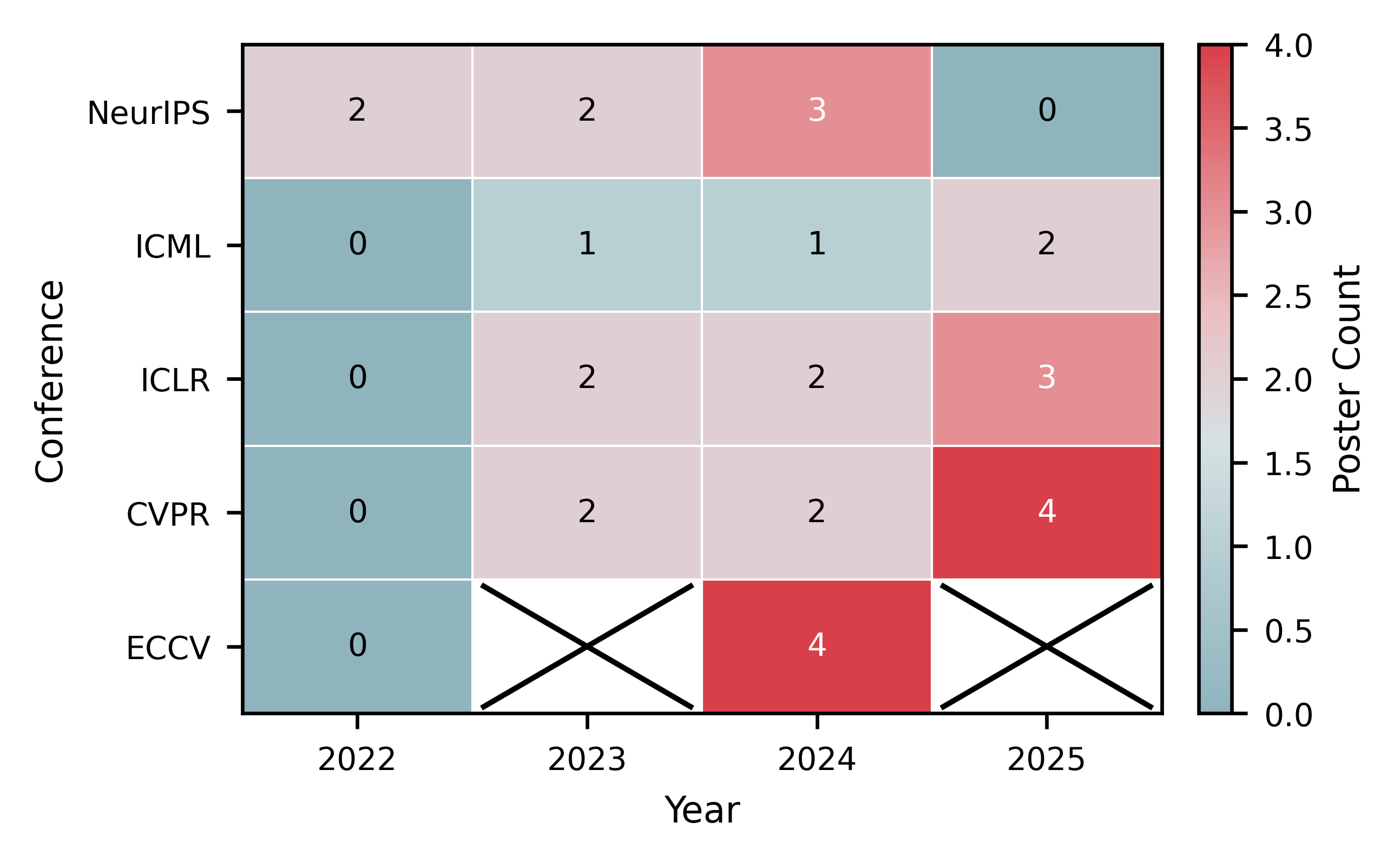}
\vspace{-8pt}
\caption{Data distribution of our evaluation benchmark from top-tier AI conferences.}
\label{fig::dataset}
\end{figure*}
\begin{figure*}[!htp]
\centering
\includegraphics[width=0.8\textwidth]{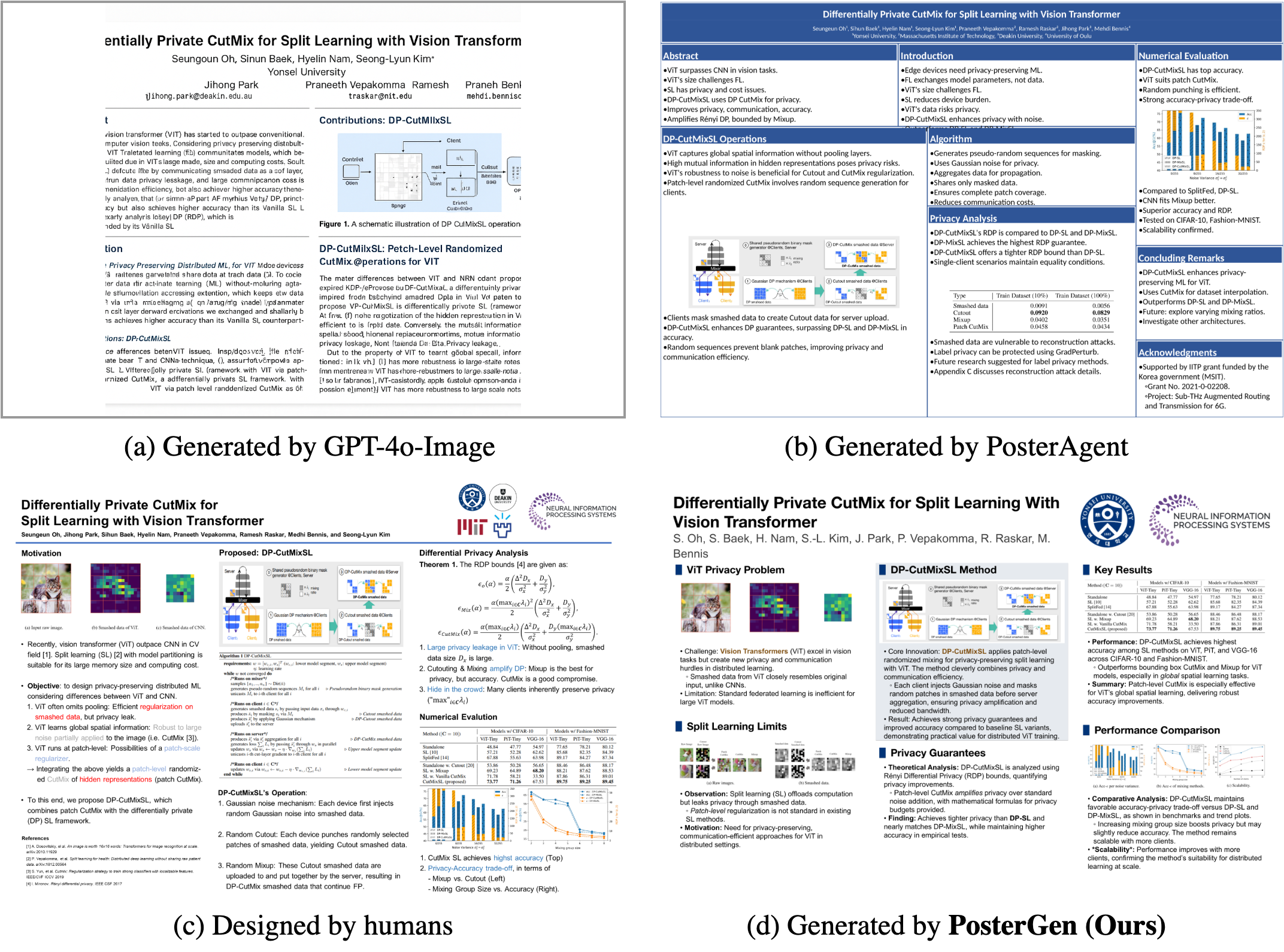}
\caption{Qualitative results of~\citet{oh2022differentially}.}
\label{fig::case_study_1}
\end{figure*}
\begin{figure*}[!htp]
\centering
\includegraphics[width=0.8\textwidth]{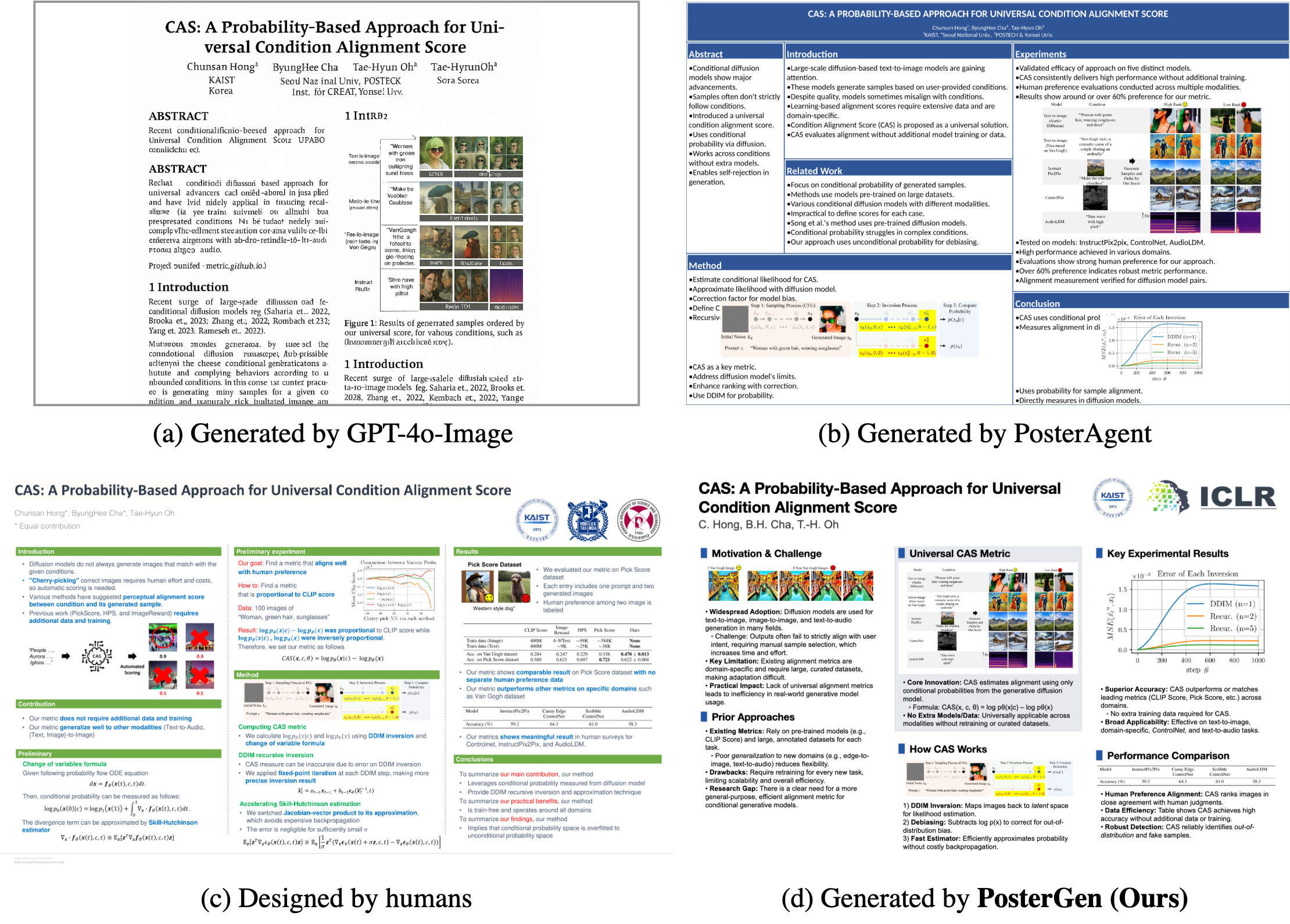}
\caption{Qualitative results of~\citet{hong2024cas}.}
\label{fig::case_study_2}
\end{figure*}
\begin{figure*}[!htp]
\centering
\includegraphics[width=0.8\textwidth]{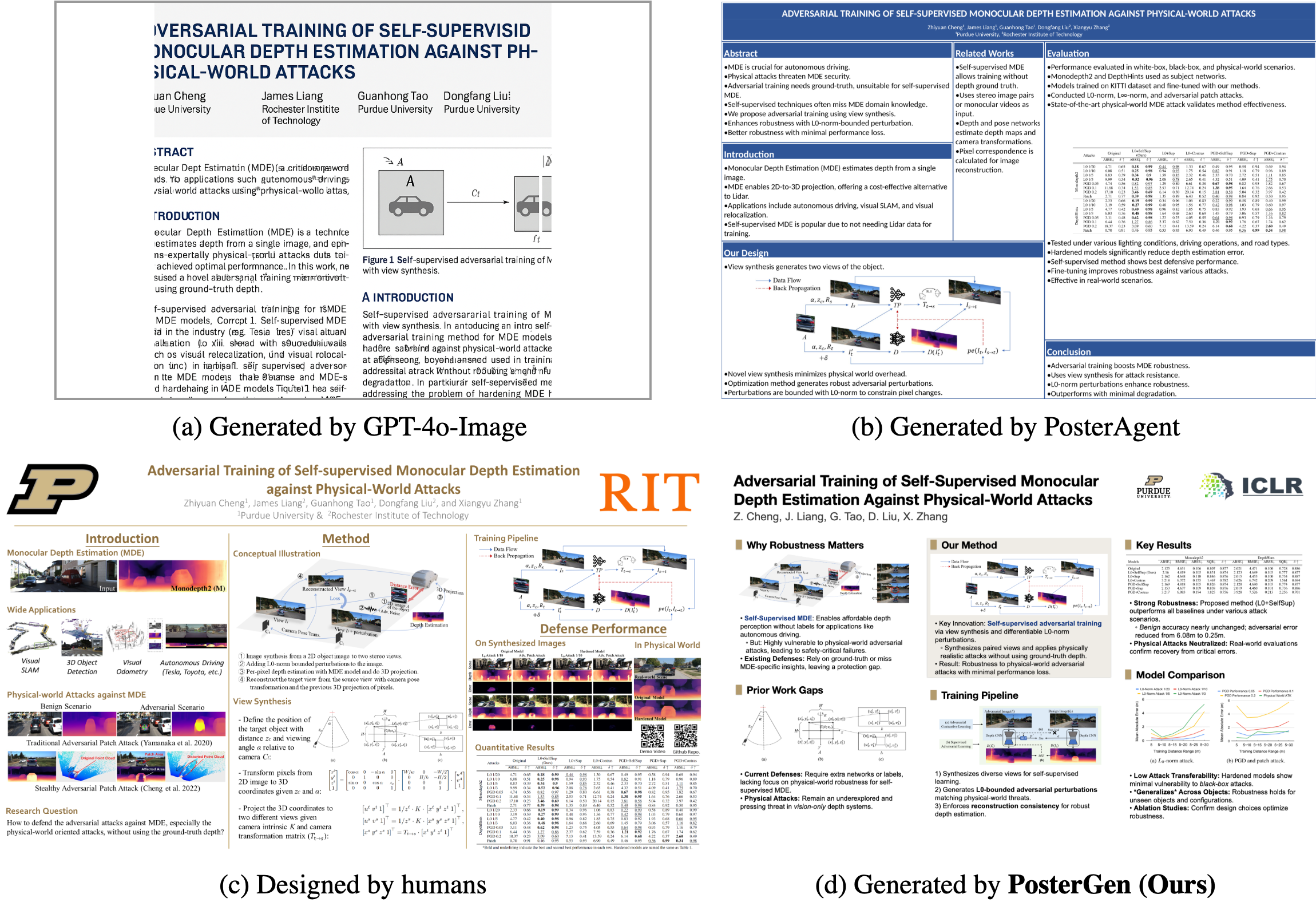}
\caption{Qualitative results of~\citet{cheng2023adversarial}.}
\label{fig::case_study_3}
\end{figure*}
\begin{figure*}[!htp]
\centering
\includegraphics[width=0.8\textwidth]{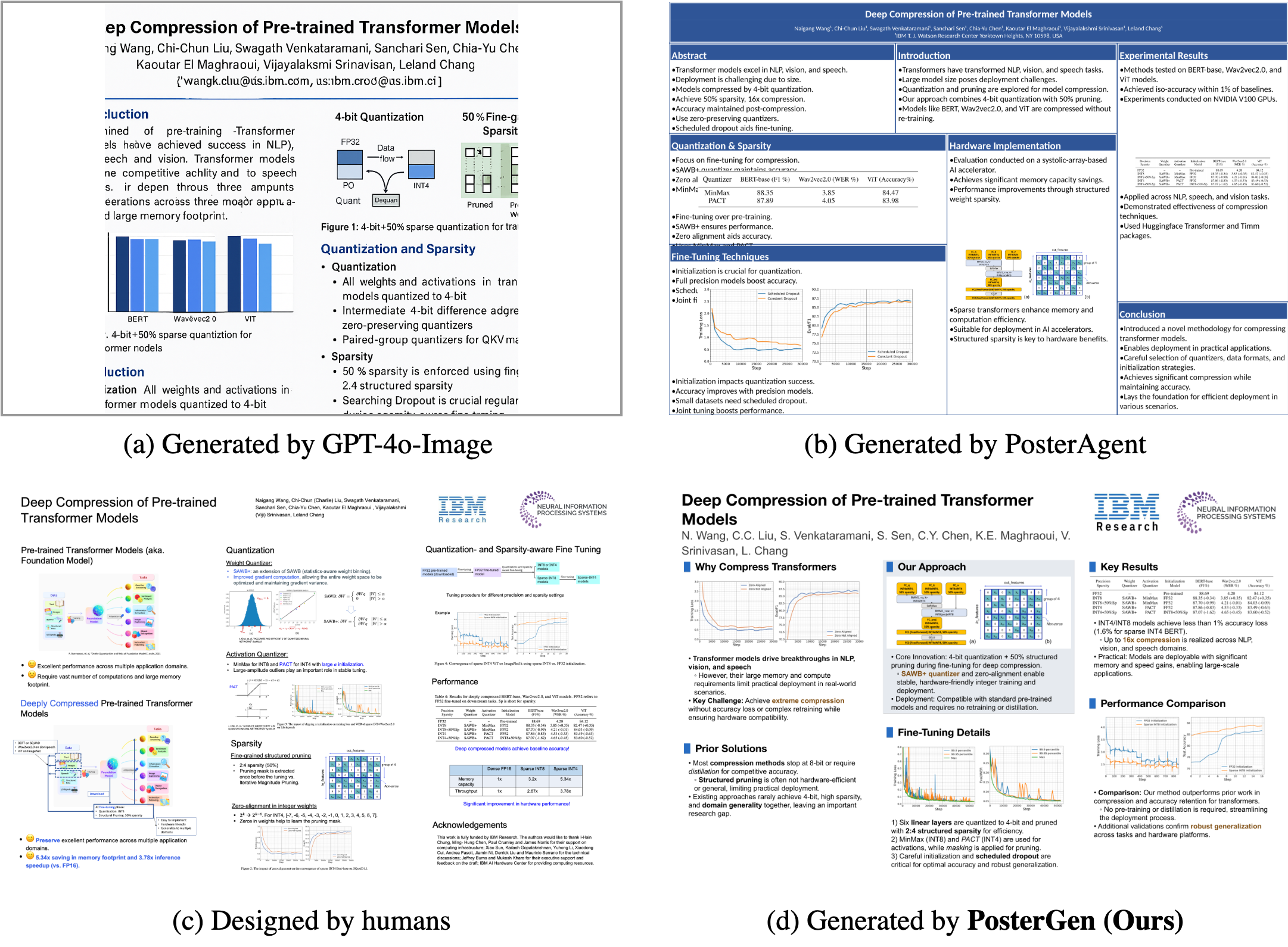}
\caption{Qualitative results of~\citet{wang2022deep}.}
\label{fig::case_study_4}
\end{figure*}
\begin{figure*}[!htp]
\centering
\includegraphics[width=0.8\textwidth]{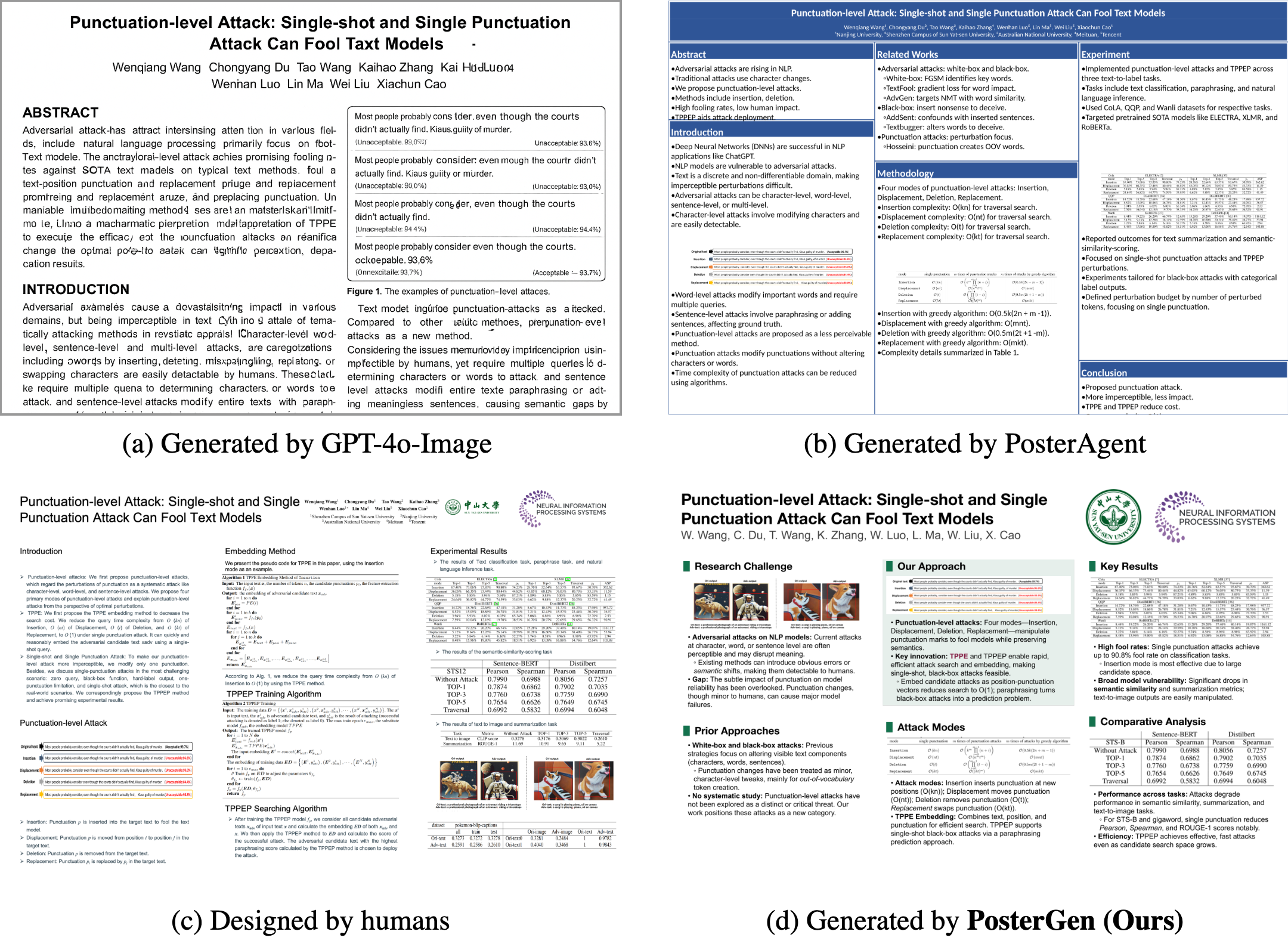}
\caption{Qualitative results of~\citet{du2023punctuation}.}
\label{fig::case_study_5}
\end{figure*}
\begin{figure*}[!htp]
\centering
\includegraphics[width=0.8\textwidth]{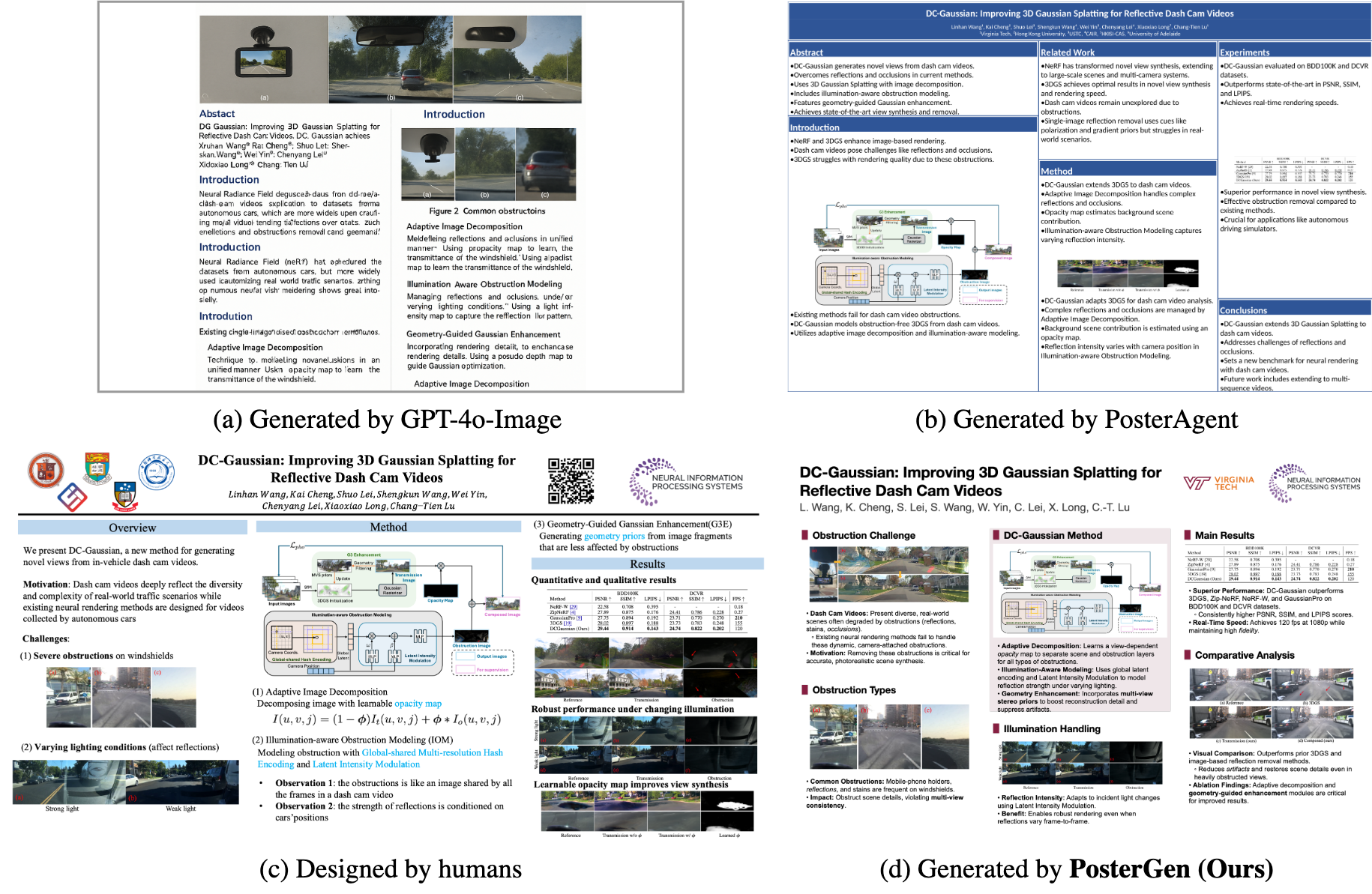}
\caption{Qualitative results of~\citet{wang2024dc}.}
\label{fig::case_study_6}
\end{figure*}
\begin{figure*}[!htp]
\centering
\includegraphics[width=0.8\textwidth]{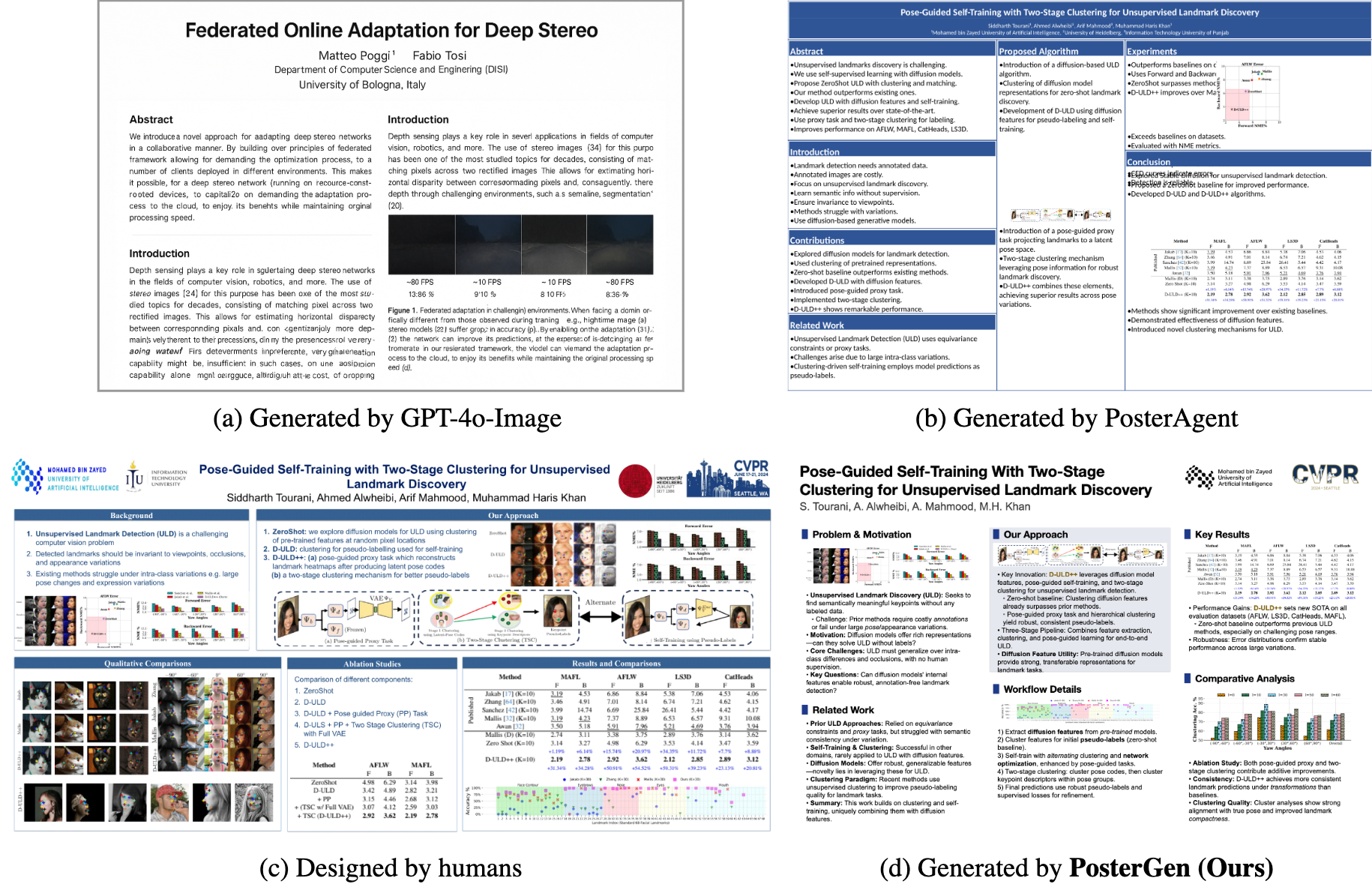}
\caption{Qualitative results of~\citet{tourani2024pose}.}
\label{fig::case_study_7}
\end{figure*}
\begin{figure*}[!htp]
\centering
\includegraphics[width=0.8\textwidth]{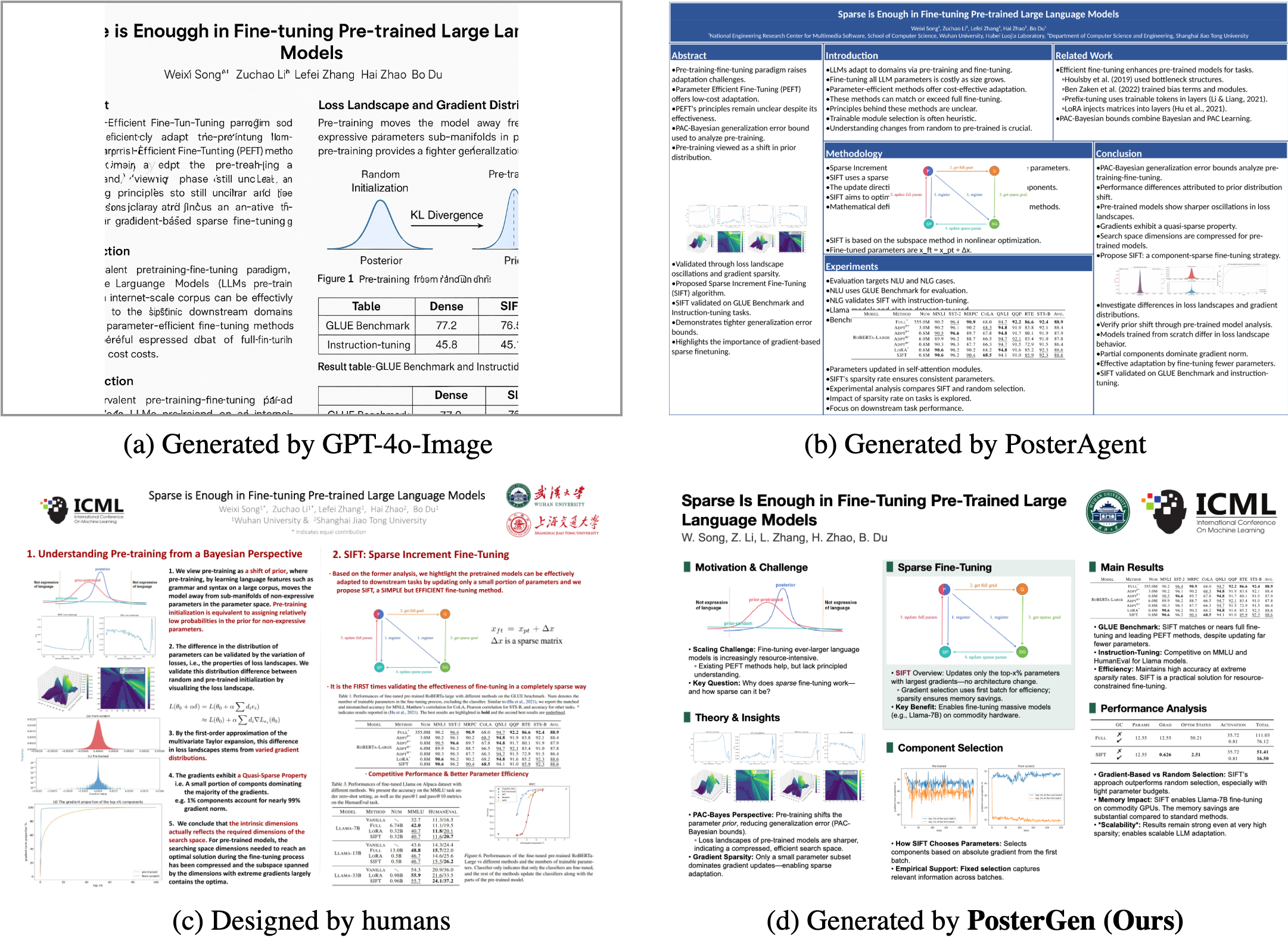}
\caption{Qualitative results of~\citet{song2024sparse}.}
\label{fig::case_study_8}
\end{figure*}
\begin{figure*}[!htp]
\centering
\includegraphics[width=0.8\textwidth]{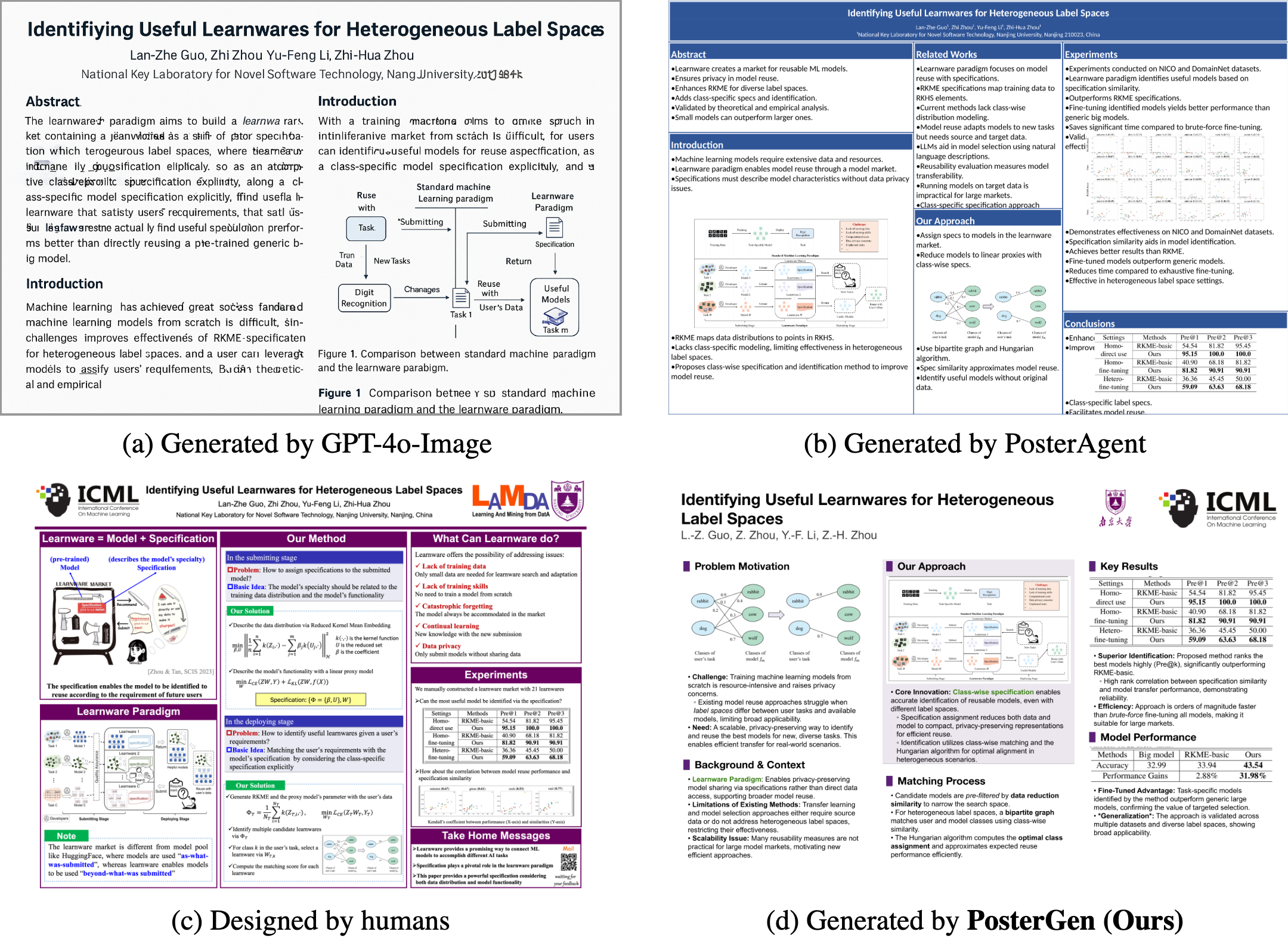}
\caption{Qualitative results of~\citet{guo2023identifying}.}
\label{fig::case_study_9}
\end{figure*}
\begin{figure*}[!htp]
\centering
\includegraphics[width=0.8\textwidth]{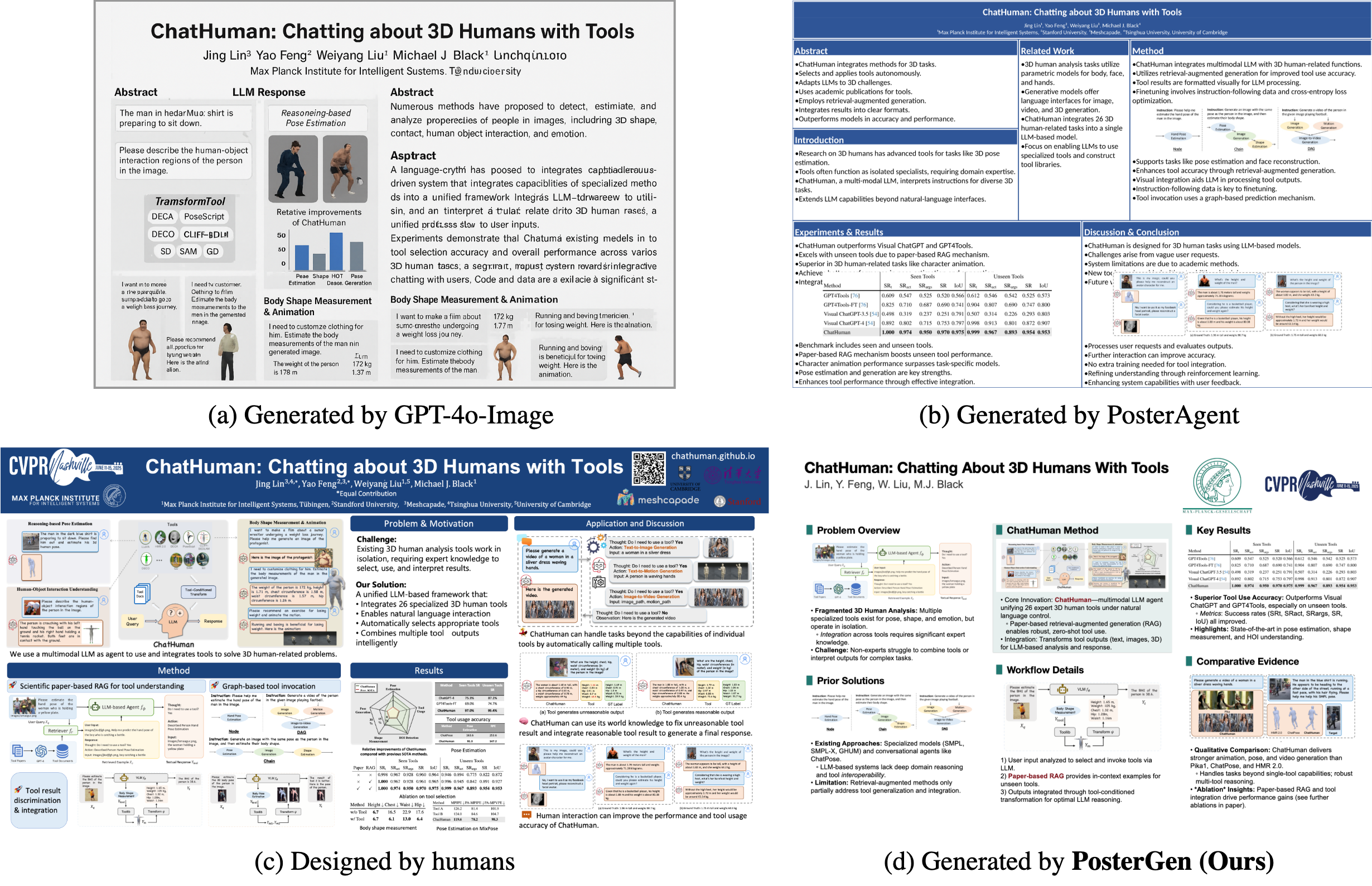}
\caption{Qualitative results of~\citet{lin2025chathuman}.}
\label{fig::case_study_10}
\end{figure*}
\begin{figure*}[!htp]
\centering
\includegraphics[width=0.8\textwidth]{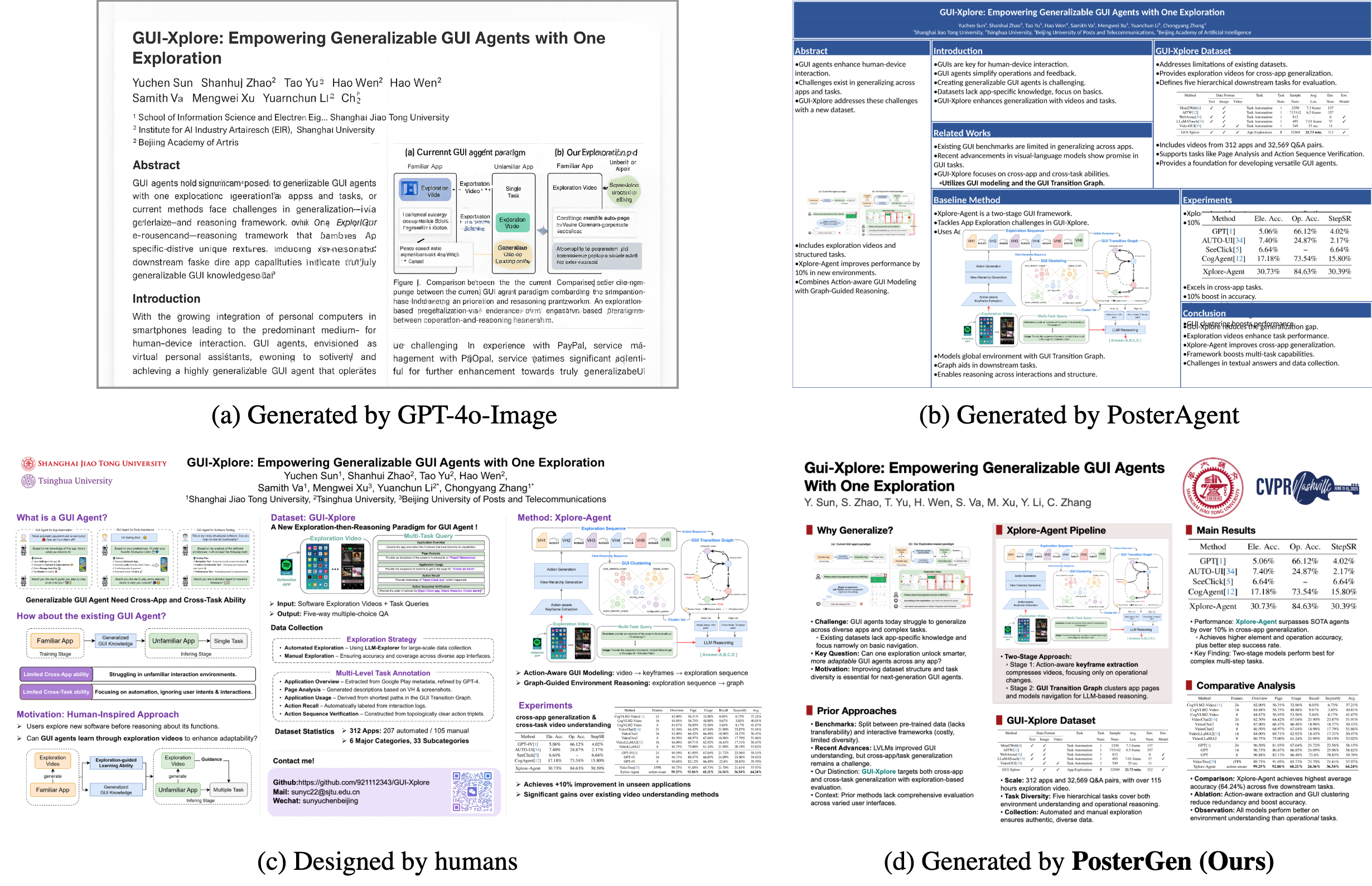}
\caption{Qualitative results of~\citet{sun2025gui}.}
\label{fig::case_study_11}
\end{figure*}
\begin{figure*}[!htp]
\centering
\includegraphics[width=0.8\textwidth]{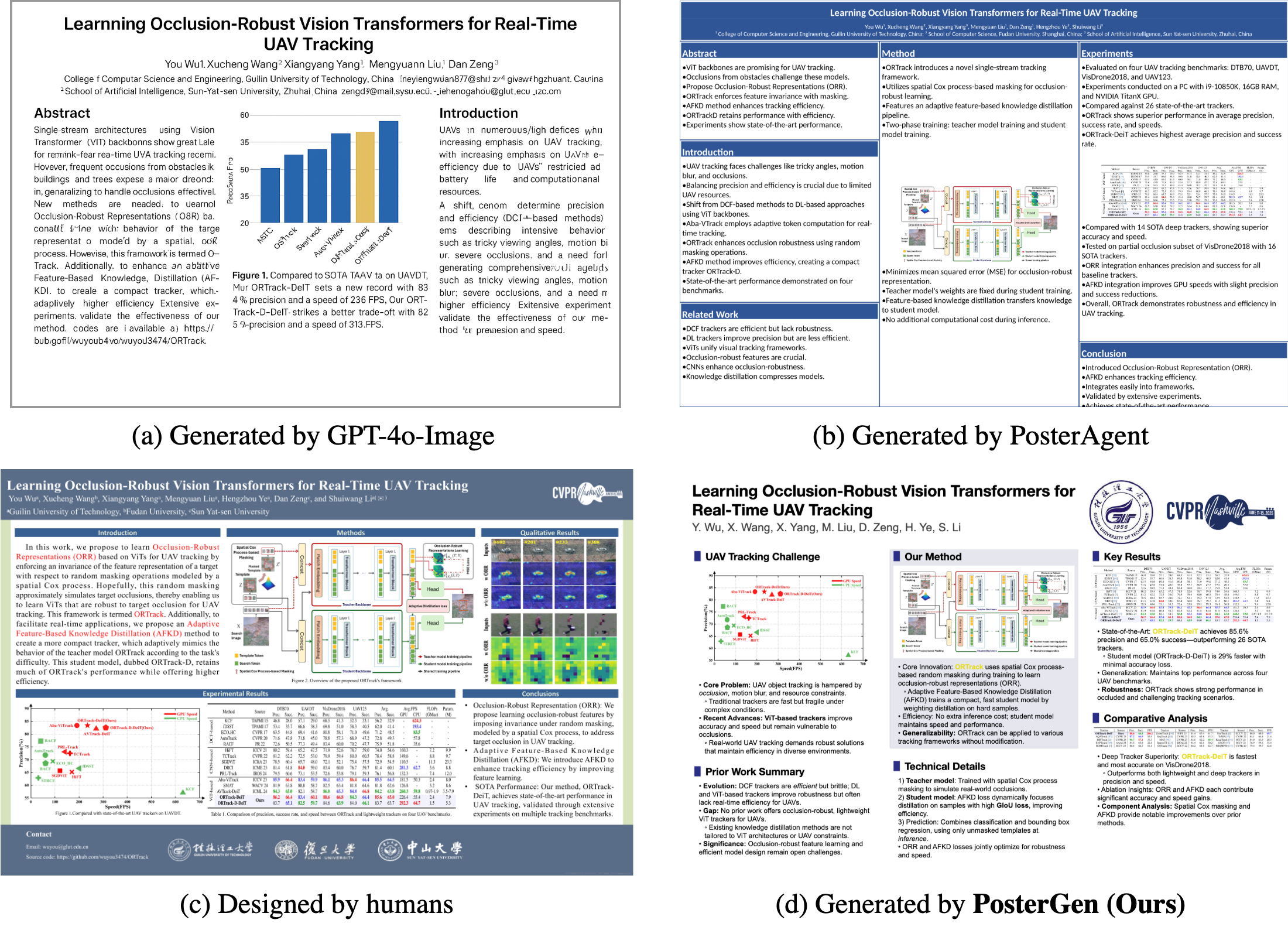}
\caption{Qualitative results of~\citet{wu2025learning}.}
\label{fig::case_study_12}
\end{figure*}
\begin{figure*}[!htp]
\centering
\includegraphics[width=0.8\textwidth]{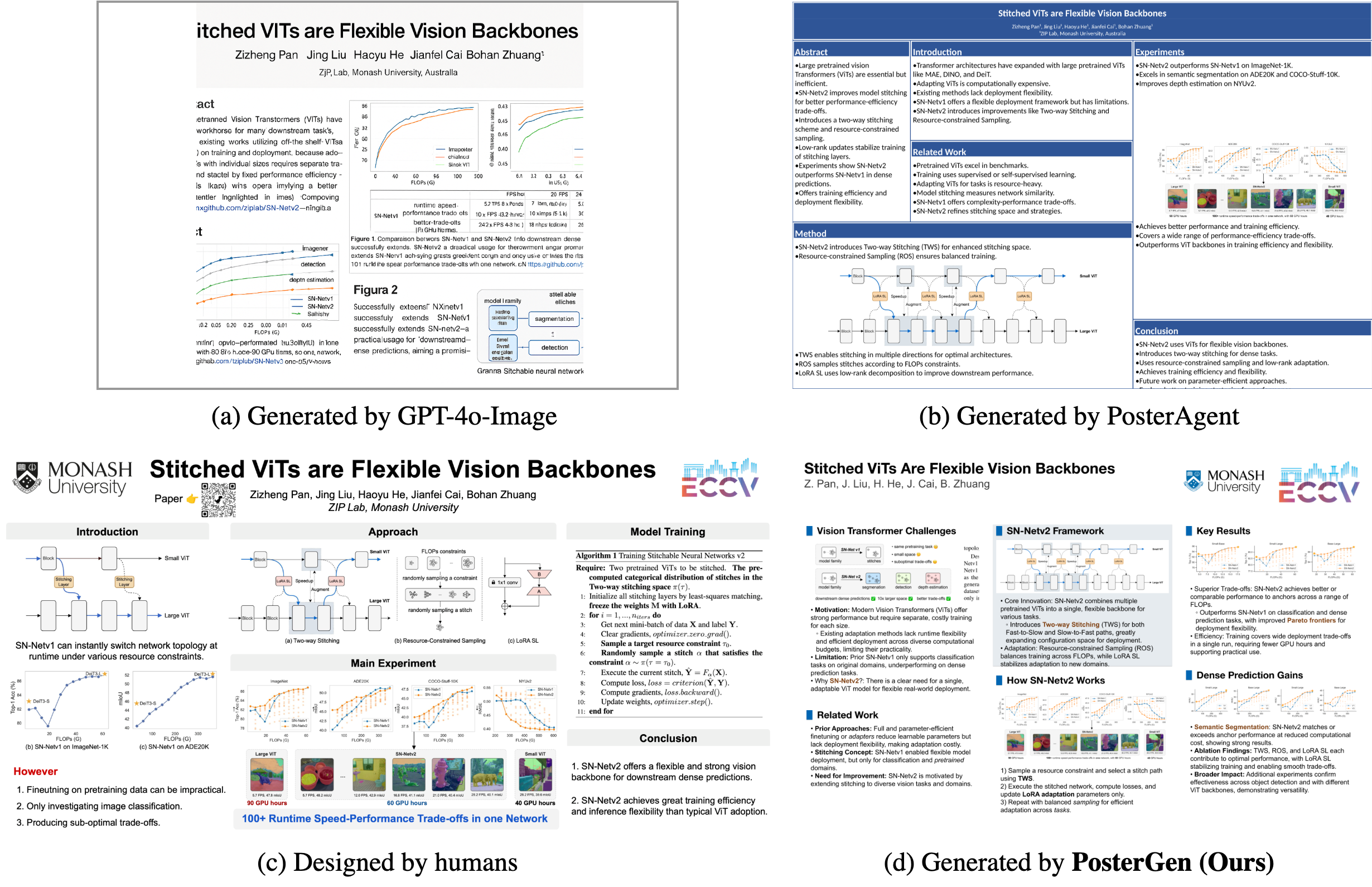}
\caption{Qualitative results of~\citet{pan2024stitched}.}
\label{fig::case_study_13}
\end{figure*}
\begin{figure*}[!htp]
\centering
\includegraphics[width=0.8\textwidth]{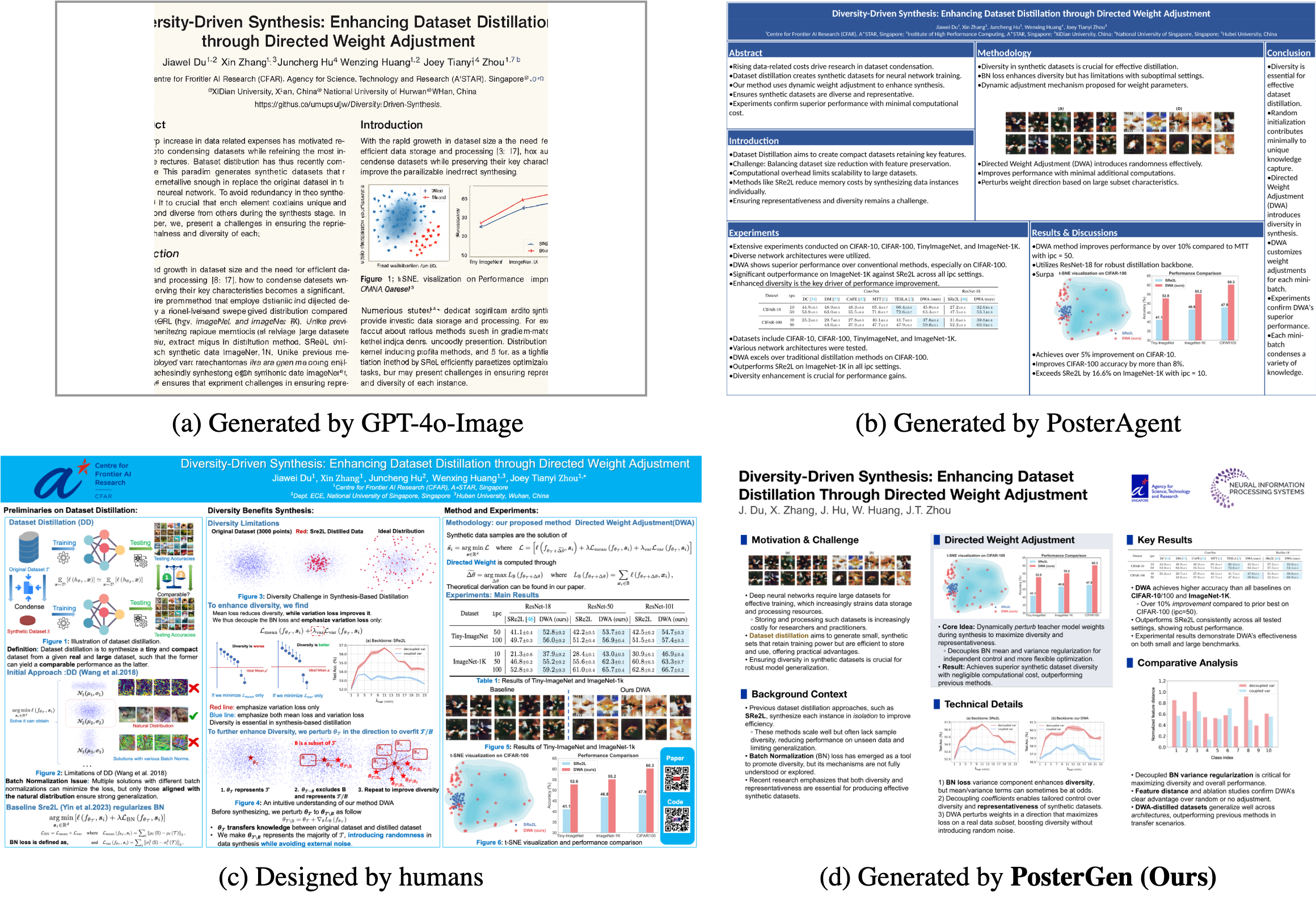}
\caption{Qualitative results of~\citet{du2024diversity}.}
\label{fig::case_study_14}
\end{figure*}
\begin{figure*}[!htp]
\centering
\includegraphics[width=0.8\textwidth]{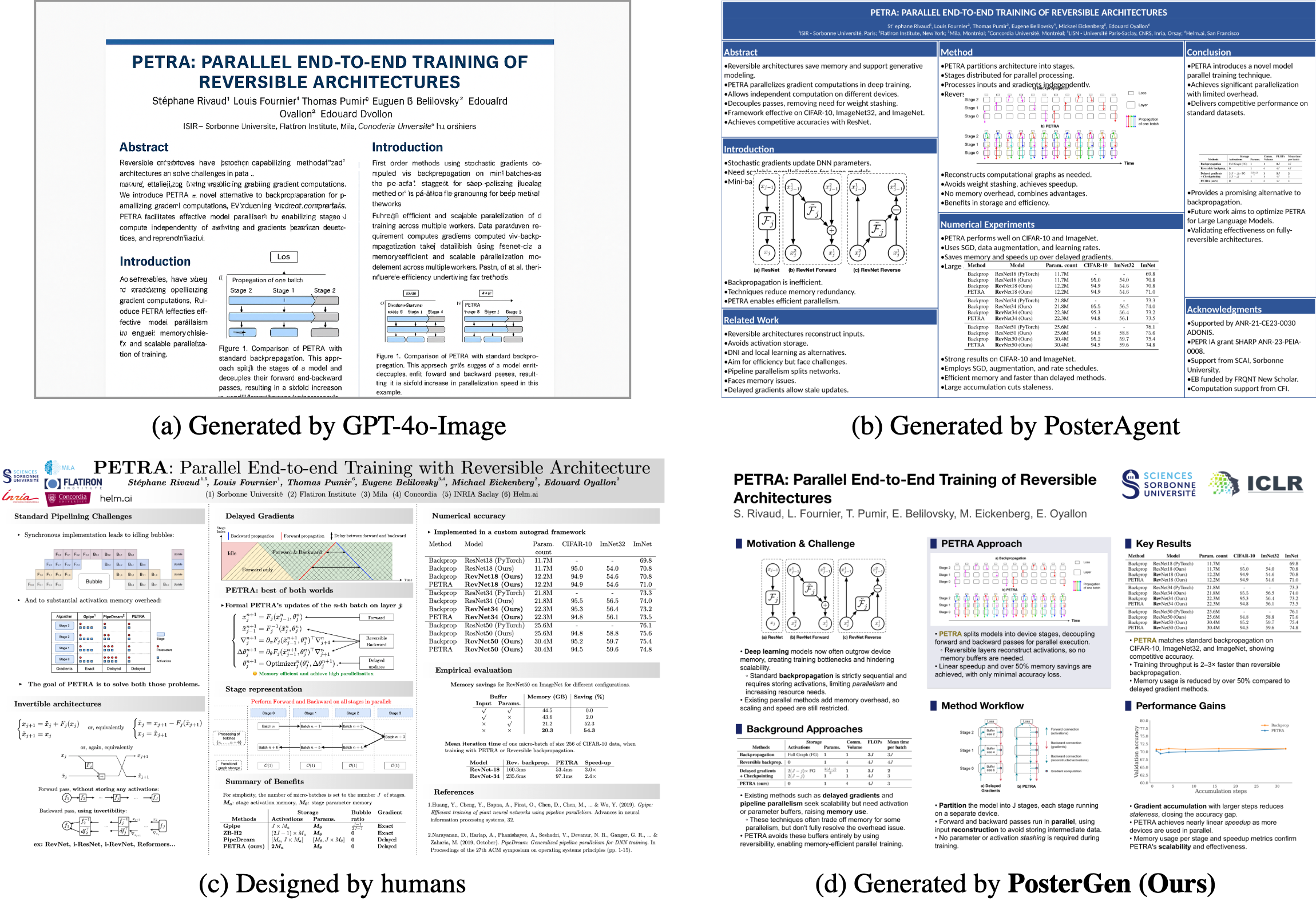}
\caption{Qualitative results of~\citet{rivaud2025petra}.}
\label{fig::case_study_15}
\end{figure*}

In this section, we provide 15 additional representative qualitative results drawn from our 30-paper benchmark, as shown in~\cref{fig::case_study_1,fig::case_study_2,fig::case_study_3,fig::case_study_4,fig::case_study_5,fig::case_study_6,fig::case_study_7,fig::case_study_8,fig::case_study_9,fig::case_study_10,fig::case_study_11,fig::case_study_12,fig::case_study_13,fig::case_study_14,fig::case_study_15}. All the 30 papers were selected from top-tier AI conferences, e.g., NeurIPS, ICML, ICLR, CVPR and ECCV in the last four years, with the distribution shown in \cref{fig::dataset}. Our selection criteria require that both the full paper and a corresponding \underline{high-quality} human-made poster are publicly available.

As illustrated in~\cref{fig::case_study_1,fig::case_study_2,fig::case_study_3,fig::case_study_4,fig::case_study_5,fig::case_study_6,fig::case_study_7,fig::case_study_8,fig::case_study_9,fig::case_study_10,fig::case_study_11,fig::case_study_12,fig::case_study_13,fig::case_study_14,fig::case_study_15}, these supplementary qualitative results further validate the observations presented within the main paper, regarding the limitations of baseline approaches for aesthetic design. The end-to-end GPT-4o-Image method is prone to basic layout constraints, manifesting serious boundary problems like too much whitespace and cut-off vertical content, thus verifying that it is unable to have robust canvas control. Though PosterAgent~\citep{pang2025paperposter} enhances content fidelity over diffusion-based methods, it consistently shows aesthetic inadequacies like improper alignment of elements, poor whitespace use, and a monotonous visual aesthetic that lacks typographical hierarchy to create visual flow. In contrast, PosterGen consistently generates high-quality artistic posters showcasing pleasant colors, effective visual hierarchy generation, and natural reading paths. Therefore, it demonstrates that PosterGen is capable of including effective aesthetic concepts within its agent workflow.

%% file: sec/S12_limitations.tex
Although \textbf{PosterGen} demonstrates significant improvement over existing methods for generating aesthetic-aware posters, we identify several limitations that present directions for future work:
\begin{itemize}
	\item Although the marker-pdf tool is very effective for most use cases for ParserAgent, occasionally it yields errors for extractions. Such artifacts may include lack of text information (for example, title or author names) or possibly for visual components at low resolutions. Areas for improvement may include development of a specialized document parser module designed for academic-style pdf files containing intricate formats.
	\item CuratorAgent is now limited to using only the visual elements it finds in the source paper. It is not capable of creating generative visual elements based on paper content only. One promising direction to take this agent further is to empower its functionality to generate additional diagrams, such as creating a flowchart to intuitively summarize a text-heavy Introduction.
	\item PosterGen is dependent on a three-column grid layout as its current paradigm. This acts as a solid and safe choice for landscape-oriented posters, but does come with its limitations in design flexibility. Going further, the exploration of LayoutAgent should include diverse ratios for various portrait-oriented positions, as well as dynamic and diverse layout designs.
\end{itemize}

%% file: sec/S13_broader_impacts.tex
In this paper, we are the first to explore the integration of design and aesthetic principles into a multi-agent framework for academic poster generation, and propose a novel workflow \textbf{PosterGen}. By mirroring the specialized workflow of professional designers, \textbf{PosterGen} achieves content fidelity that rivals the state-of-the-art while significantly enhancing the final poster's visual design and aesthetic quality. Though it may not fully eliminate the need for human fine-tuning, its core contribution lies in systematically embedding design principles into multi-agent design, a step often overlooked even in manual creation.

We hope our work will advance this emerging yet meaningful field of automated scientific communication, and firmly believe that PosterGen can substantially relieve researchers of the time and effort required for poster creation, allowing them to focus more on the one-on-one scholarly dialogue that poster sessions are meant to facilitate.

%% file: main.bib
@String(AAAI = {AAAI})

@book{faulkes2021better,
  title={Better posters: plan, design and present an academic poster},
  author={Faulkes, Zen},
  year={2021},
  publisher={Pelagic Publishing Ltd}
}

@inproceedings{chen2025posta,
  title={Posta: A go-to framework for customized artistic poster generation},
  author={Chen, Haoyu and Xu, Xiaojie and Li, Wenbo and Ren, Jingjing and Ye, Tian and Liu, Songhua and Chen, Ying-Cong and Zhu, Lei and Wang, Xinchao},
  booktitle={Proceedings of the Computer Vision and Pattern Recognition Conference},
  pages={28694--28704},
  year={2025}
}

@article{chen2025postercraft,
  title={Postercraft: Rethinking high-quality aesthetic poster generation in a unified framework},
  author={Chen, SiXiang and Lai, Jianyu and Gao, Jialin and Ye, Tian and Chen, Haoyu and Shi, Hengyu and Shao, Shitong and Lin, Yunlong and Fei, Song and Xing, Zhaohu and others},
  journal={arXiv preprint arXiv:2506.10741},
  year={2025}
}

@inproceedings{gao2025postermaker,
  title={Postermaker: Towards high-quality product poster generation with accurate text rendering},
  author={Gao, Yifan and Lin, Zihang and Liu, Chuanbin and Zhou, Min and Ge, Tiezheng and Zheng, Bo and Xie, Hongtao},
  booktitle={Proceedings of the Computer Vision and Pattern Recognition Conference},
  pages={8083--8093},
  year={2025}
}

@article{qiang2019learning,
  title={Learning to generate posters of scientific papers by probabilistic graphical models},
  author={Qiang, Yu-Ting and Fu, Yan-Wei and Yu, Xiao and Guo, Yan-Wen and Zhou, Zhi-Hua and Sigal, Leonid},
  journal={Journal of Computer Science and Technology},
  volume={34},
  number={1},
  pages={155--169},
  year={2019},
  publisher={Springer}
}

@inproceedings{xu2022posterbot,
  title={Posterbot: A system for generating posters of scientific papers with neural models},
  author={Xu, Sheng and Wan, Xiaojun},
  booktitle={Proceedings of the AAAI Conference on Artificial Intelligence},
  volume={36},
  number={11},
  pages={13233--13235},
  year={2022}
}

@inproceedings{wang2024scipostlayout,
  title={SciPostLayout: A Dataset for Layout Analysis and Layout Generation of Scientific Posters},
  author={Wang, Hao and Tanaka, Shohei and Ushiku, Yoshitaka},
  booktitle={Proceedings of the IEEE/CVF Conference on Computer Vision and Pattern Recognition},
  pages={8136--8141},
  year={2024}
}

@inproceedings{saxena2025postersum,
  title={Postersum: A multimodal benchmark for scientific poster summarization},
  author={Saxena, Rohit and Minervini, Pasquale and Keller, Frank},
  booktitle={Proceedings of the 14th International Joint Conference on Natural Language Processing and the 4th Conference of the Asia-Pacific Chapter of the Association for Computational Linguistics},
  pages={1828--1844},
  year={2025}
}

@article{liu2022character,
  title={A character-level length-control algorithm for non-autoregressive sentence summarization},
  author={Liu, Puyuan and Zhang, Xiang and Mou, Lili},
  journal={Advances in Neural Information Processing Systems},
  volume={35},
  pages={29101--29112},
  year={2022}
}

@inproceedings{sun2026pp,
  title={P2P: Automated Paper-to-Poster Generation and Fine-Grained Benchmark},
  author={Tao Sun and Enhao Pan and Zhengkai Yang and Kaixin Sui and Jiajun Shi and Xianfu Cheng and Tongliang Li and Ge Zhang and Wenhao Huang and Jian Yang and Zhoujun Li},
  booktitle={The Fourteenth International Conference on Learning Representations},
  year={2026},
  url={https://openreview.net/forum?id=JojyT9niJL}
}

@inproceedings{pang2025paperposter,
  title={Paper2Poster: Towards Multimodal Poster Automation from Scientific Papers},
  author={Wei Pang and Kevin Qinghong Lin and Xiangru Jian and Xi He and Philip Torr},
  booktitle={The Thirty-ninth Annual Conference on Neural Information Processing Systems Datasets and Benchmarks Track},
  year={2025},
  url={https://openreview.net/forum?id=p0E74lpRBD}
}

@inproceedings{seol2024posterllama,
  title={Posterllama: Bridging design ability of language model to content-aware layout generation},
  author={Seol, Jaejung and Kim, Seojun and Yoo, Jaejun},
  booktitle={European Conference on Computer Vision},
  pages={451--468},
  year={2024},
  organization={Springer}
}

@article{zhang2025creatidesign,
  title={Creatidesign: A unified multi-conditional diffusion transformer for creative graphic design},
  author={Zhang, Hui and Hong, Dexiang and Yang, Maoke and Cheng, Yutao and Zhang, Zhao and Shao, Jie and Wu, Xinglong and Wu, Zuxuan and Jiang, Yu-Gang},
  journal={arXiv preprint arXiv:2505.19114},
  year={2025}
}

@article{guo2024large,
  title={Large language model based multi-agents: A survey of progress and challenges},
  author={Guo, Taicheng and Chen, Xiuying and Wang, Yaqi and Chang, Ruidi and Pei, Shichao and Chawla, Nitesh V and Wiest, Olaf and Zhang, Xiangliang},
  journal={arXiv preprint arXiv:2402.01680},
  year={2024}
}

@article{li2023camel,
  title={Camel: Communicative agents for" mind" exploration of large language model society},
  author={Li, Guohao and Hammoud, Hasan and Itani, Hani and Khizbullin, Dmitrii and Ghanem, Bernard},
  journal={Advances in neural information processing systems},
  volume={36},
  pages={51991--52008},
  year={2023}
}

@article{yin2023ttida,
  title={Ttida: controllable generative data augmentation via text-to-text and text-to-image models},
  author={Yin, Yuwei and Kaddour, Jean and Zhang, Xiang and Nie, Yixin and Liu, Zhenguang and Kong, Lingpeng and Liu, Qi},
  journal={arXiv preprint arXiv:2304.08821},
  year={2023}
}

@inproceedings{liu2024agentbench,
  title={AgentBench: Evaluating {LLM}s as Agents},
  author={Xiao Liu and Hao Yu and Hanchen Zhang and Yifan Xu and Xuanyu Lei and Hanyu Lai and Yu Gu and Hangliang Ding and Kaiwen Men and Kejuan Yang and Shudan Zhang and Xiang Deng and Aohan Zeng and Zhengxiao Du and Chenhui Zhang and Sheng Shen and Tianjun Zhang and Yu Su and Huan Sun and Minlie Huang and Yuxiao Dong and Jie Tang},
  booktitle={The Twelfth International Conference on Learning Representations},
  year={2024},
  url={https://openreview.net/forum?id=zAdUB0aCTQ}
}

@inproceedings{jin2024contranovo,
  title={Contranovo: A contrastive learning approach to enhance de novo peptide sequencing},
  author={Jin, Zhi and Xu, Sheng and Zhang, Xiang and Ling, Tianze and Dong, Nanqing and Ouyang, Wanli and Gao, Zhiqiang and Chang, Cheng and Sun, Siqi},
  booktitle={Proceedings of the AAAI conference on artificial intelligence},
  volume={38},
  number={1},
  pages={144--152},
  year={2024}
}

@article{zhang2024autoregressive+,
  title={Autoregressive+ Chain of Thought= Recurrent: Recurrence's Role in Language Models' Computability and a Revisit of Recurrent Transformer},
  author={Zhang, Xiang and Abdul-Mageed, Muhammad and Lakshmanan, Laks VS},
  journal={arXiv preprint arXiv:2409.09239},
  year={2024}
}

@article{zhang2024cross,
  title={Cross-modal consistency in multimodal large language models},
  author={Zhang, Xiang and Li, Senyu and Shi, Ning and Hauer, Bradley and Wu, Zijun and Kondrak, Grzegorz and Abdul-Mageed, Muhammad and Lakshmanan, Laks VS},
  journal={arXiv preprint arXiv:2411.09273},
  year={2024}
}

@inproceedings{wu2024autogen,
  title={Autogen: Enabling next-gen LLM applications via multi-agent conversations},
  author={Wu, Qingyun and Bansal, Gagan and Zhang, Jieyu and Wu, Yiran and Li, Beibin and Zhu, Erkang and Jiang, Li and Zhang, Xiaoyun and Zhang, Shaokun and Liu, Jiale and others},
  booktitle={First conference on language modeling},
  year={2024}
}

@inproceedings{cao2025multi2,
  title={Multi2: Multi-agent test-time scalable framework for multi-document processing},
  author={Cao, Juntai and Zhang, Xiang and Li, Raymond and Wei, Jiaqi and Li, Chuyuan and Joty, Shafiq and Carenini, Giuseppe},
  booktitle={Proceedings of The 5th New Frontiers in Summarization Workshop},
  pages={135--156},
  year={2025}
}

@article{bo2024reflective,
  title={Reflective multi-agent collaboration based on large language models},
  author={Bo, Xiaohe and Zhang, Zeyu and Dai, Quanyu and Feng, Xueyang and Wang, Lei and Li, Rui and Chen, Xu and Wen, Ji-Rong},
  journal={Advances in Neural Information Processing Systems},
  volume={37},
  pages={138595--138631},
  year={2024}
}

@inproceedings{wei2025retrieval,
  title={Retrieval is Not Enough: Enhancing {RAG} through Test-Time Critique and Optimization},
  author={Jiaqi Wei and Hao Zhou and Xiang Zhang and Di Zhang and Zijie Qiu and Noah Wei and Jinzhe Li and Wanli Ouyang and Siqi Sun},
  booktitle={The Thirty-ninth Annual Conference on Neural Information Processing Systems},
  year={2025},
  url={https://openreview.net/forum?id=cnUq7GkS6d}
}

@article{lin2024designprobe,
  title={Designprobe: A graphic design benchmark for multimodal large language models},
  author={Lin, Jieru and Huang, Danqing and Zhao, Tiejun and Zhan, Dechen and Lin, Chin-Yew},
  journal={arXiv preprint arXiv:2404.14801},
  year={2024}
}

@inproceedings{cheng2025graphic,
  title={Graphic design with large multimodal model},
  author={Cheng, Yutao and Zhang, Zhao and Yang, Maoke and Nie, Hui and Li, Chunyuan and Wu, Xinglong and Shao, Jie},
  booktitle={Proceedings of the AAAI Conference on Artificial Intelligence},
  volume={39},
  number={3},
  pages={2473--2481},
  year={2025}
}

@inproceedings{hsu2025postero,
  title={Postero: Structuring layout trees to enable language models in generalized content-aware layout generation},
  author={Hsu, HsiaoYuan and Peng, Yuxin},
  booktitle={Proceedings of the Computer Vision and Pattern Recognition Conference},
  pages={8117--8127},
  year={2025}
}

@inproceedings{patnaik2025aesthetiq,
  title={Aesthetiq: Enhancing graphic layout design via aesthetic-aware preference alignment of multi-modal large language models},
  author={Patnaik, Sohan and Jain, Rishabh and Krishnamurthy, Balaji and Sarkar, Mausoom},
  booktitle={Proceedings of the Computer Vision and Pattern Recognition Conference},
  pages={23701--23711},
  year={2025}
}

@book{olson2019narrative,
  title={Narrative is everything: The ABT framework and narrative evolution},
  author={Olson, Randy},
  year={2019},
  publisher={Prairie Starfish Productions}
}

@misc{paruchuri2025marker,
  title={Marker: Convert PDF to Markdown and JSON Quickly with High Accuracy},
  author={Paruchuri, Vik},
  year={2025}
}

@inproceedings{zhang2025neural,
  title={Neural Encoding and Decoding at Scale},
  author={Yizi Zhang and Yanchen Wang and Mehdi Azabou and Alexandre Andre and Zixuan Wang and Hanrui Lyu and International Brain Laboratory and Eva L Dyer and Liam Paninski and Cole Lincoln Hurwitz},
  booktitle={Forty-second International Conference on Machine Learning},
  year={2025},
  url={https://openreview.net/forum?id=vOdz3zhSCj}
}

@inproceedings{zhou2024segmentation,
  title={Segmentation-guided layer-wise image vectorization with gradient fills},
  author={Zhou, Hengyu and Zhang, Hui and Wang, Bin},
  booktitle={European Conference on Computer Vision},
  pages={165--180},
  year={2024},
  organization={Springer}
}

@inproceedings{xue2025comfybench,
  title={Comfybench: Benchmarking llm-based agents in comfyui for autonomously designing collaborative ai systems},
  author={Xue, Xiangyuan and Lu, Zeyu and Huang, Di and Wang, Zidong and Ouyang, Wanli and Bai, Lei},
  booktitle={Proceedings of the computer vision and pattern recognition conference},
  pages={24614--24624},
  year={2025}
}

@inproceedings{cao2025phishagent,
  title={Phishagent: a robust multimodal agent for phishing webpage detection},
  author={Cao, Tri and Huang, Chengyu and Li, Yuexin and Huilin, Wang and He, Amy and Oo, Nay and Hooi, Bryan},
  booktitle={Proceedings of the AAAI Conference on Artificial Intelligence},
  volume={39},
  number={27},
  pages={27869--27877},
  year={2025}
}

@inproceedings{mondal2024presentations,
  title={Presentations by the humans and for the humans: Harnessing llms for generating persona-aware slides from documents},
  author={Mondal, Ishani and Shwetha, S and Natarajan, Anandhavelu and Garimella, Aparna and Bandyopadhyay, Sambaran and Boyd-Graber, Jordan},
  booktitle={Proceedings of the 18th Conference of the European Chapter of the Association for Computational Linguistics (Volume 1: Long Papers)},
  pages={2664--2684},
  year={2024}
}

@inproceedings{zheng2025pptagent,
  title={Pptagent: Generating and evaluating presentations beyond text-to-slides},
  author={Zheng, Hao and Guan, Xinyan and Kong, Hao and Zhang, Wenkai and Zheng, Jia and Zhou, Weixiang and Lin, Hongyu and Lu, Yaojie and Han, Xianpei and Sun, Le},
  booktitle={Proceedings of the 2025 Conference on Empirical Methods in Natural Language Processing},
  pages={14413--14429},
  year={2025}
}

@article{jung2025talk,
  title={Talk to your slides: Language-driven agents for efficient slide editing},
  author={Jung, Kyudan and Cho, Hojun and Yun, Jooyeol and Yang, Soyoung and Jang, Jaehyeok and Choo, Jaegul},
  journal={arXiv preprint arXiv:2505.11604},
  year={2025}
}

@inproceedings{fu2022doc2ppt,
  title={Doc2ppt: Automatic presentation slides generation from scientific documents},
  author={Fu, Tsu-Jui and Wang, William Yang and McDuff, Daniel and Song, Yale},
  booktitle={Proceedings of the AAAI Conference on Artificial Intelligence},
  volume={36},
  number={1},
  pages={634--642},
  year={2022}
}

@article{hu2014ppsgen,
  title={PPSGen: Learning-based presentation slides generation for academic papers},
  author={Hu, Yue and Wan, Xiaojun},
  journal={IEEE transactions on knowledge and data engineering},
  volume={27},
  number={4},
  pages={1085--1097},
  year={2014},
  publisher={IEEE}
}

@article{zhang2025tokenization,
  title={Tokenization constraints in llms: A study of symbolic and arithmetic reasoning limits},
  author={Zhang, Xiang and Cao, Juntai and Wei, Jiaqi and Xu, Yiwei and You, Chenyu},
  journal={arXiv preprint arXiv:2505.14178},
  year={2025}
}

@article{kumar2024slidespawn,
  title={Slidespawn: An automatic slides generation system for research publications},
  author={Kumar, Keshav and Chowdary, Ravindranath},
  journal={arXiv preprint arXiv:2411.17719},
  year={2024}
}

@inproceedings{sravanthi2009slidesgen,
  title={SlidesGen: Automatic Generation of Presentation Slides for a Technical Paper Using Summarization.},
  author={Sravanthi, M and Chowdary, C Ravindranath and Kumar, P Sreenivasa},
  booktitle={FLAIRS},
  year={2009}
}

@inproceedings{shi2025presentagent,
  title={Presentagent: Multimodal agent for presentation video generation},
  author={Shi, Jingwei and Zhang, Zeyu and Wu, Biao and Liang, Yanjie and Fang, Meng and Chen, Ling and Zhao, Yang},
  booktitle={Proceedings of the 2025 Conference on Empirical Methods in Natural Language Processing: System Demonstrations},
  pages={760--773},
  year={2025}
}

@inproceedings{zhang2025prompt,
  title={Why prompt design matters and works: A complexity analysis of prompt search space in llms},
  author={Zhang, Xiang and Cao, Juntai and You, Chenyu and Ding, Dujian},
  booktitle={Proceedings of the 63rd Annual Meeting of the Association for Computational Linguistics (Volume 1: Long Papers)},
  pages={32525--32555},
  year={2025}
}

@inproceedings{oh2022differentially,
  title={Differentially Private CutMix for Split Learning with Vision Transformer},
  author={Seungeun Oh and Jihong Park and Sihun Baek and Hyelin Nam and Praneeth Vepakomma and Ramesh Raskar and Mehdi Bennis and Seong-Lyun Kim},
  booktitle={ First Workshop on Interpolation Regularizers and Beyond at NeurIPS 2022},
  year={2022},
  url={https://openreview.net/forum?id=gRCWdltNQq}
}

@inproceedings{hong2024cas,
  title={Cas: A probability-based approach for universal condition alignment score},
  author={Hong, Chunsan and Cha, ByungHee and Oh, Tae-Hyun},
  booktitle={The Twelfth International Conference on Learning Representations},
  year={2024}
}

@inproceedings{cheng2023adversarial,
  title={Adversarial Training of Self-supervised Monocular Depth Estimation against Physical-World Attacks},
  author={Zhiyuan Cheng and James Chenhao Liang and Guanhong Tao and Dongfang Liu and Xiangyu Zhang},
  booktitle={The Eleventh International Conference on Learning Representations },
  year={2023},
  url={https://openreview.net/forum?id=LfdEuhjR5GV}
}

@article{wang2022deep,
  title={Deep compression of pre-trained transformer models},
  author={Wang, Naigang and Liu, Chi-Chun Charlie and Venkataramani, Swagath and Sen, Sanchari and Chen, Chia-Yu and El Maghraoui, Kaoutar and Srinivasan, Vijayalakshmi Viji and Chang, Leland},
  journal={Advances in Neural Information Processing Systems},
  volume={35},
  pages={14140--14154},
  year={2022}
}

@article{du2023punctuation,
  title={Punctuation-level attack: Single-shot and single punctuation can fool text models},
  author={Du, Chongyang and Wang, Tao and Zhang, Kaihao and Luo, Wenhan and Ma, Lin and Liu, Wei and Cao, Xiaochun and others},
  journal={Advances in Neural Information Processing Systems},
  volume={36},
  pages={49312--49324},
  year={2023}
}

@article{wang2024dc,
  title={DC-Gaussian: Improving 3D Gaussian splatting for reflective dash cam videos},
  author={Wang, Linhan and Cheng, Kai and Lei, Shuo and Wang, Shengkun and Yin, Wei and Lei, Chenyang and Long, Xiaoxiao and Lu, Chang-Tien},
  journal={Advances in Neural Information Processing Systems},
  volume={37},
  pages={99898--99920},
  year={2024}
}

@inproceedings{tourani2024pose,
  title={Pose-guided self-training with two-stage clustering for unsupervised landmark discovery},
  author={Tourani, Siddharth and Alwheibi, Ahmed and Mahmood, Arif and Khan, Muhammad Haris},
  booktitle={Proceedings of the IEEE/CVF Conference on Computer Vision and Pattern Recognition},
  pages={23041--23051},
  year={2024}
}

@inproceedings{song2024sparse,
  title={Sparse is Enough in Fine-tuning Pre-trained Large Language Models},
  author={Weixi Song and Zuchao Li and Lefei Zhang and hai zhao and Bo Du},
  booktitle={Forty-first International Conference on Machine Learning},
  year={2024},
  url={https://openreview.net/forum?id=10hu2D3hAg}
}

@inproceedings{guo2023identifying,
  title={Identifying useful learnwares for heterogeneous label spaces},
  author={Guo, Lan-Zhe and Zhou, Zhi and Li, Yu-Feng and Zhou, Zhi-Hua},
  booktitle={International Conference on Machine Learning},
  pages={12122--12131},
  year={2023},
  organization={PMLR}
}

@inproceedings{lin2025chathuman,
  title={Chathuman: Chatting about 3d humans with tools},
  author={Lin, Jing and Feng, Yao and Liu, Weiyang and Black, Michael J},
  booktitle={Proceedings of the Computer Vision and Pattern Recognition Conference},
  pages={8150--8161},
  year={2025}
}

@inproceedings{sun2025gui,
  title={Gui-xplore: Empowering generalizable gui agents with one exploration},
  author={Sun, Yuchen and Zhao, Shanhui and Yu, Tao and Wen, Hao and Va, Samith and Xu, Mengwei and Li, Yuanchun and Zhang, Chongyang},
  booktitle={Proceedings of the computer vision and pattern recognition conference},
  pages={19477--19486},
  year={2025}
}

@inproceedings{wu2025learning,
  title={Learning occlusion-robust vision transformers for real-time uav tracking},
  author={Wu, You and Wang, Xucheng and Yang, Xiangyang and Liu, Mengyuan and Zeng, Dan and Ye, Hengzhou and Li, Shuiwang},
  booktitle={Proceedings of the IEEE/CVF Conference on Computer Vision and Pattern Recognition},
  pages={17103--17113},
  year={2025}
}

@inproceedings{pan2024stitched,
  title={Stitched vits are flexible vision backbones},
  author={Pan, Zizheng and Liu, Jing and He, Haoyu and Cai, Jianfei and Zhuang, Bohan},
  booktitle={European Conference on Computer Vision},
  pages={258--274},
  year={2024},
  organization={Springer}
}

@article{du2024diversity,
  title={Diversity-driven synthesis: Enhancing dataset distillation through directed weight adjustment},
  author={Du, Jiawei and Zhang, Xin and Hu, Juncheng and Huang, Wenxing and Zhou, Joey T},
  journal={Advances in neural information processing systems},
  volume={37},
  pages={119443--119465},
  year={2024}
}

@inproceedings{rivaud2025petra,
  title={{PETRA}: Parallel End-to-end Training with Reversible Architectures},
  author={Stephane Rivaud and Louis Fournier and Thomas Pumir and Eugene Belilovsky and Michael Eickenberg and Edouard Oyallon},
  booktitle={The Thirteenth International Conference on Learning Representations},
  year={2025},
  url={https://openreview.net/forum?id=0fhzSFsGUT}
}

@inproceedings{you2024calibrating,
  title={Calibrating multi-modal representations: A pursuit of group robustness without annotations},
  author={You, Chenyu and Mint, Yifei and Dai, Weicheng and Sekhon, Jasjeet S and Staib, Lawrence and Duncan, James S},
  booktitle={IEEE/CVF Conference on Computer Vision and Pattern Recognition},
  pages={26140--26150},
  year={2024}
}

@article{you2025uncovering,
  title={Uncovering memorization effect in the presence of spurious correlations},
  author={You, Chenyu and Dai, Haocheng and Min, Yifei and Sekhon, Jasjeet S and Joshi, Sarang and Duncan, James S},
  journal={Nature Communications},
  volume={16},
  number={1},
  pages={5424},
  year={2025},
  publisher={Nature Publishing Group UK London}
}

@inproceedings{sun2026coma,
  title={Coma: Compositional human motion generation with multi-modal agents},
  author={Sun, Shanlin and Xu, Jiaqi and de Araujo, Gabriel and Zhou, Shenghan and Zhang, Hanwen and Huang, Ziheng and You, Chenyu and Xie, Xiaohui},
  booktitle={Proceedings of the AAAI Conference on Artificial Intelligence},
  volume={40},
  number={11},
  pages={9206--9214},
  year={2026}
}

@article{pan2025beyond,
  title={Beyond benchmarks: Dynamic, automatic and systematic red-teaming agents for trustworthy medical language models},
  author={Pan, Jiazhen and Jian, Bailiang and Hager, Paul and Zhang, Yundi and Liu, Che and Jungmann, Friedrike and Li, Hongwei Bran and You, Chenyu and Wu, Junde and Zhu, Jiayuan and others},
  journal={arXiv preprint arXiv:2508.00923},
  year={2025}
}

@article{wei2025ai,
  title={From ai for science to agentic science: A survey on autonomous scientific discovery},
  author={Wei, Jiaqi and Yang, Yuejin and Zhang, Xiang and Chen, Yuhan and Zhuang, Xiang and Gao, Zhangyang and Zhou, Dongzhan and Wang, Guangshuai and Gao, Zhiqiang and Cao, Juntai and others},
  journal={arXiv preprint arXiv:2508.14111},
  year={2025}
}

@article{xiong2025quantagent,
  title={Quantagent: Price-driven multi-agent llms for high-frequency trading},
  author={Xiong, Fei and Zhang, Xiang and Feng, Aosong and Sun, Siqi and You, Chenyu},
  journal={arXiv preprint arXiv:2509.09995},
  year={2025}
}

@article{zhao2025timeseriesscientist,
  title={Timeseriesscientist: A general-purpose ai agent for time series analysis},
  author={Zhao, Haokun and Zhang, Xiang and Wei, Jiaqi and Xu, Yiwei and He, Yuting and Sun, Siqi and You, Chenyu},
  journal={arXiv preprint arXiv:2510.01538},
  year={2025}
}

@inproceedings{sun2025docagent,
  title={Docagent: An agentic framework for multi-modal long-context document understanding},
  author={Sun, Li and He, Liu and Jia, Shuyue and He, Yangfan and You, Chenyu},
  booktitle={Proceedings of the 2025 Conference on Empirical Methods in Natural Language Processing},
  pages={17712--17727},
  year={2025}
}

@article{liang2025slidegen,
  title={Slidegen: Collaborative multimodal agents for scientific slide generation},
  author={Liang, Xin and Zhang, Xiang and Xu, Yiwei and Sun, Siqi and You, Chenyu},
  journal={arXiv preprint arXiv:2512.04529},
  year={2025}
}
